\documentclass[letterpaper, 10 pt, conference]{ieeeconf}  

\usepackage{svg}
\usepackage{amsmath, amssymb}
\usepackage{enumerate}

\usepackage{hyperref}
\usepackage{cleveref}

\usepackage{multirow}
\usepackage{tabularx,booktabs}
\newcolumntype{C}{>{\centering\arraybackslash}X} 
\newcolumntype{R}{>{\raggedleft\arraybackslash}X} 

\usepackage{esvect}

\usepackage{pdfpages}

\usepackage{blindtext}
\usepackage{xcolor}
\usepackage{tikz}
\usetikzlibrary{shapes,
	arrows,
	positioning,
	fit,
	backgrounds,
	calc,
	patterns,
	math,
	external}
\usepackage{pgfplots}
\usepackage{pgfplotstable}
\usetikzlibrary{pgfplots.statistics, pgfplots.colorbrewer, calc, external} 
\pgfplotsset{compat=1.18}
\definecolor{THICOLOR}{rgb}{0,0.112,0.47}

\definecolor{COFFEE1}{RGB}{60,47,47}

\definecolor{COFFEE2}{RGB}{133,68,66}

\definecolor{COFFEE3}{RGB}{223,189,146}

\definecolor{COFFEE4}{RGB}{111,68,54}

\definecolor{COFFEE5}{RGB}{190,155,123}

\definecolor{BLAU}{RGB}{0,0,255}

\definecolor{YlGn31}{RGB}{247,252,185}

\definecolor{YlGn3C}{RGB}{247,252,185}

\definecolor{YlGn32}{RGB}{173,221,142}

\definecolor{YlGn3F}{RGB}{173,221,142}

\definecolor{YlGn33}{RGB}{49,163,84}

\definecolor{YlGn3I}{RGB}{49,163,84}

\definecolor{YlGn41}{RGB}{255,255,204}

\definecolor{YlGn4B}{RGB}{255,255,204}

\definecolor{YlGn42}{RGB}{194,230,153}

\definecolor{YlGn4E}{RGB}{194,230,153}

\definecolor{YlGn43}{RGB}{120,198,121}

\definecolor{YlGn4G}{RGB}{120,198,121}

\definecolor{YlGn44}{RGB}{35,132,67}

\definecolor{YlGn4J}{RGB}{35,132,67}

\definecolor{YlGn51}{RGB}{255,255,204}

\definecolor{YlGn5B}{RGB}{255,255,204}

\definecolor{YlGn52}{RGB}{194,230,153}

\definecolor{YlGn5E}{RGB}{194,230,153}

\definecolor{YlGn53}{RGB}{120,198,121}

\definecolor{YlGn5G}{RGB}{120,198,121}

\definecolor{YlGn54}{RGB}{49,163,84}

\definecolor{YlGn5I}{RGB}{49,163,84}

\definecolor{YlGn55}{RGB}{0,104,55}

\definecolor{YlGn5K}{RGB}{0,104,55}

\definecolor{YlGn61}{RGB}{255,255,204}

\definecolor{YlGn6B}{RGB}{255,255,204}

\definecolor{YlGn62}{RGB}{217,240,163}

\definecolor{YlGn6D}{RGB}{217,240,163}

\definecolor{YlGn63}{RGB}{173,221,142}

\definecolor{YlGn6F}{RGB}{173,221,142}

\definecolor{YlGn64}{RGB}{120,198,121}

\definecolor{YlGn6G}{RGB}{120,198,121}

\definecolor{YlGn65}{RGB}{49,163,84}

\definecolor{YlGn6I}{RGB}{49,163,84}

\definecolor{YlGn66}{RGB}{0,104,55}

\definecolor{YlGn6K}{RGB}{0,104,55}

\definecolor{YlGn71}{RGB}{255,255,204}

\definecolor{YlGn7B}{RGB}{255,255,204}

\definecolor{YlGn72}{RGB}{217,240,163}

\definecolor{YlGn7D}{RGB}{217,240,163}

\definecolor{YlGn73}{RGB}{173,221,142}

\definecolor{YlGn7F}{RGB}{173,221,142}

\definecolor{YlGn74}{RGB}{120,198,121}

\definecolor{YlGn7G}{RGB}{120,198,121}

\definecolor{YlGn75}{RGB}{65,171,93}

\definecolor{YlGn7H}{RGB}{65,171,93}

\definecolor{YlGn76}{RGB}{35,132,67}

\definecolor{YlGn7J}{RGB}{35,132,67}

\definecolor{YlGn77}{RGB}{0,90,50}

\definecolor{YlGn7L}{RGB}{0,90,50}

\definecolor{YlGn81}{RGB}{255,255,229}

\definecolor{YlGn8A}{RGB}{255,255,229}

\definecolor{YlGn82}{RGB}{247,252,185}

\definecolor{YlGn8C}{RGB}{247,252,185}

\definecolor{YlGn83}{RGB}{217,240,163}

\definecolor{YlGn8D}{RGB}{217,240,163}

\definecolor{YlGn84}{RGB}{173,221,142}

\definecolor{YlGn8F}{RGB}{173,221,142}

\definecolor{YlGn85}{RGB}{120,198,121}

\definecolor{YlGn8G}{RGB}{120,198,121}

\definecolor{YlGn86}{RGB}{65,171,93}

\definecolor{YlGn8H}{RGB}{65,171,93}

\definecolor{YlGn87}{RGB}{35,132,67}

\definecolor{YlGn8J}{RGB}{35,132,67}

\definecolor{YlGn88}{RGB}{0,90,50}

\definecolor{YlGn8L}{RGB}{0,90,50}

\definecolor{YlGn91}{RGB}{255,255,229}

\definecolor{YlGn9A}{RGB}{255,255,229}

\definecolor{YlGn92}{RGB}{247,252,185}

\definecolor{YlGn9C}{RGB}{247,252,185}

\definecolor{YlGn93}{RGB}{217,240,163}

\definecolor{YlGn9D}{RGB}{217,240,163}

\definecolor{YlGn94}{RGB}{173,221,142}

\definecolor{YlGn9F}{RGB}{173,221,142}

\definecolor{YlGn95}{RGB}{120,198,121}

\definecolor{YlGn9G}{RGB}{120,198,121}

\definecolor{YlGn96}{RGB}{65,171,93}

\definecolor{YlGn9H}{RGB}{65,171,93}

\definecolor{YlGn97}{RGB}{35,132,67}

\definecolor{YlGn9J}{RGB}{35,132,67}

\definecolor{YlGn98}{RGB}{0,104,55}

\definecolor{YlGn9K}{RGB}{0,104,55}

\definecolor{YlGn99}{RGB}{0,69,41}

\definecolor{YlGn9M}{RGB}{0,69,41}

\definecolor{YlGnBu31}{RGB}{237,248,177}

\definecolor{YlGnBu3C}{RGB}{237,248,177}

\definecolor{YlGnBu32}{RGB}{127,205,187}

\definecolor{YlGnBu3F}{RGB}{127,205,187}

\definecolor{YlGnBu33}{RGB}{44,127,184}

\definecolor{YlGnBu3I}{RGB}{44,127,184}

\definecolor{YlGnBu41}{RGB}{255,255,204}

\definecolor{YlGnBu4B}{RGB}{255,255,204}

\definecolor{YlGnBu42}{RGB}{161,218,180}

\definecolor{YlGnBu4E}{RGB}{161,218,180}

\definecolor{YlGnBu43}{RGB}{65,182,196}

\definecolor{YlGnBu4G}{RGB}{65,182,196}

\definecolor{YlGnBu44}{RGB}{34,94,168}

\definecolor{YlGnBu4J}{RGB}{34,94,168}

\definecolor{YlGnBu51}{RGB}{255,255,204}

\definecolor{YlGnBu5B}{RGB}{255,255,204}

\definecolor{YlGnBu52}{RGB}{161,218,180}

\definecolor{YlGnBu5E}{RGB}{161,218,180}

\definecolor{YlGnBu53}{RGB}{65,182,196}

\definecolor{YlGnBu5G}{RGB}{65,182,196}

\definecolor{YlGnBu54}{RGB}{44,127,184}

\definecolor{YlGnBu5I}{RGB}{44,127,184}

\definecolor{YlGnBu55}{RGB}{37,52,148}

\definecolor{YlGnBu5K}{RGB}{37,52,148}

\definecolor{YlGnBu61}{RGB}{255,255,204}

\definecolor{YlGnBu6B}{RGB}{255,255,204}

\definecolor{YlGnBu62}{RGB}{199,233,180}

\definecolor{YlGnBu6D}{RGB}{199,233,180}

\definecolor{YlGnBu63}{RGB}{127,205,187}

\definecolor{YlGnBu6F}{RGB}{127,205,187}

\definecolor{YlGnBu64}{RGB}{65,182,196}

\definecolor{YlGnBu6G}{RGB}{65,182,196}

\definecolor{YlGnBu65}{RGB}{44,127,184}

\definecolor{YlGnBu6I}{RGB}{44,127,184}

\definecolor{YlGnBu66}{RGB}{37,52,148}

\definecolor{YlGnBu6K}{RGB}{37,52,148}

\definecolor{YlGnBu71}{RGB}{255,255,204}

\definecolor{YlGnBu7B}{RGB}{255,255,204}

\definecolor{YlGnBu72}{RGB}{199,233,180}

\definecolor{YlGnBu7D}{RGB}{199,233,180}

\definecolor{YlGnBu73}{RGB}{127,205,187}

\definecolor{YlGnBu7F}{RGB}{127,205,187}

\definecolor{YlGnBu74}{RGB}{65,182,196}

\definecolor{YlGnBu7G}{RGB}{65,182,196}

\definecolor{YlGnBu75}{RGB}{29,145,192}

\definecolor{YlGnBu7H}{RGB}{29,145,192}

\definecolor{YlGnBu76}{RGB}{34,94,168}

\definecolor{YlGnBu7J}{RGB}{34,94,168}

\definecolor{YlGnBu77}{RGB}{12,44,132}

\definecolor{YlGnBu7L}{RGB}{12,44,132}

\definecolor{YlGnBu81}{RGB}{255,255,217}

\definecolor{YlGnBu8A}{RGB}{255,255,217}

\definecolor{YlGnBu82}{RGB}{237,248,177}

\definecolor{YlGnBu8C}{RGB}{237,248,177}

\definecolor{YlGnBu83}{RGB}{199,233,180}

\definecolor{YlGnBu8D}{RGB}{199,233,180}

\definecolor{YlGnBu84}{RGB}{127,205,187}

\definecolor{YlGnBu8F}{RGB}{127,205,187}

\definecolor{YlGnBu85}{RGB}{65,182,196}

\definecolor{YlGnBu8G}{RGB}{65,182,196}

\definecolor{YlGnBu86}{RGB}{29,145,192}

\definecolor{YlGnBu8H}{RGB}{29,145,192}

\definecolor{YlGnBu87}{RGB}{34,94,168}

\definecolor{YlGnBu8J}{RGB}{34,94,168}

\definecolor{YlGnBu88}{RGB}{12,44,132}

\definecolor{YlGnBu8L}{RGB}{12,44,132}

\definecolor{YlGnBu91}{RGB}{255,255,217}

\definecolor{YlGnBu9A}{RGB}{255,255,217}

\definecolor{YlGnBu92}{RGB}{237,248,177}

\definecolor{YlGnBu9C}{RGB}{237,248,177}

\definecolor{YlGnBu93}{RGB}{199,233,180}

\definecolor{YlGnBu9D}{RGB}{199,233,180}

\definecolor{YlGnBu94}{RGB}{127,205,187}

\definecolor{YlGnBu9F}{RGB}{127,205,187}

\definecolor{YlGnBu95}{RGB}{65,182,196}

\definecolor{YlGnBu9G}{RGB}{65,182,196}

\definecolor{YlGnBu96}{RGB}{29,145,192}

\definecolor{YlGnBu9H}{RGB}{29,145,192}

\definecolor{YlGnBu97}{RGB}{34,94,168}

\definecolor{YlGnBu9J}{RGB}{34,94,168}

\definecolor{YlGnBu98}{RGB}{37,52,148}

\definecolor{YlGnBu9K}{RGB}{37,52,148}

\definecolor{YlGnBu99}{RGB}{8,29,88}

\definecolor{YlGnBu9M}{RGB}{8,29,88}

\definecolor{GnBu31}{RGB}{224,243,219}

\definecolor{GnBu3C}{RGB}{224,243,219}

\definecolor{GnBu32}{RGB}{168,221,181}

\definecolor{GnBu3F}{RGB}{168,221,181}

\definecolor{GnBu33}{RGB}{67,162,202}

\definecolor{GnBu3I}{RGB}{67,162,202}

\definecolor{GnBu41}{RGB}{240,249,232}

\definecolor{GnBu4B}{RGB}{240,249,232}

\definecolor{GnBu42}{RGB}{186,228,188}

\definecolor{GnBu4E}{RGB}{186,228,188}

\definecolor{GnBu43}{RGB}{123,204,196}

\definecolor{GnBu4G}{RGB}{123,204,196}

\definecolor{GnBu44}{RGB}{43,140,190}

\definecolor{GnBu4J}{RGB}{43,140,190}

\definecolor{GnBu51}{RGB}{240,249,232}

\definecolor{GnBu5B}{RGB}{240,249,232}

\definecolor{GnBu52}{RGB}{186,228,188}

\definecolor{GnBu5E}{RGB}{186,228,188}

\definecolor{GnBu53}{RGB}{123,204,196}

\definecolor{GnBu5G}{RGB}{123,204,196}

\definecolor{GnBu54}{RGB}{67,162,202}

\definecolor{GnBu5I}{RGB}{67,162,202}

\definecolor{GnBu55}{RGB}{8,104,172}

\definecolor{GnBu5K}{RGB}{8,104,172}

\definecolor{GnBu61}{RGB}{240,249,232}

\definecolor{GnBu6B}{RGB}{240,249,232}

\definecolor{GnBu62}{RGB}{204,235,197}

\definecolor{GnBu6D}{RGB}{204,235,197}

\definecolor{GnBu63}{RGB}{168,221,181}

\definecolor{GnBu6F}{RGB}{168,221,181}

\definecolor{GnBu64}{RGB}{123,204,196}

\definecolor{GnBu6G}{RGB}{123,204,196}

\definecolor{GnBu65}{RGB}{67,162,202}

\definecolor{GnBu6I}{RGB}{67,162,202}

\definecolor{GnBu66}{RGB}{8,104,172}

\definecolor{GnBu6K}{RGB}{8,104,172}

\definecolor{GnBu71}{RGB}{240,249,232}

\definecolor{GnBu7B}{RGB}{240,249,232}

\definecolor{GnBu72}{RGB}{204,235,197}

\definecolor{GnBu7D}{RGB}{204,235,197}

\definecolor{GnBu73}{RGB}{168,221,181}

\definecolor{GnBu7F}{RGB}{168,221,181}

\definecolor{GnBu74}{RGB}{123,204,196}

\definecolor{GnBu7G}{RGB}{123,204,196}

\definecolor{GnBu75}{RGB}{78,179,211}

\definecolor{GnBu7H}{RGB}{78,179,211}

\definecolor{GnBu76}{RGB}{43,140,190}

\definecolor{GnBu7J}{RGB}{43,140,190}

\definecolor{GnBu77}{RGB}{8,88,158}

\definecolor{GnBu7L}{RGB}{8,88,158}

\definecolor{GnBu81}{RGB}{247,252,240}

\definecolor{GnBu8A}{RGB}{247,252,240}

\definecolor{GnBu82}{RGB}{224,243,219}

\definecolor{GnBu8C}{RGB}{224,243,219}

\definecolor{GnBu83}{RGB}{204,235,197}

\definecolor{GnBu8D}{RGB}{204,235,197}

\definecolor{GnBu84}{RGB}{168,221,181}

\definecolor{GnBu8F}{RGB}{168,221,181}

\definecolor{GnBu85}{RGB}{123,204,196}

\definecolor{GnBu8G}{RGB}{123,204,196}

\definecolor{GnBu86}{RGB}{78,179,211}

\definecolor{GnBu8H}{RGB}{78,179,211}

\definecolor{GnBu87}{RGB}{43,140,190}

\definecolor{GnBu8J}{RGB}{43,140,190}

\definecolor{GnBu88}{RGB}{8,88,158}

\definecolor{GnBu8L}{RGB}{8,88,158}

\definecolor{GnBu91}{RGB}{247,252,240}

\definecolor{GnBu9A}{RGB}{247,252,240}

\definecolor{GnBu92}{RGB}{224,243,219}

\definecolor{GnBu9C}{RGB}{224,243,219}

\definecolor{GnBu93}{RGB}{204,235,197}

\definecolor{GnBu9D}{RGB}{204,235,197}

\definecolor{GnBu94}{RGB}{168,221,181}

\definecolor{GnBu9F}{RGB}{168,221,181}

\definecolor{GnBu95}{RGB}{123,204,196}

\definecolor{GnBu9G}{RGB}{123,204,196}

\definecolor{GnBu96}{RGB}{78,179,211}

\definecolor{GnBu9H}{RGB}{78,179,211}

\definecolor{GnBu97}{RGB}{43,140,190}

\definecolor{GnBu9J}{RGB}{43,140,190}

\definecolor{GnBu98}{RGB}{8,104,172}

\definecolor{GnBu9K}{RGB}{8,104,172}

\definecolor{GnBu99}{RGB}{8,64,129}

\definecolor{GnBu9M}{RGB}{8,64,129}

\definecolor{BuGn31}{RGB}{229,245,249}

\definecolor{BuGn3C}{RGB}{229,245,249}

\definecolor{BuGn32}{RGB}{153,216,201}

\definecolor{BuGn3F}{RGB}{153,216,201}

\definecolor{BuGn33}{RGB}{44,162,95}

\definecolor{BuGn3I}{RGB}{44,162,95}

\definecolor{BuGn41}{RGB}{237,248,251}

\definecolor{BuGn4B}{RGB}{237,248,251}

\definecolor{BuGn42}{RGB}{178,226,226}

\definecolor{BuGn4E}{RGB}{178,226,226}

\definecolor{BuGn43}{RGB}{102,194,164}

\definecolor{BuGn4G}{RGB}{102,194,164}

\definecolor{BuGn44}{RGB}{35,139,69}

\definecolor{BuGn4J}{RGB}{35,139,69}

\definecolor{BuGn51}{RGB}{237,248,251}

\definecolor{BuGn5B}{RGB}{237,248,251}

\definecolor{BuGn52}{RGB}{178,226,226}

\definecolor{BuGn5E}{RGB}{178,226,226}

\definecolor{BuGn53}{RGB}{102,194,164}

\definecolor{BuGn5G}{RGB}{102,194,164}

\definecolor{BuGn54}{RGB}{44,162,95}

\definecolor{BuGn5I}{RGB}{44,162,95}

\definecolor{BuGn55}{RGB}{0,109,44}

\definecolor{BuGn5K}{RGB}{0,109,44}

\definecolor{BuGn61}{RGB}{237,248,251}

\definecolor{BuGn6B}{RGB}{237,248,251}

\definecolor{BuGn62}{RGB}{204,236,230}

\definecolor{BuGn6D}{RGB}{204,236,230}

\definecolor{BuGn63}{RGB}{153,216,201}

\definecolor{BuGn6F}{RGB}{153,216,201}

\definecolor{BuGn64}{RGB}{102,194,164}

\definecolor{BuGn6G}{RGB}{102,194,164}

\definecolor{BuGn65}{RGB}{44,162,95}

\definecolor{BuGn6I}{RGB}{44,162,95}

\definecolor{BuGn66}{RGB}{0,109,44}

\definecolor{BuGn6K}{RGB}{0,109,44}

\definecolor{BuGn71}{RGB}{237,248,251}

\definecolor{BuGn7B}{RGB}{237,248,251}

\definecolor{BuGn72}{RGB}{204,236,230}

\definecolor{BuGn7D}{RGB}{204,236,230}

\definecolor{BuGn73}{RGB}{153,216,201}

\definecolor{BuGn7F}{RGB}{153,216,201}

\definecolor{BuGn74}{RGB}{102,194,164}

\definecolor{BuGn7G}{RGB}{102,194,164}

\definecolor{BuGn75}{RGB}{65,174,118}

\definecolor{BuGn7H}{RGB}{65,174,118}

\definecolor{BuGn76}{RGB}{35,139,69}

\definecolor{BuGn7J}{RGB}{35,139,69}

\definecolor{BuGn77}{RGB}{0,88,36}

\definecolor{BuGn7L}{RGB}{0,88,36}

\definecolor{BuGn81}{RGB}{247,252,253}

\definecolor{BuGn8A}{RGB}{247,252,253}

\definecolor{BuGn82}{RGB}{229,245,249}

\definecolor{BuGn8C}{RGB}{229,245,249}

\definecolor{BuGn83}{RGB}{204,236,230}

\definecolor{BuGn8D}{RGB}{204,236,230}

\definecolor{BuGn84}{RGB}{153,216,201}

\definecolor{BuGn8F}{RGB}{153,216,201}

\definecolor{BuGn85}{RGB}{102,194,164}

\definecolor{BuGn8G}{RGB}{102,194,164}

\definecolor{BuGn86}{RGB}{65,174,118}

\definecolor{BuGn8H}{RGB}{65,174,118}

\definecolor{BuGn87}{RGB}{35,139,69}

\definecolor{BuGn8J}{RGB}{35,139,69}

\definecolor{BuGn88}{RGB}{0,88,36}

\definecolor{BuGn8L}{RGB}{0,88,36}

\definecolor{BuGn91}{RGB}{247,252,253}

\definecolor{BuGn9A}{RGB}{247,252,253}

\definecolor{BuGn92}{RGB}{229,245,249}

\definecolor{BuGn9C}{RGB}{229,245,249}

\definecolor{BuGn93}{RGB}{204,236,230}

\definecolor{BuGn9D}{RGB}{204,236,230}

\definecolor{BuGn94}{RGB}{153,216,201}

\definecolor{BuGn9F}{RGB}{153,216,201}

\definecolor{BuGn95}{RGB}{102,194,164}

\definecolor{BuGn9G}{RGB}{102,194,164}

\definecolor{BuGn96}{RGB}{65,174,118}

\definecolor{BuGn9H}{RGB}{65,174,118}

\definecolor{BuGn97}{RGB}{35,139,69}

\definecolor{BuGn9J}{RGB}{35,139,69}

\definecolor{BuGn98}{RGB}{0,109,44}

\definecolor{BuGn9K}{RGB}{0,109,44}

\definecolor{BuGn99}{RGB}{0,68,27}

\definecolor{BuGn9M}{RGB}{0,68,27}

\definecolor{PuBuGn31}{RGB}{236,226,240}

\definecolor{PuBuGn3C}{RGB}{236,226,240}

\definecolor{PuBuGn32}{RGB}{166,189,219}

\definecolor{PuBuGn3F}{RGB}{166,189,219}

\definecolor{PuBuGn33}{RGB}{28,144,153}

\definecolor{PuBuGn3I}{RGB}{28,144,153}

\definecolor{PuBuGn41}{RGB}{246,239,247}

\definecolor{PuBuGn4B}{RGB}{246,239,247}

\definecolor{PuBuGn42}{RGB}{189,201,225}

\definecolor{PuBuGn4E}{RGB}{189,201,225}

\definecolor{PuBuGn43}{RGB}{103,169,207}

\definecolor{PuBuGn4G}{RGB}{103,169,207}

\definecolor{PuBuGn44}{RGB}{2,129,138}

\definecolor{PuBuGn4J}{RGB}{2,129,138}

\definecolor{PuBuGn51}{RGB}{246,239,247}

\definecolor{PuBuGn5B}{RGB}{246,239,247}

\definecolor{PuBuGn52}{RGB}{189,201,225}

\definecolor{PuBuGn5E}{RGB}{189,201,225}

\definecolor{PuBuGn53}{RGB}{103,169,207}

\definecolor{PuBuGn5G}{RGB}{103,169,207}

\definecolor{PuBuGn54}{RGB}{28,144,153}

\definecolor{PuBuGn5I}{RGB}{28,144,153}

\definecolor{PuBuGn55}{RGB}{1,108,89}

\definecolor{PuBuGn5K}{RGB}{1,108,89}

\definecolor{PuBuGn61}{RGB}{246,239,247}

\definecolor{PuBuGn6B}{RGB}{246,239,247}

\definecolor{PuBuGn62}{RGB}{208,209,230}

\definecolor{PuBuGn6D}{RGB}{208,209,230}

\definecolor{PuBuGn63}{RGB}{166,189,219}

\definecolor{PuBuGn6F}{RGB}{166,189,219}

\definecolor{PuBuGn64}{RGB}{103,169,207}

\definecolor{PuBuGn6G}{RGB}{103,169,207}

\definecolor{PuBuGn65}{RGB}{28,144,153}

\definecolor{PuBuGn6I}{RGB}{28,144,153}

\definecolor{PuBuGn66}{RGB}{1,108,89}

\definecolor{PuBuGn6K}{RGB}{1,108,89}

\definecolor{PuBuGn71}{RGB}{246,239,247}

\definecolor{PuBuGn7B}{RGB}{246,239,247}

\definecolor{PuBuGn72}{RGB}{208,209,230}

\definecolor{PuBuGn7D}{RGB}{208,209,230}

\definecolor{PuBuGn73}{RGB}{166,189,219}

\definecolor{PuBuGn7F}{RGB}{166,189,219}

\definecolor{PuBuGn74}{RGB}{103,169,207}

\definecolor{PuBuGn7G}{RGB}{103,169,207}

\definecolor{PuBuGn75}{RGB}{54,144,192}

\definecolor{PuBuGn7H}{RGB}{54,144,192}

\definecolor{PuBuGn76}{RGB}{2,129,138}

\definecolor{PuBuGn7J}{RGB}{2,129,138}

\definecolor{PuBuGn77}{RGB}{1,100,80}

\definecolor{PuBuGn7L}{RGB}{1,100,80}

\definecolor{PuBuGn81}{RGB}{255,247,251}

\definecolor{PuBuGn8A}{RGB}{255,247,251}

\definecolor{PuBuGn82}{RGB}{236,226,240}

\definecolor{PuBuGn8C}{RGB}{236,226,240}

\definecolor{PuBuGn83}{RGB}{208,209,230}

\definecolor{PuBuGn8D}{RGB}{208,209,230}

\definecolor{PuBuGn84}{RGB}{166,189,219}

\definecolor{PuBuGn8F}{RGB}{166,189,219}

\definecolor{PuBuGn85}{RGB}{103,169,207}

\definecolor{PuBuGn8G}{RGB}{103,169,207}

\definecolor{PuBuGn86}{RGB}{54,144,192}

\definecolor{PuBuGn8H}{RGB}{54,144,192}

\definecolor{PuBuGn87}{RGB}{2,129,138}

\definecolor{PuBuGn8J}{RGB}{2,129,138}

\definecolor{PuBuGn88}{RGB}{1,100,80}

\definecolor{PuBuGn8L}{RGB}{1,100,80}

\definecolor{PuBuGn91}{RGB}{255,247,251}

\definecolor{PuBuGn9A}{RGB}{255,247,251}

\definecolor{PuBuGn92}{RGB}{236,226,240}

\definecolor{PuBuGn9C}{RGB}{236,226,240}

\definecolor{PuBuGn93}{RGB}{208,209,230}

\definecolor{PuBuGn9D}{RGB}{208,209,230}

\definecolor{PuBuGn94}{RGB}{166,189,219}

\definecolor{PuBuGn9F}{RGB}{166,189,219}

\definecolor{PuBuGn95}{RGB}{103,169,207}

\definecolor{PuBuGn9G}{RGB}{103,169,207}

\definecolor{PuBuGn96}{RGB}{54,144,192}

\definecolor{PuBuGn9H}{RGB}{54,144,192}

\definecolor{PuBuGn97}{RGB}{2,129,138}

\definecolor{PuBuGn9J}{RGB}{2,129,138}

\definecolor{PuBuGn98}{RGB}{1,108,89}

\definecolor{PuBuGn9K}{RGB}{1,108,89}

\definecolor{PuBuGn99}{RGB}{1,70,54}

\definecolor{PuBuGn9M}{RGB}{1,70,54}

\definecolor{PuBu31}{RGB}{236,231,242}

\definecolor{PuBu3C}{RGB}{236,231,242}

\definecolor{PuBu32}{RGB}{166,189,219}

\definecolor{PuBu3F}{RGB}{166,189,219}

\definecolor{PuBu33}{RGB}{43,140,190}

\definecolor{PuBu3I}{RGB}{43,140,190}

\definecolor{PuBu41}{RGB}{241,238,246}

\definecolor{PuBu4B}{RGB}{241,238,246}

\definecolor{PuBu42}{RGB}{189,201,225}

\definecolor{PuBu4E}{RGB}{189,201,225}

\definecolor{PuBu43}{RGB}{116,169,207}

\definecolor{PuBu4G}{RGB}{116,169,207}

\definecolor{PuBu44}{RGB}{5,112,176}

\definecolor{PuBu4J}{RGB}{5,112,176}

\definecolor{PuBu51}{RGB}{241,238,246}

\definecolor{PuBu5B}{RGB}{241,238,246}

\definecolor{PuBu52}{RGB}{189,201,225}

\definecolor{PuBu5E}{RGB}{189,201,225}

\definecolor{PuBu53}{RGB}{116,169,207}

\definecolor{PuBu5G}{RGB}{116,169,207}

\definecolor{PuBu54}{RGB}{43,140,190}

\definecolor{PuBu5I}{RGB}{43,140,190}

\definecolor{PuBu55}{RGB}{4,90,141}

\definecolor{PuBu5K}{RGB}{4,90,141}

\definecolor{PuBu61}{RGB}{241,238,246}

\definecolor{PuBu6B}{RGB}{241,238,246}

\definecolor{PuBu62}{RGB}{208,209,230}

\definecolor{PuBu6D}{RGB}{208,209,230}

\definecolor{PuBu63}{RGB}{166,189,219}

\definecolor{PuBu6F}{RGB}{166,189,219}

\definecolor{PuBu64}{RGB}{116,169,207}

\definecolor{PuBu6G}{RGB}{116,169,207}

\definecolor{PuBu65}{RGB}{43,140,190}

\definecolor{PuBu6I}{RGB}{43,140,190}

\definecolor{PuBu66}{RGB}{4,90,141}

\definecolor{PuBu6K}{RGB}{4,90,141}

\definecolor{PuBu71}{RGB}{241,238,246}

\definecolor{PuBu7B}{RGB}{241,238,246}

\definecolor{PuBu72}{RGB}{208,209,230}

\definecolor{PuBu7D}{RGB}{208,209,230}

\definecolor{PuBu73}{RGB}{166,189,219}

\definecolor{PuBu7F}{RGB}{166,189,219}

\definecolor{PuBu74}{RGB}{116,169,207}

\definecolor{PuBu7G}{RGB}{116,169,207}

\definecolor{PuBu75}{RGB}{54,144,192}

\definecolor{PuBu7H}{RGB}{54,144,192}

\definecolor{PuBu76}{RGB}{5,112,176}

\definecolor{PuBu7J}{RGB}{5,112,176}

\definecolor{PuBu77}{RGB}{3,78,123}

\definecolor{PuBu7L}{RGB}{3,78,123}

\definecolor{PuBu81}{RGB}{255,247,251}

\definecolor{PuBu8A}{RGB}{255,247,251}

\definecolor{PuBu82}{RGB}{236,231,242}

\definecolor{PuBu8C}{RGB}{236,231,242}

\definecolor{PuBu83}{RGB}{208,209,230}

\definecolor{PuBu8D}{RGB}{208,209,230}

\definecolor{PuBu84}{RGB}{166,189,219}

\definecolor{PuBu8F}{RGB}{166,189,219}

\definecolor{PuBu85}{RGB}{116,169,207}

\definecolor{PuBu8G}{RGB}{116,169,207}

\definecolor{PuBu86}{RGB}{54,144,192}

\definecolor{PuBu8H}{RGB}{54,144,192}

\definecolor{PuBu87}{RGB}{5,112,176}

\definecolor{PuBu8J}{RGB}{5,112,176}

\definecolor{PuBu88}{RGB}{3,78,123}

\definecolor{PuBu8L}{RGB}{3,78,123}

\definecolor{PuBu91}{RGB}{255,247,251}

\definecolor{PuBu9A}{RGB}{255,247,251}

\definecolor{PuBu92}{RGB}{236,231,242}

\definecolor{PuBu9C}{RGB}{236,231,242}

\definecolor{PuBu93}{RGB}{208,209,230}

\definecolor{PuBu9D}{RGB}{208,209,230}

\definecolor{PuBu94}{RGB}{166,189,219}

\definecolor{PuBu9F}{RGB}{166,189,219}

\definecolor{PuBu95}{RGB}{116,169,207}

\definecolor{PuBu9G}{RGB}{116,169,207}

\definecolor{PuBu96}{RGB}{54,144,192}

\definecolor{PuBu9H}{RGB}{54,144,192}

\definecolor{PuBu97}{RGB}{5,112,176}

\definecolor{PuBu9J}{RGB}{5,112,176}

\definecolor{PuBu98}{RGB}{4,90,141}

\definecolor{PuBu9K}{RGB}{4,90,141}

\definecolor{PuBu99}{RGB}{2,56,88}

\definecolor{PuBu9M}{RGB}{2,56,88}

\definecolor{BuPu31}{RGB}{224,236,244}

\definecolor{BuPu3C}{RGB}{224,236,244}

\definecolor{BuPu32}{RGB}{158,188,218}

\definecolor{BuPu3F}{RGB}{158,188,218}

\definecolor{BuPu33}{RGB}{136,86,167}

\definecolor{BuPu3I}{RGB}{136,86,167}

\definecolor{BuPu41}{RGB}{237,248,251}

\definecolor{BuPu4B}{RGB}{237,248,251}

\definecolor{BuPu42}{RGB}{179,205,227}

\definecolor{BuPu4E}{RGB}{179,205,227}

\definecolor{BuPu43}{RGB}{140,150,198}

\definecolor{BuPu4G}{RGB}{140,150,198}

\definecolor{BuPu44}{RGB}{136,65,157}

\definecolor{BuPu4J}{RGB}{136,65,157}

\definecolor{BuPu51}{RGB}{237,248,251}

\definecolor{BuPu5B}{RGB}{237,248,251}

\definecolor{BuPu52}{RGB}{179,205,227}

\definecolor{BuPu5E}{RGB}{179,205,227}

\definecolor{BuPu53}{RGB}{140,150,198}

\definecolor{BuPu5G}{RGB}{140,150,198}

\definecolor{BuPu54}{RGB}{136,86,167}

\definecolor{BuPu5I}{RGB}{136,86,167}

\definecolor{BuPu55}{RGB}{129,15,124}

\definecolor{BuPu5K}{RGB}{129,15,124}

\definecolor{BuPu61}{RGB}{237,248,251}

\definecolor{BuPu6B}{RGB}{237,248,251}

\definecolor{BuPu62}{RGB}{191,211,230}

\definecolor{BuPu6D}{RGB}{191,211,230}

\definecolor{BuPu63}{RGB}{158,188,218}

\definecolor{BuPu6F}{RGB}{158,188,218}

\definecolor{BuPu64}{RGB}{140,150,198}

\definecolor{BuPu6G}{RGB}{140,150,198}

\definecolor{BuPu65}{RGB}{136,86,167}

\definecolor{BuPu6I}{RGB}{136,86,167}

\definecolor{BuPu66}{RGB}{129,15,124}

\definecolor{BuPu6K}{RGB}{129,15,124}

\definecolor{BuPu71}{RGB}{237,248,251}

\definecolor{BuPu7B}{RGB}{237,248,251}

\definecolor{BuPu72}{RGB}{191,211,230}

\definecolor{BuPu7D}{RGB}{191,211,230}

\definecolor{BuPu73}{RGB}{158,188,218}

\definecolor{BuPu7F}{RGB}{158,188,218}

\definecolor{BuPu74}{RGB}{140,150,198}

\definecolor{BuPu7G}{RGB}{140,150,198}

\definecolor{BuPu75}{RGB}{140,107,177}

\definecolor{BuPu7H}{RGB}{140,107,177}

\definecolor{BuPu76}{RGB}{136,65,157}

\definecolor{BuPu7J}{RGB}{136,65,157}

\definecolor{BuPu77}{RGB}{110,1,107}

\definecolor{BuPu7L}{RGB}{110,1,107}

\definecolor{BuPu81}{RGB}{247,252,253}

\definecolor{BuPu8A}{RGB}{247,252,253}

\definecolor{BuPu82}{RGB}{224,236,244}

\definecolor{BuPu8C}{RGB}{224,236,244}

\definecolor{BuPu83}{RGB}{191,211,230}

\definecolor{BuPu8D}{RGB}{191,211,230}

\definecolor{BuPu84}{RGB}{158,188,218}

\definecolor{BuPu8F}{RGB}{158,188,218}

\definecolor{BuPu85}{RGB}{140,150,198}

\definecolor{BuPu8G}{RGB}{140,150,198}

\definecolor{BuPu86}{RGB}{140,107,177}

\definecolor{BuPu8H}{RGB}{140,107,177}

\definecolor{BuPu87}{RGB}{136,65,157}

\definecolor{BuPu8J}{RGB}{136,65,157}

\definecolor{BuPu88}{RGB}{110,1,107}

\definecolor{BuPu8L}{RGB}{110,1,107}

\definecolor{BuPu91}{RGB}{247,252,253}

\definecolor{BuPu9A}{RGB}{247,252,253}

\definecolor{BuPu92}{RGB}{224,236,244}

\definecolor{BuPu9C}{RGB}{224,236,244}

\definecolor{BuPu93}{RGB}{191,211,230}

\definecolor{BuPu9D}{RGB}{191,211,230}

\definecolor{BuPu94}{RGB}{158,188,218}

\definecolor{BuPu9F}{RGB}{158,188,218}

\definecolor{BuPu95}{RGB}{140,150,198}

\definecolor{BuPu9G}{RGB}{140,150,198}

\definecolor{BuPu96}{RGB}{140,107,177}

\definecolor{BuPu9H}{RGB}{140,107,177}

\definecolor{BuPu97}{RGB}{136,65,157}

\definecolor{BuPu9J}{RGB}{136,65,157}

\definecolor{BuPu98}{RGB}{129,15,124}

\definecolor{BuPu9K}{RGB}{129,15,124}

\definecolor{BuPu99}{RGB}{77,0,75}

\definecolor{BuPu9M}{RGB}{77,0,75}

\definecolor{RdPu31}{RGB}{253,224,221}

\definecolor{RdPu3C}{RGB}{253,224,221}

\definecolor{RdPu32}{RGB}{250,159,181}

\definecolor{RdPu3F}{RGB}{250,159,181}

\definecolor{RdPu33}{RGB}{197,27,138}

\definecolor{RdPu3I}{RGB}{197,27,138}

\definecolor{RdPu41}{RGB}{254,235,226}

\definecolor{RdPu4B}{RGB}{254,235,226}

\definecolor{RdPu42}{RGB}{251,180,185}

\definecolor{RdPu4E}{RGB}{251,180,185}

\definecolor{RdPu43}{RGB}{247,104,161}

\definecolor{RdPu4G}{RGB}{247,104,161}

\definecolor{RdPu44}{RGB}{174,1,126}

\definecolor{RdPu4J}{RGB}{174,1,126}

\definecolor{RdPu51}{RGB}{254,235,226}

\definecolor{RdPu5B}{RGB}{254,235,226}

\definecolor{RdPu52}{RGB}{251,180,185}

\definecolor{RdPu5E}{RGB}{251,180,185}

\definecolor{RdPu53}{RGB}{247,104,161}

\definecolor{RdPu5G}{RGB}{247,104,161}

\definecolor{RdPu54}{RGB}{197,27,138}

\definecolor{RdPu5I}{RGB}{197,27,138}

\definecolor{RdPu55}{RGB}{122,1,119}

\definecolor{RdPu5K}{RGB}{122,1,119}

\definecolor{RdPu61}{RGB}{254,235,226}

\definecolor{RdPu6B}{RGB}{254,235,226}

\definecolor{RdPu62}{RGB}{252,197,192}

\definecolor{RdPu6D}{RGB}{252,197,192}

\definecolor{RdPu63}{RGB}{250,159,181}

\definecolor{RdPu6F}{RGB}{250,159,181}

\definecolor{RdPu64}{RGB}{247,104,161}

\definecolor{RdPu6G}{RGB}{247,104,161}

\definecolor{RdPu65}{RGB}{197,27,138}

\definecolor{RdPu6I}{RGB}{197,27,138}

\definecolor{RdPu66}{RGB}{122,1,119}

\definecolor{RdPu6K}{RGB}{122,1,119}

\definecolor{RdPu71}{RGB}{254,235,226}

\definecolor{RdPu7B}{RGB}{254,235,226}

\definecolor{RdPu72}{RGB}{252,197,192}

\definecolor{RdPu7D}{RGB}{252,197,192}

\definecolor{RdPu73}{RGB}{250,159,181}

\definecolor{RdPu7F}{RGB}{250,159,181}

\definecolor{RdPu74}{RGB}{247,104,161}

\definecolor{RdPu7G}{RGB}{247,104,161}

\definecolor{RdPu75}{RGB}{221,52,151}

\definecolor{RdPu7H}{RGB}{221,52,151}

\definecolor{RdPu76}{RGB}{174,1,126}

\definecolor{RdPu7J}{RGB}{174,1,126}

\definecolor{RdPu77}{RGB}{122,1,119}

\definecolor{RdPu7L}{RGB}{122,1,119}

\definecolor{RdPu81}{RGB}{255,247,243}

\definecolor{RdPu8A}{RGB}{255,247,243}

\definecolor{RdPu82}{RGB}{253,224,221}

\definecolor{RdPu8C}{RGB}{253,224,221}

\definecolor{RdPu83}{RGB}{252,197,192}

\definecolor{RdPu8D}{RGB}{252,197,192}

\definecolor{RdPu84}{RGB}{250,159,181}

\definecolor{RdPu8F}{RGB}{250,159,181}

\definecolor{RdPu85}{RGB}{247,104,161}

\definecolor{RdPu8G}{RGB}{247,104,161}

\definecolor{RdPu86}{RGB}{221,52,151}

\definecolor{RdPu8H}{RGB}{221,52,151}

\definecolor{RdPu87}{RGB}{174,1,126}

\definecolor{RdPu8J}{RGB}{174,1,126}

\definecolor{RdPu88}{RGB}{122,1,119}

\definecolor{RdPu8L}{RGB}{122,1,119}

\definecolor{RdPu91}{RGB}{255,247,243}

\definecolor{RdPu9A}{RGB}{255,247,243}

\definecolor{RdPu92}{RGB}{253,224,221}

\definecolor{RdPu9C}{RGB}{253,224,221}

\definecolor{RdPu93}{RGB}{252,197,192}

\definecolor{RdPu9D}{RGB}{252,197,192}

\definecolor{RdPu94}{RGB}{250,159,181}

\definecolor{RdPu9F}{RGB}{250,159,181}

\definecolor{RdPu95}{RGB}{247,104,161}

\definecolor{RdPu9G}{RGB}{247,104,161}

\definecolor{RdPu96}{RGB}{221,52,151}

\definecolor{RdPu9H}{RGB}{221,52,151}

\definecolor{RdPu97}{RGB}{174,1,126}

\definecolor{RdPu9J}{RGB}{174,1,126}

\definecolor{RdPu98}{RGB}{122,1,119}

\definecolor{RdPu9K}{RGB}{122,1,119}

\definecolor{RdPu99}{RGB}{73,0,106}

\definecolor{RdPu9M}{RGB}{73,0,106}

\definecolor{PuRd31}{RGB}{231,225,239}

\definecolor{PuRd3C}{RGB}{231,225,239}

\definecolor{PuRd32}{RGB}{201,148,199}

\definecolor{PuRd3F}{RGB}{201,148,199}

\definecolor{PuRd33}{RGB}{221,28,119}

\definecolor{PuRd3I}{RGB}{221,28,119}

\definecolor{PuRd41}{RGB}{241,238,246}

\definecolor{PuRd4B}{RGB}{241,238,246}

\definecolor{PuRd42}{RGB}{215,181,216}

\definecolor{PuRd4E}{RGB}{215,181,216}

\definecolor{PuRd43}{RGB}{223,101,176}

\definecolor{PuRd4G}{RGB}{223,101,176}

\definecolor{PuRd44}{RGB}{206,18,86}

\definecolor{PuRd4J}{RGB}{206,18,86}

\definecolor{PuRd51}{RGB}{241,238,246}

\definecolor{PuRd5B}{RGB}{241,238,246}

\definecolor{PuRd52}{RGB}{215,181,216}

\definecolor{PuRd5E}{RGB}{215,181,216}

\definecolor{PuRd53}{RGB}{223,101,176}

\definecolor{PuRd5G}{RGB}{223,101,176}

\definecolor{PuRd54}{RGB}{221,28,119}

\definecolor{PuRd5I}{RGB}{221,28,119}

\definecolor{PuRd55}{RGB}{152,0,67}

\definecolor{PuRd5K}{RGB}{152,0,67}

\definecolor{PuRd61}{RGB}{241,238,246}

\definecolor{PuRd6B}{RGB}{241,238,246}

\definecolor{PuRd62}{RGB}{212,185,218}

\definecolor{PuRd6D}{RGB}{212,185,218}

\definecolor{PuRd63}{RGB}{201,148,199}

\definecolor{PuRd6F}{RGB}{201,148,199}

\definecolor{PuRd64}{RGB}{223,101,176}

\definecolor{PuRd6G}{RGB}{223,101,176}

\definecolor{PuRd65}{RGB}{221,28,119}

\definecolor{PuRd6I}{RGB}{221,28,119}

\definecolor{PuRd66}{RGB}{152,0,67}

\definecolor{PuRd6K}{RGB}{152,0,67}

\definecolor{PuRd71}{RGB}{241,238,246}

\definecolor{PuRd7B}{RGB}{241,238,246}

\definecolor{PuRd72}{RGB}{212,185,218}

\definecolor{PuRd7D}{RGB}{212,185,218}

\definecolor{PuRd73}{RGB}{201,148,199}

\definecolor{PuRd7F}{RGB}{201,148,199}

\definecolor{PuRd74}{RGB}{223,101,176}

\definecolor{PuRd7G}{RGB}{223,101,176}

\definecolor{PuRd75}{RGB}{231,41,138}

\definecolor{PuRd7H}{RGB}{231,41,138}

\definecolor{PuRd76}{RGB}{206,18,86}

\definecolor{PuRd7J}{RGB}{206,18,86}

\definecolor{PuRd77}{RGB}{145,0,63}

\definecolor{PuRd7L}{RGB}{145,0,63}

\definecolor{PuRd81}{RGB}{247,244,249}

\definecolor{PuRd8A}{RGB}{247,244,249}

\definecolor{PuRd82}{RGB}{231,225,239}

\definecolor{PuRd8C}{RGB}{231,225,239}

\definecolor{PuRd83}{RGB}{212,185,218}

\definecolor{PuRd8D}{RGB}{212,185,218}

\definecolor{PuRd84}{RGB}{201,148,199}

\definecolor{PuRd8F}{RGB}{201,148,199}

\definecolor{PuRd85}{RGB}{223,101,176}

\definecolor{PuRd8G}{RGB}{223,101,176}

\definecolor{PuRd86}{RGB}{231,41,138}

\definecolor{PuRd8H}{RGB}{231,41,138}

\definecolor{PuRd87}{RGB}{206,18,86}

\definecolor{PuRd8J}{RGB}{206,18,86}

\definecolor{PuRd88}{RGB}{145,0,63}

\definecolor{PuRd8L}{RGB}{145,0,63}

\definecolor{PuRd91}{RGB}{247,244,249}

\definecolor{PuRd9A}{RGB}{247,244,249}

\definecolor{PuRd92}{RGB}{231,225,239}

\definecolor{PuRd9C}{RGB}{231,225,239}

\definecolor{PuRd93}{RGB}{212,185,218}

\definecolor{PuRd9D}{RGB}{212,185,218}

\definecolor{PuRd94}{RGB}{201,148,199}

\definecolor{PuRd9F}{RGB}{201,148,199}

\definecolor{PuRd95}{RGB}{223,101,176}

\definecolor{PuRd9G}{RGB}{223,101,176}

\definecolor{PuRd96}{RGB}{231,41,138}

\definecolor{PuRd9H}{RGB}{231,41,138}

\definecolor{PuRd97}{RGB}{206,18,86}

\definecolor{PuRd9J}{RGB}{206,18,86}

\definecolor{PuRd98}{RGB}{152,0,67}

\definecolor{PuRd9K}{RGB}{152,0,67}

\definecolor{PuRd99}{RGB}{103,0,31}

\definecolor{PuRd9M}{RGB}{103,0,31}

\definecolor{OrRd31}{RGB}{254,232,200}

\definecolor{OrRd3C}{RGB}{254,232,200}

\definecolor{OrRd32}{RGB}{253,187,132}

\definecolor{OrRd3F}{RGB}{253,187,132}

\definecolor{OrRd33}{RGB}{227,74,51}

\definecolor{OrRd3I}{RGB}{227,74,51}

\definecolor{OrRd41}{RGB}{254,240,217}

\definecolor{OrRd4B}{RGB}{254,240,217}

\definecolor{OrRd42}{RGB}{253,204,138}

\definecolor{OrRd4E}{RGB}{253,204,138}

\definecolor{OrRd43}{RGB}{252,141,89}

\definecolor{OrRd4G}{RGB}{252,141,89}

\definecolor{OrRd44}{RGB}{215,48,31}

\definecolor{OrRd4J}{RGB}{215,48,31}

\definecolor{OrRd51}{RGB}{254,240,217}

\definecolor{OrRd5B}{RGB}{254,240,217}

\definecolor{OrRd52}{RGB}{253,204,138}

\definecolor{OrRd5E}{RGB}{253,204,138}

\definecolor{OrRd53}{RGB}{252,141,89}

\definecolor{OrRd5G}{RGB}{252,141,89}

\definecolor{OrRd54}{RGB}{227,74,51}

\definecolor{OrRd5I}{RGB}{227,74,51}

\definecolor{OrRd55}{RGB}{179,0,0}

\definecolor{OrRd5K}{RGB}{179,0,0}

\definecolor{OrRd61}{RGB}{254,240,217}

\definecolor{OrRd6B}{RGB}{254,240,217}

\definecolor{OrRd62}{RGB}{253,212,158}

\definecolor{OrRd6D}{RGB}{253,212,158}

\definecolor{OrRd63}{RGB}{253,187,132}

\definecolor{OrRd6F}{RGB}{253,187,132}

\definecolor{OrRd64}{RGB}{252,141,89}

\definecolor{OrRd6G}{RGB}{252,141,89}

\definecolor{OrRd65}{RGB}{227,74,51}

\definecolor{OrRd6I}{RGB}{227,74,51}

\definecolor{OrRd66}{RGB}{179,0,0}

\definecolor{OrRd6K}{RGB}{179,0,0}

\definecolor{OrRd71}{RGB}{254,240,217}

\definecolor{OrRd7B}{RGB}{254,240,217}

\definecolor{OrRd72}{RGB}{253,212,158}

\definecolor{OrRd7D}{RGB}{253,212,158}

\definecolor{OrRd73}{RGB}{253,187,132}

\definecolor{OrRd7F}{RGB}{253,187,132}

\definecolor{OrRd74}{RGB}{252,141,89}

\definecolor{OrRd7G}{RGB}{252,141,89}

\definecolor{OrRd75}{RGB}{239,101,72}

\definecolor{OrRd7H}{RGB}{239,101,72}

\definecolor{OrRd76}{RGB}{215,48,31}

\definecolor{OrRd7J}{RGB}{215,48,31}

\definecolor{OrRd77}{RGB}{153,0,0}

\definecolor{OrRd7L}{RGB}{153,0,0}

\definecolor{OrRd81}{RGB}{255,247,236}

\definecolor{OrRd8A}{RGB}{255,247,236}

\definecolor{OrRd82}{RGB}{254,232,200}

\definecolor{OrRd8C}{RGB}{254,232,200}

\definecolor{OrRd83}{RGB}{253,212,158}

\definecolor{OrRd8D}{RGB}{253,212,158}

\definecolor{OrRd84}{RGB}{253,187,132}

\definecolor{OrRd8F}{RGB}{253,187,132}

\definecolor{OrRd85}{RGB}{252,141,89}

\definecolor{OrRd8G}{RGB}{252,141,89}

\definecolor{OrRd86}{RGB}{239,101,72}

\definecolor{OrRd8H}{RGB}{239,101,72}

\definecolor{OrRd87}{RGB}{215,48,31}

\definecolor{OrRd8J}{RGB}{215,48,31}

\definecolor{OrRd88}{RGB}{153,0,0}

\definecolor{OrRd8L}{RGB}{153,0,0}

\definecolor{OrRd91}{RGB}{255,247,236}

\definecolor{OrRd9A}{RGB}{255,247,236}

\definecolor{OrRd92}{RGB}{254,232,200}

\definecolor{OrRd9C}{RGB}{254,232,200}

\definecolor{OrRd93}{RGB}{253,212,158}

\definecolor{OrRd9D}{RGB}{253,212,158}

\definecolor{OrRd94}{RGB}{253,187,132}

\definecolor{OrRd9F}{RGB}{253,187,132}

\definecolor{OrRd95}{RGB}{252,141,89}

\definecolor{OrRd9G}{RGB}{252,141,89}

\definecolor{OrRd96}{RGB}{239,101,72}

\definecolor{OrRd9H}{RGB}{239,101,72}

\definecolor{OrRd97}{RGB}{215,48,31}

\definecolor{OrRd9J}{RGB}{215,48,31}

\definecolor{OrRd98}{RGB}{179,0,0}

\definecolor{OrRd9K}{RGB}{179,0,0}

\definecolor{OrRd99}{RGB}{127,0,0}

\definecolor{OrRd9M}{RGB}{127,0,0}

\definecolor{YlOrRd31}{RGB}{255,237,160}

\definecolor{YlOrRd3C}{RGB}{255,237,160}

\definecolor{YlOrRd32}{RGB}{254,178,76}

\definecolor{YlOrRd3F}{RGB}{254,178,76}

\definecolor{YlOrRd33}{RGB}{240,59,32}

\definecolor{YlOrRd3I}{RGB}{240,59,32}

\definecolor{YlOrRd41}{RGB}{255,255,178}

\definecolor{YlOrRd4B}{RGB}{255,255,178}

\definecolor{YlOrRd42}{RGB}{254,204,92}

\definecolor{YlOrRd4E}{RGB}{254,204,92}

\definecolor{YlOrRd43}{RGB}{253,141,60}

\definecolor{YlOrRd4G}{RGB}{253,141,60}

\definecolor{YlOrRd44}{RGB}{227,26,28}

\definecolor{YlOrRd4J}{RGB}{227,26,28}

\definecolor{YlOrRd51}{RGB}{255,255,178}

\definecolor{YlOrRd5B}{RGB}{255,255,178}

\definecolor{YlOrRd52}{RGB}{254,204,92}

\definecolor{YlOrRd5E}{RGB}{254,204,92}

\definecolor{YlOrRd53}{RGB}{253,141,60}

\definecolor{YlOrRd5G}{RGB}{253,141,60}

\definecolor{YlOrRd54}{RGB}{240,59,32}

\definecolor{YlOrRd5I}{RGB}{240,59,32}

\definecolor{YlOrRd55}{RGB}{189,0,38}

\definecolor{YlOrRd5K}{RGB}{189,0,38}

\definecolor{YlOrRd61}{RGB}{255,255,178}

\definecolor{YlOrRd6B}{RGB}{255,255,178}

\definecolor{YlOrRd62}{RGB}{254,217,118}

\definecolor{YlOrRd6D}{RGB}{254,217,118}

\definecolor{YlOrRd63}{RGB}{254,178,76}

\definecolor{YlOrRd6F}{RGB}{254,178,76}

\definecolor{YlOrRd64}{RGB}{253,141,60}

\definecolor{YlOrRd6G}{RGB}{253,141,60}

\definecolor{YlOrRd65}{RGB}{240,59,32}

\definecolor{YlOrRd6I}{RGB}{240,59,32}

\definecolor{YlOrRd66}{RGB}{189,0,38}

\definecolor{YlOrRd6K}{RGB}{189,0,38}

\definecolor{YlOrRd71}{RGB}{255,255,178}

\definecolor{YlOrRd7B}{RGB}{255,255,178}

\definecolor{YlOrRd72}{RGB}{254,217,118}

\definecolor{YlOrRd7D}{RGB}{254,217,118}

\definecolor{YlOrRd73}{RGB}{254,178,76}

\definecolor{YlOrRd7F}{RGB}{254,178,76}

\definecolor{YlOrRd74}{RGB}{253,141,60}

\definecolor{YlOrRd7G}{RGB}{253,141,60}

\definecolor{YlOrRd75}{RGB}{252,78,42}

\definecolor{YlOrRd7H}{RGB}{252,78,42}

\definecolor{YlOrRd76}{RGB}{227,26,28}

\definecolor{YlOrRd7J}{RGB}{227,26,28}

\definecolor{YlOrRd77}{RGB}{177,0,38}

\definecolor{YlOrRd7L}{RGB}{177,0,38}

\definecolor{YlOrRd81}{RGB}{255,255,204}

\definecolor{YlOrRd8A}{RGB}{255,255,204}

\definecolor{YlOrRd82}{RGB}{255,237,160}

\definecolor{YlOrRd8C}{RGB}{255,237,160}

\definecolor{YlOrRd83}{RGB}{254,217,118}

\definecolor{YlOrRd8D}{RGB}{254,217,118}

\definecolor{YlOrRd84}{RGB}{254,178,76}

\definecolor{YlOrRd8F}{RGB}{254,178,76}

\definecolor{YlOrRd85}{RGB}{253,141,60}

\definecolor{YlOrRd8G}{RGB}{253,141,60}

\definecolor{YlOrRd86}{RGB}{252,78,42}

\definecolor{YlOrRd8H}{RGB}{252,78,42}

\definecolor{YlOrRd87}{RGB}{227,26,28}

\definecolor{YlOrRd8J}{RGB}{227,26,28}

\definecolor{YlOrRd88}{RGB}{177,0,38}

\definecolor{YlOrRd8L}{RGB}{177,0,38}

\definecolor{YlOrRd91}{RGB}{255,255,204}

\definecolor{YlOrRd9A}{RGB}{255,255,204}

\definecolor{YlOrRd92}{RGB}{255,237,160}

\definecolor{YlOrRd9C}{RGB}{255,237,160}

\definecolor{YlOrRd93}{RGB}{254,217,118}

\definecolor{YlOrRd9D}{RGB}{254,217,118}

\definecolor{YlOrRd94}{RGB}{254,178,76}

\definecolor{YlOrRd9F}{RGB}{254,178,76}

\definecolor{YlOrRd95}{RGB}{253,141,60}

\definecolor{YlOrRd9G}{RGB}{253,141,60}

\definecolor{YlOrRd96}{RGB}{252,78,42}

\definecolor{YlOrRd9H}{RGB}{252,78,42}

\definecolor{YlOrRd97}{RGB}{227,26,28}

\definecolor{YlOrRd9J}{RGB}{227,26,28}

\definecolor{YlOrRd98}{RGB}{189,0,38}

\definecolor{YlOrRd9K}{RGB}{189,0,38}

\definecolor{YlOrRd99}{RGB}{128,0,38}

\definecolor{YlOrRd9M}{RGB}{128,0,38}

\definecolor{YlOrBr31}{RGB}{255,247,188}

\definecolor{YlOrBr3C}{RGB}{255,247,188}

\definecolor{YlOrBr32}{RGB}{254,196,79}

\definecolor{YlOrBr3F}{RGB}{254,196,79}

\definecolor{YlOrBr33}{RGB}{217,95,14}

\definecolor{YlOrBr3I}{RGB}{217,95,14}

\definecolor{YlOrBr41}{RGB}{255,255,212}

\definecolor{YlOrBr4B}{RGB}{255,255,212}

\definecolor{YlOrBr42}{RGB}{254,217,142}

\definecolor{YlOrBr4E}{RGB}{254,217,142}

\definecolor{YlOrBr43}{RGB}{254,153,41}

\definecolor{YlOrBr4G}{RGB}{254,153,41}

\definecolor{YlOrBr44}{RGB}{204,76,2}

\definecolor{YlOrBr4J}{RGB}{204,76,2}

\definecolor{YlOrBr51}{RGB}{255,255,212}

\definecolor{YlOrBr5B}{RGB}{255,255,212}

\definecolor{YlOrBr52}{RGB}{254,217,142}

\definecolor{YlOrBr5E}{RGB}{254,217,142}

\definecolor{YlOrBr53}{RGB}{254,153,41}

\definecolor{YlOrBr5G}{RGB}{254,153,41}

\definecolor{YlOrBr54}{RGB}{217,95,14}

\definecolor{YlOrBr5I}{RGB}{217,95,14}

\definecolor{YlOrBr55}{RGB}{153,52,4}

\definecolor{YlOrBr5K}{RGB}{153,52,4}

\definecolor{YlOrBr61}{RGB}{255,255,212}

\definecolor{YlOrBr6B}{RGB}{255,255,212}

\definecolor{YlOrBr62}{RGB}{254,227,145}

\definecolor{YlOrBr6D}{RGB}{254,227,145}

\definecolor{YlOrBr63}{RGB}{254,196,79}

\definecolor{YlOrBr6F}{RGB}{254,196,79}

\definecolor{YlOrBr64}{RGB}{254,153,41}

\definecolor{YlOrBr6G}{RGB}{254,153,41}

\definecolor{YlOrBr65}{RGB}{217,95,14}

\definecolor{YlOrBr6I}{RGB}{217,95,14}

\definecolor{YlOrBr66}{RGB}{153,52,4}

\definecolor{YlOrBr6K}{RGB}{153,52,4}

\definecolor{YlOrBr71}{RGB}{255,255,212}

\definecolor{YlOrBr7B}{RGB}{255,255,212}

\definecolor{YlOrBr72}{RGB}{254,227,145}

\definecolor{YlOrBr7D}{RGB}{254,227,145}

\definecolor{YlOrBr73}{RGB}{254,196,79}

\definecolor{YlOrBr7F}{RGB}{254,196,79}

\definecolor{YlOrBr74}{RGB}{254,153,41}

\definecolor{YlOrBr7G}{RGB}{254,153,41}

\definecolor{YlOrBr75}{RGB}{236,112,20}

\definecolor{YlOrBr7H}{RGB}{236,112,20}

\definecolor{YlOrBr76}{RGB}{204,76,2}

\definecolor{YlOrBr7J}{RGB}{204,76,2}

\definecolor{YlOrBr77}{RGB}{140,45,4}

\definecolor{YlOrBr7L}{RGB}{140,45,4}

\definecolor{YlOrBr81}{RGB}{255,255,229}

\definecolor{YlOrBr8A}{RGB}{255,255,229}

\definecolor{YlOrBr82}{RGB}{255,247,188}

\definecolor{YlOrBr8C}{RGB}{255,247,188}

\definecolor{YlOrBr83}{RGB}{254,227,145}

\definecolor{YlOrBr8D}{RGB}{254,227,145}

\definecolor{YlOrBr84}{RGB}{254,196,79}

\definecolor{YlOrBr8F}{RGB}{254,196,79}

\definecolor{YlOrBr85}{RGB}{254,153,41}

\definecolor{YlOrBr8G}{RGB}{254,153,41}

\definecolor{YlOrBr86}{RGB}{236,112,20}

\definecolor{YlOrBr8H}{RGB}{236,112,20}

\definecolor{YlOrBr87}{RGB}{204,76,2}

\definecolor{YlOrBr8J}{RGB}{204,76,2}

\definecolor{YlOrBr88}{RGB}{140,45,4}

\definecolor{YlOrBr8L}{RGB}{140,45,4}

\definecolor{YlOrBr91}{RGB}{255,255,229}

\definecolor{YlOrBr9A}{RGB}{255,255,229}

\definecolor{YlOrBr92}{RGB}{255,247,188}

\definecolor{YlOrBr9C}{RGB}{255,247,188}

\definecolor{YlOrBr93}{RGB}{254,227,145}

\definecolor{YlOrBr9D}{RGB}{254,227,145}

\definecolor{YlOrBr94}{RGB}{254,196,79}

\definecolor{YlOrBr9F}{RGB}{254,196,79}

\definecolor{YlOrBr95}{RGB}{254,153,41}

\definecolor{YlOrBr9G}{RGB}{254,153,41}

\definecolor{YlOrBr96}{RGB}{236,112,20}

\definecolor{YlOrBr9H}{RGB}{236,112,20}

\definecolor{YlOrBr97}{RGB}{204,76,2}

\definecolor{YlOrBr9J}{RGB}{204,76,2}

\definecolor{YlOrBr98}{RGB}{153,52,4}

\definecolor{YlOrBr9K}{RGB}{153,52,4}

\definecolor{YlOrBr99}{RGB}{102,37,6}

\definecolor{YlOrBr9M}{RGB}{102,37,6}

\definecolor{Purples31}{RGB}{239,237,245}

\definecolor{Purples3C}{RGB}{239,237,245}

\definecolor{Purples32}{RGB}{188,189,220}

\definecolor{Purples3F}{RGB}{188,189,220}

\definecolor{Purples33}{RGB}{117,107,177}

\definecolor{Purples3I}{RGB}{117,107,177}

\definecolor{Purples41}{RGB}{242,240,247}

\definecolor{Purples4B}{RGB}{242,240,247}

\definecolor{Purples42}{RGB}{203,201,226}

\definecolor{Purples4E}{RGB}{203,201,226}

\definecolor{Purples43}{RGB}{158,154,200}

\definecolor{Purples4G}{RGB}{158,154,200}

\definecolor{Purples44}{RGB}{106,81,163}

\definecolor{Purples4J}{RGB}{106,81,163}

\definecolor{Purples51}{RGB}{242,240,247}

\definecolor{Purples5B}{RGB}{242,240,247}

\definecolor{Purples52}{RGB}{203,201,226}

\definecolor{Purples5E}{RGB}{203,201,226}

\definecolor{Purples53}{RGB}{158,154,200}

\definecolor{Purples5G}{RGB}{158,154,200}

\definecolor{Purples54}{RGB}{117,107,177}

\definecolor{Purples5I}{RGB}{117,107,177}

\definecolor{Purples55}{RGB}{84,39,143}

\definecolor{Purples5K}{RGB}{84,39,143}

\definecolor{Purples61}{RGB}{242,240,247}

\definecolor{Purples6B}{RGB}{242,240,247}

\definecolor{Purples62}{RGB}{218,218,235}

\definecolor{Purples6D}{RGB}{218,218,235}

\definecolor{Purples63}{RGB}{188,189,220}

\definecolor{Purples6F}{RGB}{188,189,220}

\definecolor{Purples64}{RGB}{158,154,200}

\definecolor{Purples6G}{RGB}{158,154,200}

\definecolor{Purples65}{RGB}{117,107,177}

\definecolor{Purples6I}{RGB}{117,107,177}

\definecolor{Purples66}{RGB}{84,39,143}

\definecolor{Purples6K}{RGB}{84,39,143}

\definecolor{Purples71}{RGB}{242,240,247}

\definecolor{Purples7B}{RGB}{242,240,247}

\definecolor{Purples72}{RGB}{218,218,235}

\definecolor{Purples7D}{RGB}{218,218,235}

\definecolor{Purples73}{RGB}{188,189,220}

\definecolor{Purples7F}{RGB}{188,189,220}

\definecolor{Purples74}{RGB}{158,154,200}

\definecolor{Purples7G}{RGB}{158,154,200}

\definecolor{Purples75}{RGB}{128,125,186}

\definecolor{Purples7H}{RGB}{128,125,186}

\definecolor{Purples76}{RGB}{106,81,163}

\definecolor{Purples7J}{RGB}{106,81,163}

\definecolor{Purples77}{RGB}{74,20,134}

\definecolor{Purples7L}{RGB}{74,20,134}

\definecolor{Purples81}{RGB}{252,251,253}

\definecolor{Purples8A}{RGB}{252,251,253}

\definecolor{Purples82}{RGB}{239,237,245}

\definecolor{Purples8C}{RGB}{239,237,245}

\definecolor{Purples83}{RGB}{218,218,235}

\definecolor{Purples8D}{RGB}{218,218,235}

\definecolor{Purples84}{RGB}{188,189,220}

\definecolor{Purples8F}{RGB}{188,189,220}

\definecolor{Purples85}{RGB}{158,154,200}

\definecolor{Purples8G}{RGB}{158,154,200}

\definecolor{Purples86}{RGB}{128,125,186}

\definecolor{Purples8H}{RGB}{128,125,186}

\definecolor{Purples87}{RGB}{106,81,163}

\definecolor{Purples8J}{RGB}{106,81,163}

\definecolor{Purples88}{RGB}{74,20,134}

\definecolor{Purples8L}{RGB}{74,20,134}

\definecolor{Purples91}{RGB}{252,251,253}

\definecolor{Purples9A}{RGB}{252,251,253}

\definecolor{Purples92}{RGB}{239,237,245}

\definecolor{Purples9C}{RGB}{239,237,245}

\definecolor{Purples93}{RGB}{218,218,235}

\definecolor{Purples9D}{RGB}{218,218,235}

\definecolor{Purples94}{RGB}{188,189,220}

\definecolor{Purples9F}{RGB}{188,189,220}

\definecolor{Purples95}{RGB}{158,154,200}

\definecolor{Purples9G}{RGB}{158,154,200}

\definecolor{Purples96}{RGB}{128,125,186}

\definecolor{Purples9H}{RGB}{128,125,186}

\definecolor{Purples97}{RGB}{106,81,163}

\definecolor{Purples9J}{RGB}{106,81,163}

\definecolor{Purples98}{RGB}{84,39,143}

\definecolor{Purples9K}{RGB}{84,39,143}

\definecolor{Purples99}{RGB}{63,0,125}

\definecolor{Purples9M}{RGB}{63,0,125}

\definecolor{Blues31}{RGB}{222,235,247}

\definecolor{Blues3C}{RGB}{222,235,247}

\definecolor{Blues32}{RGB}{158,202,225}

\definecolor{Blues3F}{RGB}{158,202,225}

\definecolor{Blues33}{RGB}{49,130,189}

\definecolor{Blues3I}{RGB}{49,130,189}

\definecolor{Blues41}{RGB}{239,243,255}

\definecolor{Blues4B}{RGB}{239,243,255}

\definecolor{Blues42}{RGB}{189,215,231}

\definecolor{Blues4E}{RGB}{189,215,231}

\definecolor{Blues43}{RGB}{107,174,214}

\definecolor{Blues4G}{RGB}{107,174,214}

\definecolor{Blues44}{RGB}{33,113,181}

\definecolor{Blues4J}{RGB}{33,113,181}

\definecolor{Blues51}{RGB}{239,243,255}

\definecolor{Blues5B}{RGB}{239,243,255}

\definecolor{Blues52}{RGB}{189,215,231}

\definecolor{Blues5E}{RGB}{189,215,231}

\definecolor{Blues53}{RGB}{107,174,214}

\definecolor{Blues5G}{RGB}{107,174,214}

\definecolor{Blues54}{RGB}{49,130,189}

\definecolor{Blues5I}{RGB}{49,130,189}

\definecolor{Blues55}{RGB}{8,81,156}

\definecolor{Blues5K}{RGB}{8,81,156}

\definecolor{Blues61}{RGB}{239,243,255}

\definecolor{Blues6B}{RGB}{239,243,255}

\definecolor{Blues62}{RGB}{198,219,239}

\definecolor{Blues6D}{RGB}{198,219,239}

\definecolor{Blues63}{RGB}{158,202,225}

\definecolor{Blues6F}{RGB}{158,202,225}

\definecolor{Blues64}{RGB}{107,174,214}

\definecolor{Blues6G}{RGB}{107,174,214}

\definecolor{Blues65}{RGB}{49,130,189}

\definecolor{Blues6I}{RGB}{49,130,189}

\definecolor{Blues66}{RGB}{8,81,156}

\definecolor{Blues6K}{RGB}{8,81,156}

\definecolor{Blues71}{RGB}{239,243,255}

\definecolor{Blues7B}{RGB}{239,243,255}

\definecolor{Blues72}{RGB}{198,219,239}

\definecolor{Blues7D}{RGB}{198,219,239}

\definecolor{Blues73}{RGB}{158,202,225}

\definecolor{Blues7F}{RGB}{158,202,225}

\definecolor{Blues74}{RGB}{107,174,214}

\definecolor{Blues7G}{RGB}{107,174,214}

\definecolor{Blues75}{RGB}{66,146,198}

\definecolor{Blues7H}{RGB}{66,146,198}

\definecolor{Blues76}{RGB}{33,113,181}

\definecolor{Blues7J}{RGB}{33,113,181}

\definecolor{Blues77}{RGB}{8,69,148}

\definecolor{Blues7L}{RGB}{8,69,148}

\definecolor{Blues81}{RGB}{247,251,255}

\definecolor{Blues8A}{RGB}{247,251,255}

\definecolor{Blues82}{RGB}{222,235,247}

\definecolor{Blues8C}{RGB}{222,235,247}

\definecolor{Blues83}{RGB}{198,219,239}

\definecolor{Blues8D}{RGB}{198,219,239}

\definecolor{Blues84}{RGB}{158,202,225}

\definecolor{Blues8F}{RGB}{158,202,225}

\definecolor{Blues85}{RGB}{107,174,214}

\definecolor{Blues8G}{RGB}{107,174,214}

\definecolor{Blues86}{RGB}{66,146,198}

\definecolor{Blues8H}{RGB}{66,146,198}

\definecolor{Blues87}{RGB}{33,113,181}

\definecolor{Blues8J}{RGB}{33,113,181}

\definecolor{Blues88}{RGB}{8,69,148}

\definecolor{Blues8L}{RGB}{8,69,148}

\definecolor{Blues91}{RGB}{247,251,255}

\definecolor{Blues9A}{RGB}{247,251,255}

\definecolor{Blues92}{RGB}{222,235,247}

\definecolor{Blues9C}{RGB}{222,235,247}

\definecolor{Blues93}{RGB}{198,219,239}

\definecolor{Blues9D}{RGB}{198,219,239}

\definecolor{Blues94}{RGB}{158,202,225}

\definecolor{Blues9F}{RGB}{158,202,225}

\definecolor{Blues95}{RGB}{107,174,214}

\definecolor{Blues9G}{RGB}{107,174,214}

\definecolor{Blues96}{RGB}{66,146,198}

\definecolor{Blues9H}{RGB}{66,146,198}

\definecolor{Blues97}{RGB}{33,113,181}

\definecolor{Blues9J}{RGB}{33,113,181}

\definecolor{Blues98}{RGB}{8,81,156}

\definecolor{Blues9K}{RGB}{8,81,156}

\definecolor{Blues99}{RGB}{8,48,107}

\definecolor{Blues9M}{RGB}{8,48,107}

\definecolor{Greens31}{RGB}{229,245,224}

\definecolor{Greens3C}{RGB}{229,245,224}

\definecolor{Greens32}{RGB}{161,217,155}

\definecolor{Greens3F}{RGB}{161,217,155}

\definecolor{Greens33}{RGB}{49,163,84}

\definecolor{Greens3I}{RGB}{49,163,84}

\definecolor{Greens41}{RGB}{237,248,233}

\definecolor{Greens4B}{RGB}{237,248,233}

\definecolor{Greens42}{RGB}{186,228,179}

\definecolor{Greens4E}{RGB}{186,228,179}

\definecolor{Greens43}{RGB}{116,196,118}

\definecolor{Greens4G}{RGB}{116,196,118}

\definecolor{Greens44}{RGB}{35,139,69}

\definecolor{Greens4J}{RGB}{35,139,69}

\definecolor{Greens51}{RGB}{237,248,233}

\definecolor{Greens5B}{RGB}{237,248,233}

\definecolor{Greens52}{RGB}{186,228,179}

\definecolor{Greens5E}{RGB}{186,228,179}

\definecolor{Greens53}{RGB}{116,196,118}

\definecolor{Greens5G}{RGB}{116,196,118}

\definecolor{Greens54}{RGB}{49,163,84}

\definecolor{Greens5I}{RGB}{49,163,84}

\definecolor{Greens55}{RGB}{0,109,44}

\definecolor{Greens5K}{RGB}{0,109,44}

\definecolor{Greens61}{RGB}{237,248,233}

\definecolor{Greens6B}{RGB}{237,248,233}

\definecolor{Greens62}{RGB}{199,233,192}

\definecolor{Greens6D}{RGB}{199,233,192}

\definecolor{Greens63}{RGB}{161,217,155}

\definecolor{Greens6F}{RGB}{161,217,155}

\definecolor{Greens64}{RGB}{116,196,118}

\definecolor{Greens6G}{RGB}{116,196,118}

\definecolor{Greens65}{RGB}{49,163,84}

\definecolor{Greens6I}{RGB}{49,163,84}

\definecolor{Greens66}{RGB}{0,109,44}

\definecolor{Greens6K}{RGB}{0,109,44}

\definecolor{Greens71}{RGB}{237,248,233}

\definecolor{Greens7B}{RGB}{237,248,233}

\definecolor{Greens72}{RGB}{199,233,192}

\definecolor{Greens7D}{RGB}{199,233,192}

\definecolor{Greens73}{RGB}{161,217,155}

\definecolor{Greens7F}{RGB}{161,217,155}

\definecolor{Greens74}{RGB}{116,196,118}

\definecolor{Greens7G}{RGB}{116,196,118}

\definecolor{Greens75}{RGB}{65,171,93}

\definecolor{Greens7H}{RGB}{65,171,93}

\definecolor{Greens76}{RGB}{35,139,69}

\definecolor{Greens7J}{RGB}{35,139,69}

\definecolor{Greens77}{RGB}{0,90,50}

\definecolor{Greens7L}{RGB}{0,90,50}

\definecolor{Greens81}{RGB}{247,252,245}

\definecolor{Greens8A}{RGB}{247,252,245}

\definecolor{Greens82}{RGB}{229,245,224}

\definecolor{Greens8C}{RGB}{229,245,224}

\definecolor{Greens83}{RGB}{199,233,192}

\definecolor{Greens8D}{RGB}{199,233,192}

\definecolor{Greens84}{RGB}{161,217,155}

\definecolor{Greens8F}{RGB}{161,217,155}

\definecolor{Greens85}{RGB}{116,196,118}

\definecolor{Greens8G}{RGB}{116,196,118}

\definecolor{Greens86}{RGB}{65,171,93}

\definecolor{Greens8H}{RGB}{65,171,93}

\definecolor{Greens87}{RGB}{35,139,69}

\definecolor{Greens8J}{RGB}{35,139,69}

\definecolor{Greens88}{RGB}{0,90,50}

\definecolor{Greens8L}{RGB}{0,90,50}

\definecolor{Greens91}{RGB}{247,252,245}

\definecolor{Greens9A}{RGB}{247,252,245}

\definecolor{Greens92}{RGB}{229,245,224}

\definecolor{Greens9C}{RGB}{229,245,224}

\definecolor{Greens93}{RGB}{199,233,192}

\definecolor{Greens9D}{RGB}{199,233,192}

\definecolor{Greens94}{RGB}{161,217,155}

\definecolor{Greens9F}{RGB}{161,217,155}

\definecolor{Greens95}{RGB}{116,196,118}

\definecolor{Greens9G}{RGB}{116,196,118}

\definecolor{Greens96}{RGB}{65,171,93}

\definecolor{Greens9H}{RGB}{65,171,93}

\definecolor{Greens97}{RGB}{35,139,69}

\definecolor{Greens9J}{RGB}{35,139,69}

\definecolor{Greens98}{RGB}{0,109,44}

\definecolor{Greens9K}{RGB}{0,109,44}

\definecolor{Greens99}{RGB}{0,68,27}

\definecolor{Greens9M}{RGB}{0,68,27}

\definecolor{Oranges31}{RGB}{254,230,206}

\definecolor{Oranges3C}{RGB}{254,230,206}

\definecolor{Oranges32}{RGB}{253,174,107}

\definecolor{Oranges3F}{RGB}{253,174,107}

\definecolor{Oranges33}{RGB}{230,85,13}

\definecolor{Oranges3I}{RGB}{230,85,13}

\definecolor{Oranges41}{RGB}{254,237,222}

\definecolor{Oranges4B}{RGB}{254,237,222}

\definecolor{Oranges42}{RGB}{253,190,133}

\definecolor{Oranges4E}{RGB}{253,190,133}

\definecolor{Oranges43}{RGB}{253,141,60}

\definecolor{Oranges4G}{RGB}{253,141,60}

\definecolor{Oranges44}{RGB}{217,71,1}

\definecolor{Oranges4J}{RGB}{217,71,1}

\definecolor{Oranges51}{RGB}{254,237,222}

\definecolor{Oranges5B}{RGB}{254,237,222}

\definecolor{Oranges52}{RGB}{253,190,133}

\definecolor{Oranges5E}{RGB}{253,190,133}

\definecolor{Oranges53}{RGB}{253,141,60}

\definecolor{Oranges5G}{RGB}{253,141,60}

\definecolor{Oranges54}{RGB}{230,85,13}

\definecolor{Oranges5I}{RGB}{230,85,13}

\definecolor{Oranges55}{RGB}{166,54,3}

\definecolor{Oranges5K}{RGB}{166,54,3}

\definecolor{Oranges61}{RGB}{254,237,222}

\definecolor{Oranges6B}{RGB}{254,237,222}

\definecolor{Oranges62}{RGB}{253,208,162}

\definecolor{Oranges6D}{RGB}{253,208,162}

\definecolor{Oranges63}{RGB}{253,174,107}

\definecolor{Oranges6F}{RGB}{253,174,107}

\definecolor{Oranges64}{RGB}{253,141,60}

\definecolor{Oranges6G}{RGB}{253,141,60}

\definecolor{Oranges65}{RGB}{230,85,13}

\definecolor{Oranges6I}{RGB}{230,85,13}

\definecolor{Oranges66}{RGB}{166,54,3}

\definecolor{Oranges6K}{RGB}{166,54,3}

\definecolor{Oranges71}{RGB}{254,237,222}

\definecolor{Oranges7B}{RGB}{254,237,222}

\definecolor{Oranges72}{RGB}{253,208,162}

\definecolor{Oranges7D}{RGB}{253,208,162}

\definecolor{Oranges73}{RGB}{253,174,107}

\definecolor{Oranges7F}{RGB}{253,174,107}

\definecolor{Oranges74}{RGB}{253,141,60}

\definecolor{Oranges7G}{RGB}{253,141,60}

\definecolor{Oranges75}{RGB}{241,105,19}

\definecolor{Oranges7H}{RGB}{241,105,19}

\definecolor{Oranges76}{RGB}{217,72,1}

\definecolor{Oranges7J}{RGB}{217,72,1}

\definecolor{Oranges77}{RGB}{140,45,4}

\definecolor{Oranges7L}{RGB}{140,45,4}

\definecolor{Oranges81}{RGB}{255,245,235}

\definecolor{Oranges8A}{RGB}{255,245,235}

\definecolor{Oranges82}{RGB}{254,230,206}

\definecolor{Oranges8C}{RGB}{254,230,206}

\definecolor{Oranges83}{RGB}{253,208,162}

\definecolor{Oranges8D}{RGB}{253,208,162}

\definecolor{Oranges84}{RGB}{253,174,107}

\definecolor{Oranges8F}{RGB}{253,174,107}

\definecolor{Oranges85}{RGB}{253,141,60}

\definecolor{Oranges8G}{RGB}{253,141,60}

\definecolor{Oranges86}{RGB}{241,105,19}

\definecolor{Oranges8H}{RGB}{241,105,19}

\definecolor{Oranges87}{RGB}{217,72,1}

\definecolor{Oranges8J}{RGB}{217,72,1}

\definecolor{Oranges88}{RGB}{140,45,4}

\definecolor{Oranges8L}{RGB}{140,45,4}

\definecolor{Oranges91}{RGB}{255,245,235}

\definecolor{Oranges9A}{RGB}{255,245,235}

\definecolor{Oranges92}{RGB}{254,230,206}

\definecolor{Oranges9C}{RGB}{254,230,206}

\definecolor{Oranges93}{RGB}{253,208,162}

\definecolor{Oranges9D}{RGB}{253,208,162}

\definecolor{Oranges94}{RGB}{253,174,107}

\definecolor{Oranges9F}{RGB}{253,174,107}

\definecolor{Oranges95}{RGB}{253,141,60}

\definecolor{Oranges9G}{RGB}{253,141,60}

\definecolor{Oranges96}{RGB}{241,105,19}

\definecolor{Oranges9H}{RGB}{241,105,19}

\definecolor{Oranges97}{RGB}{217,72,1}

\definecolor{Oranges9J}{RGB}{217,72,1}

\definecolor{Oranges98}{RGB}{166,54,3}

\definecolor{Oranges9K}{RGB}{166,54,3}

\definecolor{Oranges99}{RGB}{127,39,4}

\definecolor{Oranges9M}{RGB}{127,39,4}

\definecolor{Reds31}{RGB}{254,224,210}

\definecolor{Reds3C}{RGB}{254,224,210}

\definecolor{Reds32}{RGB}{252,146,114}

\definecolor{Reds3F}{RGB}{252,146,114}

\definecolor{Reds33}{RGB}{222,45,38}

\definecolor{Reds3I}{RGB}{222,45,38}

\definecolor{Reds41}{RGB}{254,229,217}

\definecolor{Reds4B}{RGB}{254,229,217}

\definecolor{Reds42}{RGB}{252,174,145}

\definecolor{Reds4E}{RGB}{252,174,145}

\definecolor{Reds43}{RGB}{251,106,74}

\definecolor{Reds4G}{RGB}{251,106,74}

\definecolor{Reds44}{RGB}{203,24,29}

\definecolor{Reds4J}{RGB}{203,24,29}

\definecolor{Reds51}{RGB}{254,229,217}

\definecolor{Reds5B}{RGB}{254,229,217}

\definecolor{Reds52}{RGB}{252,174,145}

\definecolor{Reds5E}{RGB}{252,174,145}

\definecolor{Reds53}{RGB}{251,106,74}

\definecolor{Reds5G}{RGB}{251,106,74}

\definecolor{Reds54}{RGB}{222,45,38}

\definecolor{Reds5I}{RGB}{222,45,38}

\definecolor{Reds55}{RGB}{165,15,21}

\definecolor{Reds5K}{RGB}{165,15,21}

\definecolor{Reds61}{RGB}{254,229,217}

\definecolor{Reds6B}{RGB}{254,229,217}

\definecolor{Reds62}{RGB}{252,187,161}

\definecolor{Reds6D}{RGB}{252,187,161}

\definecolor{Reds63}{RGB}{252,146,114}

\definecolor{Reds6F}{RGB}{252,146,114}

\definecolor{Reds64}{RGB}{251,106,74}

\definecolor{Reds6G}{RGB}{251,106,74}

\definecolor{Reds65}{RGB}{222,45,38}

\definecolor{Reds6I}{RGB}{222,45,38}

\definecolor{Reds66}{RGB}{165,15,21}

\definecolor{Reds6K}{RGB}{165,15,21}

\definecolor{Reds71}{RGB}{254,229,217}

\definecolor{Reds7B}{RGB}{254,229,217}

\definecolor{Reds72}{RGB}{252,187,161}

\definecolor{Reds7D}{RGB}{252,187,161}

\definecolor{Reds73}{RGB}{252,146,114}

\definecolor{Reds7F}{RGB}{252,146,114}

\definecolor{Reds74}{RGB}{251,106,74}

\definecolor{Reds7G}{RGB}{251,106,74}

\definecolor{Reds75}{RGB}{239,59,44}

\definecolor{Reds7H}{RGB}{239,59,44}

\definecolor{Reds76}{RGB}{203,24,29}

\definecolor{Reds7J}{RGB}{203,24,29}

\definecolor{Reds77}{RGB}{153,0,13}

\definecolor{Reds7L}{RGB}{153,0,13}

\definecolor{Reds81}{RGB}{255,245,240}

\definecolor{Reds8A}{RGB}{255,245,240}

\definecolor{Reds82}{RGB}{254,224,210}

\definecolor{Reds8C}{RGB}{254,224,210}

\definecolor{Reds83}{RGB}{252,187,161}

\definecolor{Reds8D}{RGB}{252,187,161}

\definecolor{Reds84}{RGB}{252,146,114}

\definecolor{Reds8F}{RGB}{252,146,114}

\definecolor{Reds85}{RGB}{251,106,74}

\definecolor{Reds8G}{RGB}{251,106,74}

\definecolor{Reds86}{RGB}{239,59,44}

\definecolor{Reds8H}{RGB}{239,59,44}

\definecolor{Reds87}{RGB}{203,24,29}

\definecolor{Reds8J}{RGB}{203,24,29}

\definecolor{Reds88}{RGB}{153,0,13}

\definecolor{Reds8L}{RGB}{153,0,13}

\definecolor{Reds91}{RGB}{255,245,240}

\definecolor{Reds9A}{RGB}{255,245,240}

\definecolor{Reds92}{RGB}{254,224,210}

\definecolor{Reds9C}{RGB}{254,224,210}

\definecolor{Reds93}{RGB}{252,187,161}

\definecolor{Reds9D}{RGB}{252,187,161}

\definecolor{Reds94}{RGB}{252,146,114}

\definecolor{Reds9F}{RGB}{252,146,114}

\definecolor{Reds95}{RGB}{251,106,74}

\definecolor{Reds9G}{RGB}{251,106,74}

\definecolor{Reds96}{RGB}{239,59,44}

\definecolor{Reds9H}{RGB}{239,59,44}

\definecolor{Reds97}{RGB}{203,24,29}

\definecolor{Reds9J}{RGB}{203,24,29}

\definecolor{Reds98}{RGB}{165,15,21}

\definecolor{Reds9K}{RGB}{165,15,21}

\definecolor{Reds99}{RGB}{103,0,13}

\definecolor{Reds9M}{RGB}{103,0,13}

\definecolor{Greys31}{RGB}{240,240,240}

\definecolor{Greys3C}{RGB}{240,240,240}

\definecolor{Greys32}{RGB}{189,189,189}

\definecolor{Greys3F}{RGB}{189,189,189}

\definecolor{Greys33}{RGB}{99,99,99}

\definecolor{Greys3I}{RGB}{99,99,99}

\definecolor{Greys41}{RGB}{247,247,247}

\definecolor{Greys4B}{RGB}{247,247,247}

\definecolor{Greys42}{RGB}{204,204,204}

\definecolor{Greys4E}{RGB}{204,204,204}

\definecolor{Greys43}{RGB}{150,150,150}

\definecolor{Greys4G}{RGB}{150,150,150}

\definecolor{Greys44}{RGB}{82,82,82}

\definecolor{Greys4J}{RGB}{82,82,82}

\definecolor{Greys51}{RGB}{247,247,247}

\definecolor{Greys5B}{RGB}{247,247,247}

\definecolor{Greys52}{RGB}{204,204,204}

\definecolor{Greys5E}{RGB}{204,204,204}

\definecolor{Greys53}{RGB}{150,150,150}

\definecolor{Greys5G}{RGB}{150,150,150}

\definecolor{Greys54}{RGB}{99,99,99}

\definecolor{Greys5I}{RGB}{99,99,99}

\definecolor{Greys55}{RGB}{37,37,37}

\definecolor{Greys5K}{RGB}{37,37,37}

\definecolor{Greys61}{RGB}{247,247,247}

\definecolor{Greys6B}{RGB}{247,247,247}

\definecolor{Greys62}{RGB}{217,217,217}

\definecolor{Greys6D}{RGB}{217,217,217}

\definecolor{Greys63}{RGB}{189,189,189}

\definecolor{Greys6F}{RGB}{189,189,189}

\definecolor{Greys64}{RGB}{150,150,150}

\definecolor{Greys6G}{RGB}{150,150,150}

\definecolor{Greys65}{RGB}{99,99,99}

\definecolor{Greys6I}{RGB}{99,99,99}

\definecolor{Greys66}{RGB}{37,37,37}

\definecolor{Greys6K}{RGB}{37,37,37}

\definecolor{Greys71}{RGB}{247,247,247}

\definecolor{Greys7B}{RGB}{247,247,247}

\definecolor{Greys72}{RGB}{217,217,217}

\definecolor{Greys7D}{RGB}{217,217,217}

\definecolor{Greys73}{RGB}{189,189,189}

\definecolor{Greys7F}{RGB}{189,189,189}

\definecolor{Greys74}{RGB}{150,150,150}

\definecolor{Greys7G}{RGB}{150,150,150}

\definecolor{Greys75}{RGB}{115,115,115}

\definecolor{Greys7H}{RGB}{115,115,115}

\definecolor{Greys76}{RGB}{82,82,82}

\definecolor{Greys7J}{RGB}{82,82,82}

\definecolor{Greys77}{RGB}{37,37,37}

\definecolor{Greys7L}{RGB}{37,37,37}

\definecolor{Greys81}{RGB}{255,255,255}

\definecolor{Greys8A}{RGB}{255,255,255}

\definecolor{Greys82}{RGB}{240,240,240}

\definecolor{Greys8C}{RGB}{240,240,240}

\definecolor{Greys83}{RGB}{217,217,217}

\definecolor{Greys8D}{RGB}{217,217,217}

\definecolor{Greys84}{RGB}{189,189,189}

\definecolor{Greys8F}{RGB}{189,189,189}

\definecolor{Greys85}{RGB}{150,150,150}

\definecolor{Greys8G}{RGB}{150,150,150}

\definecolor{Greys86}{RGB}{115,115,115}

\definecolor{Greys8H}{RGB}{115,115,115}

\definecolor{Greys87}{RGB}{82,82,82}

\definecolor{Greys8J}{RGB}{82,82,82}

\definecolor{Greys88}{RGB}{37,37,37}

\definecolor{Greys8L}{RGB}{37,37,37}

\definecolor{Greys91}{RGB}{255,255,255}

\definecolor{Greys9A}{RGB}{255,255,255}

\definecolor{Greys92}{RGB}{240,240,240}

\definecolor{Greys9C}{RGB}{240,240,240}

\definecolor{Greys93}{RGB}{217,217,217}

\definecolor{Greys9D}{RGB}{217,217,217}

\definecolor{Greys94}{RGB}{189,189,189}

\definecolor{Greys9F}{RGB}{189,189,189}

\definecolor{Greys95}{RGB}{150,150,150}

\definecolor{Greys9G}{RGB}{150,150,150}

\definecolor{Greys96}{RGB}{115,115,115}

\definecolor{Greys9H}{RGB}{115,115,115}

\definecolor{Greys97}{RGB}{82,82,82}

\definecolor{Greys9J}{RGB}{82,82,82}

\definecolor{Greys98}{RGB}{37,37,37}

\definecolor{Greys9K}{RGB}{37,37,37}

\definecolor{Greys99}{RGB}{0,0,0}

\definecolor{Greys9M}{RGB}{0,0,0}

\definecolor{PuOr31}{RGB}{241,163,64}

\definecolor{PuOr3E}{RGB}{241,163,64}

\definecolor{PuOr32}{RGB}{247,247,247}

\definecolor{PuOr3H}{RGB}{247,247,247}

\definecolor{PuOr33}{RGB}{153,142,195}

\definecolor{PuOr3K}{RGB}{153,142,195}

\definecolor{PuOr41}{RGB}{230,97,1}

\definecolor{PuOr4C}{RGB}{230,97,1}

\definecolor{PuOr42}{RGB}{253,184,99}

\definecolor{PuOr4F}{RGB}{253,184,99}

\definecolor{PuOr43}{RGB}{178,171,210}

\definecolor{PuOr4J}{RGB}{178,171,210}

\definecolor{PuOr44}{RGB}{94,60,153}

\definecolor{PuOr4M}{RGB}{94,60,153}

\definecolor{PuOr51}{RGB}{230,97,1}

\definecolor{PuOr5C}{RGB}{230,97,1}

\definecolor{PuOr52}{RGB}{253,184,99}

\definecolor{PuOr5F}{RGB}{253,184,99}

\definecolor{PuOr53}{RGB}{247,247,247}

\definecolor{PuOr5H}{RGB}{247,247,247}

\definecolor{PuOr54}{RGB}{178,171,210}

\definecolor{PuOr5J}{RGB}{178,171,210}

\definecolor{PuOr55}{RGB}{94,60,153}

\definecolor{PuOr5M}{RGB}{94,60,153}

\definecolor{PuOr61}{RGB}{179,88,6}

\definecolor{PuOr6B}{RGB}{179,88,6}

\definecolor{PuOr62}{RGB}{241,163,64}

\definecolor{PuOr6E}{RGB}{241,163,64}

\definecolor{PuOr63}{RGB}{254,224,182}

\definecolor{PuOr6G}{RGB}{254,224,182}

\definecolor{PuOr64}{RGB}{216,218,235}

\definecolor{PuOr6I}{RGB}{216,218,235}

\definecolor{PuOr65}{RGB}{153,142,195}

\definecolor{PuOr6K}{RGB}{153,142,195}

\definecolor{PuOr66}{RGB}{84,39,136}

\definecolor{PuOr6N}{RGB}{84,39,136}

\definecolor{PuOr71}{RGB}{179,88,6}

\definecolor{PuOr7B}{RGB}{179,88,6}

\definecolor{PuOr72}{RGB}{241,163,64}

\definecolor{PuOr7E}{RGB}{241,163,64}

\definecolor{PuOr73}{RGB}{254,224,182}

\definecolor{PuOr7G}{RGB}{254,224,182}

\definecolor{PuOr74}{RGB}{247,247,247}

\definecolor{PuOr7H}{RGB}{247,247,247}

\definecolor{PuOr75}{RGB}{216,218,235}

\definecolor{PuOr7I}{RGB}{216,218,235}

\definecolor{PuOr76}{RGB}{153,142,195}

\definecolor{PuOr7K}{RGB}{153,142,195}

\definecolor{PuOr77}{RGB}{84,39,136}

\definecolor{PuOr7N}{RGB}{84,39,136}

\definecolor{PuOr81}{RGB}{179,88,6}

\definecolor{PuOr8B}{RGB}{179,88,6}

\definecolor{PuOr82}{RGB}{224,130,20}

\definecolor{PuOr8D}{RGB}{224,130,20}

\definecolor{PuOr83}{RGB}{253,184,99}

\definecolor{PuOr8F}{RGB}{253,184,99}

\definecolor{PuOr84}{RGB}{254,224,182}

\definecolor{PuOr8G}{RGB}{254,224,182}

\definecolor{PuOr85}{RGB}{216,218,235}

\definecolor{PuOr8I}{RGB}{216,218,235}

\definecolor{PuOr86}{RGB}{178,171,210}

\definecolor{PuOr8J}{RGB}{178,171,210}

\definecolor{PuOr87}{RGB}{128,115,172}

\definecolor{PuOr8L}{RGB}{128,115,172}

\definecolor{PuOr88}{RGB}{84,39,136}

\definecolor{PuOr8N}{RGB}{84,39,136}

\definecolor{PuOr91}{RGB}{179,88,6}

\definecolor{PuOr9B}{RGB}{179,88,6}

\definecolor{PuOr92}{RGB}{224,130,20}

\definecolor{PuOr9D}{RGB}{224,130,20}

\definecolor{PuOr93}{RGB}{253,184,99}

\definecolor{PuOr9F}{RGB}{253,184,99}

\definecolor{PuOr94}{RGB}{254,224,182}

\definecolor{PuOr9G}{RGB}{254,224,182}

\definecolor{PuOr95}{RGB}{247,247,247}

\definecolor{PuOr9H}{RGB}{247,247,247}

\definecolor{PuOr96}{RGB}{216,218,235}

\definecolor{PuOr9I}{RGB}{216,218,235}

\definecolor{PuOr97}{RGB}{178,171,210}

\definecolor{PuOr9J}{RGB}{178,171,210}

\definecolor{PuOr98}{RGB}{128,115,172}

\definecolor{PuOr9L}{RGB}{128,115,172}

\definecolor{PuOr99}{RGB}{84,39,136}

\definecolor{PuOr9N}{RGB}{84,39,136}

\definecolor{PuOr101}{RGB}{127,59,8}

\definecolor{PuOr10A}{RGB}{127,59,8}

\definecolor{PuOr102}{RGB}{179,88,6}

\definecolor{PuOr10B}{RGB}{179,88,6}

\definecolor{PuOr103}{RGB}{224,130,20}

\definecolor{PuOr10D}{RGB}{224,130,20}

\definecolor{PuOr104}{RGB}{253,184,99}

\definecolor{PuOr10F}{RGB}{253,184,99}

\definecolor{PuOr105}{RGB}{254,224,182}

\definecolor{PuOr10G}{RGB}{254,224,182}

\definecolor{PuOr106}{RGB}{216,218,235}

\definecolor{PuOr10I}{RGB}{216,218,235}

\definecolor{PuOr107}{RGB}{178,171,210}

\definecolor{PuOr10J}{RGB}{178,171,210}

\definecolor{PuOr108}{RGB}{128,115,172}

\definecolor{PuOr10L}{RGB}{128,115,172}

\definecolor{PuOr109}{RGB}{84,39,136}

\definecolor{PuOr10N}{RGB}{84,39,136}

\definecolor{PuOr1010}{RGB}{45,0,75}

\definecolor{PuOr10O}{RGB}{45,0,75}

\definecolor{PuOr111}{RGB}{127,59,8}

\definecolor{PuOr11A}{RGB}{127,59,8}

\definecolor{PuOr112}{RGB}{179,88,6}

\definecolor{PuOr11B}{RGB}{179,88,6}

\definecolor{PuOr113}{RGB}{224,130,20}

\definecolor{PuOr11D}{RGB}{224,130,20}

\definecolor{PuOr114}{RGB}{253,184,99}

\definecolor{PuOr11F}{RGB}{253,184,99}

\definecolor{PuOr115}{RGB}{254,224,182}

\definecolor{PuOr11G}{RGB}{254,224,182}

\definecolor{PuOr116}{RGB}{247,247,247}

\definecolor{PuOr11H}{RGB}{247,247,247}

\definecolor{PuOr117}{RGB}{216,218,235}

\definecolor{PuOr11I}{RGB}{216,218,235}

\definecolor{PuOr118}{RGB}{178,171,210}

\definecolor{PuOr11J}{RGB}{178,171,210}

\definecolor{PuOr119}{RGB}{128,115,172}

\definecolor{PuOr11L}{RGB}{128,115,172}

\definecolor{PuOr1110}{RGB}{84,39,136}

\definecolor{PuOr11N}{RGB}{84,39,136}

\definecolor{PuOr1111}{RGB}{45,0,75}

\definecolor{PuOr11O}{RGB}{45,0,75}

\definecolor{BrBG31}{RGB}{216,179,101}

\definecolor{BrBG3E}{RGB}{216,179,101}

\definecolor{BrBG32}{RGB}{245,245,245}

\definecolor{BrBG3H}{RGB}{245,245,245}

\definecolor{BrBG33}{RGB}{90,180,172}

\definecolor{BrBG3K}{RGB}{90,180,172}

\definecolor{BrBG41}{RGB}{166,97,26}

\definecolor{BrBG4C}{RGB}{166,97,26}

\definecolor{BrBG42}{RGB}{223,194,125}

\definecolor{BrBG4F}{RGB}{223,194,125}

\definecolor{BrBG43}{RGB}{128,205,193}

\definecolor{BrBG4J}{RGB}{128,205,193}

\definecolor{BrBG44}{RGB}{1,133,113}

\definecolor{BrBG4M}{RGB}{1,133,113}

\definecolor{BrBG51}{RGB}{166,97,26}

\definecolor{BrBG5C}{RGB}{166,97,26}

\definecolor{BrBG52}{RGB}{223,194,125}

\definecolor{BrBG5F}{RGB}{223,194,125}

\definecolor{BrBG53}{RGB}{245,245,245}

\definecolor{BrBG5H}{RGB}{245,245,245}

\definecolor{BrBG54}{RGB}{128,205,193}

\definecolor{BrBG5J}{RGB}{128,205,193}

\definecolor{BrBG55}{RGB}{1,133,113}

\definecolor{BrBG5M}{RGB}{1,133,113}

\definecolor{BrBG61}{RGB}{140,81,10}

\definecolor{BrBG6B}{RGB}{140,81,10}

\definecolor{BrBG62}{RGB}{216,179,101}

\definecolor{BrBG6E}{RGB}{216,179,101}

\definecolor{BrBG63}{RGB}{246,232,195}

\definecolor{BrBG6G}{RGB}{246,232,195}

\definecolor{BrBG64}{RGB}{199,234,229}

\definecolor{BrBG6I}{RGB}{199,234,229}

\definecolor{BrBG65}{RGB}{90,180,172}

\definecolor{BrBG6K}{RGB}{90,180,172}

\definecolor{BrBG66}{RGB}{1,102,94}

\definecolor{BrBG6N}{RGB}{1,102,94}

\definecolor{BrBG71}{RGB}{140,81,10}

\definecolor{BrBG7B}{RGB}{140,81,10}

\definecolor{BrBG72}{RGB}{216,179,101}

\definecolor{BrBG7E}{RGB}{216,179,101}

\definecolor{BrBG73}{RGB}{246,232,195}

\definecolor{BrBG7G}{RGB}{246,232,195}

\definecolor{BrBG74}{RGB}{245,245,245}

\definecolor{BrBG7H}{RGB}{245,245,245}

\definecolor{BrBG75}{RGB}{199,234,229}

\definecolor{BrBG7I}{RGB}{199,234,229}

\definecolor{BrBG76}{RGB}{90,180,172}

\definecolor{BrBG7K}{RGB}{90,180,172}

\definecolor{BrBG77}{RGB}{1,102,94}

\definecolor{BrBG7N}{RGB}{1,102,94}

\definecolor{BrBG81}{RGB}{140,81,10}

\definecolor{BrBG8B}{RGB}{140,81,10}

\definecolor{BrBG82}{RGB}{191,129,45}

\definecolor{BrBG8D}{RGB}{191,129,45}

\definecolor{BrBG83}{RGB}{223,194,125}

\definecolor{BrBG8F}{RGB}{223,194,125}

\definecolor{BrBG84}{RGB}{246,232,195}

\definecolor{BrBG8G}{RGB}{246,232,195}

\definecolor{BrBG85}{RGB}{199,234,229}

\definecolor{BrBG8I}{RGB}{199,234,229}

\definecolor{BrBG86}{RGB}{128,205,193}

\definecolor{BrBG8J}{RGB}{128,205,193}

\definecolor{BrBG87}{RGB}{53,151,143}

\definecolor{BrBG8L}{RGB}{53,151,143}

\definecolor{BrBG88}{RGB}{1,102,94}

\definecolor{BrBG8N}{RGB}{1,102,94}

\definecolor{BrBG91}{RGB}{140,81,10}

\definecolor{BrBG9B}{RGB}{140,81,10}

\definecolor{BrBG92}{RGB}{191,129,45}

\definecolor{BrBG9D}{RGB}{191,129,45}

\definecolor{BrBG93}{RGB}{223,194,125}

\definecolor{BrBG9F}{RGB}{223,194,125}

\definecolor{BrBG94}{RGB}{246,232,195}

\definecolor{BrBG9G}{RGB}{246,232,195}

\definecolor{BrBG95}{RGB}{245,245,245}

\definecolor{BrBG9H}{RGB}{245,245,245}

\definecolor{BrBG96}{RGB}{199,234,229}

\definecolor{BrBG9I}{RGB}{199,234,229}

\definecolor{BrBG97}{RGB}{128,205,193}

\definecolor{BrBG9J}{RGB}{128,205,193}

\definecolor{BrBG98}{RGB}{53,151,143}

\definecolor{BrBG9L}{RGB}{53,151,143}

\definecolor{BrBG99}{RGB}{1,102,94}

\definecolor{BrBG9N}{RGB}{1,102,94}

\definecolor{BrBG101}{RGB}{84,48,5}

\definecolor{BrBG10A}{RGB}{84,48,5}

\definecolor{BrBG102}{RGB}{140,81,10}

\definecolor{BrBG10B}{RGB}{140,81,10}

\definecolor{BrBG103}{RGB}{191,129,45}

\definecolor{BrBG10D}{RGB}{191,129,45}

\definecolor{BrBG104}{RGB}{223,194,125}

\definecolor{BrBG10F}{RGB}{223,194,125}

\definecolor{BrBG105}{RGB}{246,232,195}

\definecolor{BrBG10G}{RGB}{246,232,195}

\definecolor{BrBG106}{RGB}{199,234,229}

\definecolor{BrBG10I}{RGB}{199,234,229}

\definecolor{BrBG107}{RGB}{128,205,193}

\definecolor{BrBG10J}{RGB}{128,205,193}

\definecolor{BrBG108}{RGB}{53,151,143}

\definecolor{BrBG10L}{RGB}{53,151,143}

\definecolor{BrBG109}{RGB}{1,102,94}

\definecolor{BrBG10N}{RGB}{1,102,94}

\definecolor{BrBG1010}{RGB}{0,60,48}

\definecolor{BrBG10O}{RGB}{0,60,48}

\definecolor{BrBG111}{RGB}{84,48,5}

\definecolor{BrBG11A}{RGB}{84,48,5}

\definecolor{BrBG112}{RGB}{140,81,10}

\definecolor{BrBG11B}{RGB}{140,81,10}

\definecolor{BrBG113}{RGB}{191,129,45}

\definecolor{BrBG11D}{RGB}{191,129,45}

\definecolor{BrBG114}{RGB}{223,194,125}

\definecolor{BrBG11F}{RGB}{223,194,125}

\definecolor{BrBG115}{RGB}{246,232,195}

\definecolor{BrBG11G}{RGB}{246,232,195}

\definecolor{BrBG116}{RGB}{245,245,245}

\definecolor{BrBG11H}{RGB}{245,245,245}

\definecolor{BrBG117}{RGB}{199,234,229}

\definecolor{BrBG11I}{RGB}{199,234,229}

\definecolor{BrBG118}{RGB}{128,205,193}

\definecolor{BrBG11J}{RGB}{128,205,193}

\definecolor{BrBG119}{RGB}{53,151,143}

\definecolor{BrBG11L}{RGB}{53,151,143}

\definecolor{BrBG1110}{RGB}{1,102,94}

\definecolor{BrBG11N}{RGB}{1,102,94}

\definecolor{BrBG1111}{RGB}{0,60,48}

\definecolor{BrBG11O}{RGB}{0,60,48}

\definecolor{PRGn31}{RGB}{175,141,195}

\definecolor{PRGn3E}{RGB}{175,141,195}

\definecolor{PRGn32}{RGB}{247,247,247}

\definecolor{PRGn3H}{RGB}{247,247,247}

\definecolor{PRGn33}{RGB}{127,191,123}

\definecolor{PRGn3K}{RGB}{127,191,123}

\definecolor{PRGn41}{RGB}{123,50,148}

\definecolor{PRGn4C}{RGB}{123,50,148}

\definecolor{PRGn42}{RGB}{194,165,207}

\definecolor{PRGn4F}{RGB}{194,165,207}

\definecolor{PRGn43}{RGB}{166,219,160}

\definecolor{PRGn4J}{RGB}{166,219,160}

\definecolor{PRGn44}{RGB}{0,136,55}

\definecolor{PRGn4M}{RGB}{0,136,55}

\definecolor{PRGn51}{RGB}{123,50,148}

\definecolor{PRGn5C}{RGB}{123,50,148}

\definecolor{PRGn52}{RGB}{194,165,207}

\definecolor{PRGn5F}{RGB}{194,165,207}

\definecolor{PRGn53}{RGB}{247,247,247}

\definecolor{PRGn5H}{RGB}{247,247,247}

\definecolor{PRGn54}{RGB}{166,219,160}

\definecolor{PRGn5J}{RGB}{166,219,160}

\definecolor{PRGn55}{RGB}{0,136,55}

\definecolor{PRGn5M}{RGB}{0,136,55}

\definecolor{PRGn61}{RGB}{118,42,131}

\definecolor{PRGn6B}{RGB}{118,42,131}

\definecolor{PRGn62}{RGB}{175,141,195}

\definecolor{PRGn6E}{RGB}{175,141,195}

\definecolor{PRGn63}{RGB}{231,212,232}

\definecolor{PRGn6G}{RGB}{231,212,232}

\definecolor{PRGn64}{RGB}{217,240,211}

\definecolor{PRGn6I}{RGB}{217,240,211}

\definecolor{PRGn65}{RGB}{127,191,123}

\definecolor{PRGn6K}{RGB}{127,191,123}

\definecolor{PRGn66}{RGB}{27,120,55}

\definecolor{PRGn6N}{RGB}{27,120,55}

\definecolor{PRGn71}{RGB}{118,42,131}

\definecolor{PRGn7B}{RGB}{118,42,131}

\definecolor{PRGn72}{RGB}{175,141,195}

\definecolor{PRGn7E}{RGB}{175,141,195}

\definecolor{PRGn73}{RGB}{231,212,232}

\definecolor{PRGn7G}{RGB}{231,212,232}

\definecolor{PRGn74}{RGB}{247,247,247}

\definecolor{PRGn7H}{RGB}{247,247,247}

\definecolor{PRGn75}{RGB}{217,240,211}

\definecolor{PRGn7I}{RGB}{217,240,211}

\definecolor{PRGn76}{RGB}{127,191,123}

\definecolor{PRGn7K}{RGB}{127,191,123}

\definecolor{PRGn77}{RGB}{27,120,55}

\definecolor{PRGn7N}{RGB}{27,120,55}

\definecolor{PRGn81}{RGB}{118,42,131}

\definecolor{PRGn8B}{RGB}{118,42,131}

\definecolor{PRGn82}{RGB}{153,112,171}

\definecolor{PRGn8D}{RGB}{153,112,171}

\definecolor{PRGn83}{RGB}{194,165,207}

\definecolor{PRGn8F}{RGB}{194,165,207}

\definecolor{PRGn84}{RGB}{231,212,232}

\definecolor{PRGn8G}{RGB}{231,212,232}

\definecolor{PRGn85}{RGB}{217,240,211}

\definecolor{PRGn8I}{RGB}{217,240,211}

\definecolor{PRGn86}{RGB}{166,219,160}

\definecolor{PRGn8J}{RGB}{166,219,160}

\definecolor{PRGn87}{RGB}{90,174,97}

\definecolor{PRGn8L}{RGB}{90,174,97}

\definecolor{PRGn88}{RGB}{27,120,55}

\definecolor{PRGn8N}{RGB}{27,120,55}

\definecolor{PRGn91}{RGB}{118,42,131}

\definecolor{PRGn9B}{RGB}{118,42,131}

\definecolor{PRGn92}{RGB}{153,112,171}

\definecolor{PRGn9D}{RGB}{153,112,171}

\definecolor{PRGn93}{RGB}{194,165,207}

\definecolor{PRGn9F}{RGB}{194,165,207}

\definecolor{PRGn94}{RGB}{231,212,232}

\definecolor{PRGn9G}{RGB}{231,212,232}

\definecolor{PRGn95}{RGB}{247,247,247}

\definecolor{PRGn9H}{RGB}{247,247,247}

\definecolor{PRGn96}{RGB}{217,240,211}

\definecolor{PRGn9I}{RGB}{217,240,211}

\definecolor{PRGn97}{RGB}{166,219,160}

\definecolor{PRGn9J}{RGB}{166,219,160}

\definecolor{PRGn98}{RGB}{90,174,97}

\definecolor{PRGn9L}{RGB}{90,174,97}

\definecolor{PRGn99}{RGB}{27,120,55}

\definecolor{PRGn9N}{RGB}{27,120,55}

\definecolor{PRGn101}{RGB}{64,0,75}

\definecolor{PRGn10A}{RGB}{64,0,75}

\definecolor{PRGn102}{RGB}{118,42,131}

\definecolor{PRGn10B}{RGB}{118,42,131}

\definecolor{PRGn103}{RGB}{153,112,171}

\definecolor{PRGn10D}{RGB}{153,112,171}

\definecolor{PRGn104}{RGB}{194,165,207}

\definecolor{PRGn10F}{RGB}{194,165,207}

\definecolor{PRGn105}{RGB}{231,212,232}

\definecolor{PRGn10G}{RGB}{231,212,232}

\definecolor{PRGn106}{RGB}{217,240,211}

\definecolor{PRGn10I}{RGB}{217,240,211}

\definecolor{PRGn107}{RGB}{166,219,160}

\definecolor{PRGn10J}{RGB}{166,219,160}

\definecolor{PRGn108}{RGB}{90,174,97}

\definecolor{PRGn10L}{RGB}{90,174,97}

\definecolor{PRGn109}{RGB}{27,120,55}

\definecolor{PRGn10N}{RGB}{27,120,55}

\definecolor{PRGn1010}{RGB}{0,68,27}

\definecolor{PRGn10O}{RGB}{0,68,27}

\definecolor{PRGn111}{RGB}{64,0,75}

\definecolor{PRGn11A}{RGB}{64,0,75}

\definecolor{PRGn112}{RGB}{118,42,131}

\definecolor{PRGn11B}{RGB}{118,42,131}

\definecolor{PRGn113}{RGB}{153,112,171}

\definecolor{PRGn11D}{RGB}{153,112,171}

\definecolor{PRGn114}{RGB}{194,165,207}

\definecolor{PRGn11F}{RGB}{194,165,207}

\definecolor{PRGn115}{RGB}{231,212,232}

\definecolor{PRGn11G}{RGB}{231,212,232}

\definecolor{PRGn116}{RGB}{247,247,247}

\definecolor{PRGn11H}{RGB}{247,247,247}

\definecolor{PRGn117}{RGB}{217,240,211}

\definecolor{PRGn11I}{RGB}{217,240,211}

\definecolor{PRGn118}{RGB}{166,219,160}

\definecolor{PRGn11J}{RGB}{166,219,160}

\definecolor{PRGn119}{RGB}{90,174,97}

\definecolor{PRGn11L}{RGB}{90,174,97}

\definecolor{PRGn1110}{RGB}{27,120,55}

\definecolor{PRGn11N}{RGB}{27,120,55}

\definecolor{PRGn1111}{RGB}{0,68,27}

\definecolor{PRGn11O}{RGB}{0,68,27}

\definecolor{PiYG31}{RGB}{233,163,201}

\definecolor{PiYG3E}{RGB}{233,163,201}

\definecolor{PiYG32}{RGB}{247,247,247}

\definecolor{PiYG3H}{RGB}{247,247,247}

\definecolor{PiYG33}{RGB}{161,215,106}

\definecolor{PiYG3K}{RGB}{161,215,106}

\definecolor{PiYG41}{RGB}{208,28,139}

\definecolor{PiYG4C}{RGB}{208,28,139}

\definecolor{PiYG42}{RGB}{241,182,218}

\definecolor{PiYG4F}{RGB}{241,182,218}

\definecolor{PiYG43}{RGB}{184,225,134}

\definecolor{PiYG4J}{RGB}{184,225,134}

\definecolor{PiYG44}{RGB}{77,172,38}

\definecolor{PiYG4M}{RGB}{77,172,38}

\definecolor{PiYG51}{RGB}{208,28,139}

\definecolor{PiYG5C}{RGB}{208,28,139}

\definecolor{PiYG52}{RGB}{241,182,218}

\definecolor{PiYG5F}{RGB}{241,182,218}

\definecolor{PiYG53}{RGB}{247,247,247}

\definecolor{PiYG5H}{RGB}{247,247,247}

\definecolor{PiYG54}{RGB}{184,225,134}

\definecolor{PiYG5J}{RGB}{184,225,134}

\definecolor{PiYG55}{RGB}{77,172,38}

\definecolor{PiYG5M}{RGB}{77,172,38}

\definecolor{PiYG61}{RGB}{197,27,125}

\definecolor{PiYG6B}{RGB}{197,27,125}

\definecolor{PiYG62}{RGB}{233,163,201}

\definecolor{PiYG6E}{RGB}{233,163,201}

\definecolor{PiYG63}{RGB}{253,224,239}

\definecolor{PiYG6G}{RGB}{253,224,239}

\definecolor{PiYG64}{RGB}{230,245,208}

\definecolor{PiYG6I}{RGB}{230,245,208}

\definecolor{PiYG65}{RGB}{161,215,106}

\definecolor{PiYG6K}{RGB}{161,215,106}

\definecolor{PiYG66}{RGB}{77,146,33}

\definecolor{PiYG6N}{RGB}{77,146,33}

\definecolor{PiYG71}{RGB}{197,27,125}

\definecolor{PiYG7B}{RGB}{197,27,125}

\definecolor{PiYG72}{RGB}{233,163,201}

\definecolor{PiYG7E}{RGB}{233,163,201}

\definecolor{PiYG73}{RGB}{253,224,239}

\definecolor{PiYG7G}{RGB}{253,224,239}

\definecolor{PiYG74}{RGB}{247,247,247}

\definecolor{PiYG7H}{RGB}{247,247,247}

\definecolor{PiYG75}{RGB}{230,245,208}

\definecolor{PiYG7I}{RGB}{230,245,208}

\definecolor{PiYG76}{RGB}{161,215,106}

\definecolor{PiYG7K}{RGB}{161,215,106}

\definecolor{PiYG77}{RGB}{77,146,33}

\definecolor{PiYG7N}{RGB}{77,146,33}

\definecolor{PiYG81}{RGB}{197,27,125}

\definecolor{PiYG8B}{RGB}{197,27,125}

\definecolor{PiYG82}{RGB}{222,119,174}

\definecolor{PiYG8D}{RGB}{222,119,174}

\definecolor{PiYG83}{RGB}{241,182,218}

\definecolor{PiYG8F}{RGB}{241,182,218}

\definecolor{PiYG84}{RGB}{253,224,239}

\definecolor{PiYG8G}{RGB}{253,224,239}

\definecolor{PiYG85}{RGB}{230,245,208}

\definecolor{PiYG8I}{RGB}{230,245,208}

\definecolor{PiYG86}{RGB}{184,225,134}

\definecolor{PiYG8J}{RGB}{184,225,134}

\definecolor{PiYG87}{RGB}{127,188,65}

\definecolor{PiYG8L}{RGB}{127,188,65}

\definecolor{PiYG88}{RGB}{77,146,33}

\definecolor{PiYG8N}{RGB}{77,146,33}

\definecolor{PiYG91}{RGB}{197,27,125}

\definecolor{PiYG9B}{RGB}{197,27,125}

\definecolor{PiYG92}{RGB}{222,119,174}

\definecolor{PiYG9D}{RGB}{222,119,174}

\definecolor{PiYG93}{RGB}{241,182,218}

\definecolor{PiYG9F}{RGB}{241,182,218}

\definecolor{PiYG94}{RGB}{253,224,239}

\definecolor{PiYG9G}{RGB}{253,224,239}

\definecolor{PiYG95}{RGB}{247,247,247}

\definecolor{PiYG9H}{RGB}{247,247,247}

\definecolor{PiYG96}{RGB}{230,245,208}

\definecolor{PiYG9I}{RGB}{230,245,208}

\definecolor{PiYG97}{RGB}{184,225,134}

\definecolor{PiYG9J}{RGB}{184,225,134}

\definecolor{PiYG98}{RGB}{127,188,65}

\definecolor{PiYG9L}{RGB}{127,188,65}

\definecolor{PiYG99}{RGB}{77,146,33}

\definecolor{PiYG9N}{RGB}{77,146,33}

\definecolor{PiYG101}{RGB}{142,1,82}

\definecolor{PiYG10A}{RGB}{142,1,82}

\definecolor{PiYG102}{RGB}{197,27,125}

\definecolor{PiYG10B}{RGB}{197,27,125}

\definecolor{PiYG103}{RGB}{222,119,174}

\definecolor{PiYG10D}{RGB}{222,119,174}

\definecolor{PiYG104}{RGB}{241,182,218}

\definecolor{PiYG10F}{RGB}{241,182,218}

\definecolor{PiYG105}{RGB}{253,224,239}

\definecolor{PiYG10G}{RGB}{253,224,239}

\definecolor{PiYG106}{RGB}{230,245,208}

\definecolor{PiYG10I}{RGB}{230,245,208}

\definecolor{PiYG107}{RGB}{184,225,134}

\definecolor{PiYG10J}{RGB}{184,225,134}

\definecolor{PiYG108}{RGB}{127,188,65}

\definecolor{PiYG10L}{RGB}{127,188,65}

\definecolor{PiYG109}{RGB}{77,146,33}

\definecolor{PiYG10N}{RGB}{77,146,33}

\definecolor{PiYG1010}{RGB}{39,100,25}

\definecolor{PiYG10O}{RGB}{39,100,25}

\definecolor{PiYG111}{RGB}{142,1,82}

\definecolor{PiYG11A}{RGB}{142,1,82}

\definecolor{PiYG112}{RGB}{197,27,125}

\definecolor{PiYG11B}{RGB}{197,27,125}

\definecolor{PiYG113}{RGB}{222,119,174}

\definecolor{PiYG11D}{RGB}{222,119,174}

\definecolor{PiYG114}{RGB}{241,182,218}

\definecolor{PiYG11F}{RGB}{241,182,218}

\definecolor{PiYG115}{RGB}{253,224,239}

\definecolor{PiYG11G}{RGB}{253,224,239}

\definecolor{PiYG116}{RGB}{247,247,247}

\definecolor{PiYG11H}{RGB}{247,247,247}

\definecolor{PiYG117}{RGB}{230,245,208}

\definecolor{PiYG11I}{RGB}{230,245,208}

\definecolor{PiYG118}{RGB}{184,225,134}

\definecolor{PiYG11J}{RGB}{184,225,134}

\definecolor{PiYG119}{RGB}{127,188,65}

\definecolor{PiYG11L}{RGB}{127,188,65}

\definecolor{PiYG1110}{RGB}{77,146,33}

\definecolor{PiYG11N}{RGB}{77,146,33}

\definecolor{PiYG1111}{RGB}{39,100,25}

\definecolor{PiYG11O}{RGB}{39,100,25}

\definecolor{RdBu31}{RGB}{239,138,98}

\definecolor{RdBu3E}{RGB}{239,138,98}

\definecolor{RdBu32}{RGB}{247,247,247}

\definecolor{RdBu3H}{RGB}{247,247,247}

\definecolor{RdBu33}{RGB}{103,169,207}

\definecolor{RdBu3K}{RGB}{103,169,207}

\definecolor{RdBu41}{RGB}{202,0,32}

\definecolor{RdBu4C}{RGB}{202,0,32}

\definecolor{RdBu42}{RGB}{244,165,130}

\definecolor{RdBu4F}{RGB}{244,165,130}

\definecolor{RdBu43}{RGB}{146,197,222}

\definecolor{RdBu4J}{RGB}{146,197,222}

\definecolor{RdBu44}{RGB}{5,113,176}

\definecolor{RdBu4M}{RGB}{5,113,176}

\definecolor{RdBu51}{RGB}{202,0,32}

\definecolor{RdBu5C}{RGB}{202,0,32}

\definecolor{RdBu52}{RGB}{244,165,130}

\definecolor{RdBu5F}{RGB}{244,165,130}

\definecolor{RdBu53}{RGB}{247,247,247}

\definecolor{RdBu5H}{RGB}{247,247,247}

\definecolor{RdBu54}{RGB}{146,197,222}

\definecolor{RdBu5J}{RGB}{146,197,222}

\definecolor{RdBu55}{RGB}{5,113,176}

\definecolor{RdBu5M}{RGB}{5,113,176}

\definecolor{RdBu61}{RGB}{178,24,43}

\definecolor{RdBu6B}{RGB}{178,24,43}

\definecolor{RdBu62}{RGB}{239,138,98}

\definecolor{RdBu6E}{RGB}{239,138,98}

\definecolor{RdBu63}{RGB}{253,219,199}

\definecolor{RdBu6G}{RGB}{253,219,199}

\definecolor{RdBu64}{RGB}{209,229,240}

\definecolor{RdBu6I}{RGB}{209,229,240}

\definecolor{RdBu65}{RGB}{103,169,207}

\definecolor{RdBu6K}{RGB}{103,169,207}

\definecolor{RdBu66}{RGB}{33,102,172}

\definecolor{RdBu6N}{RGB}{33,102,172}

\definecolor{RdBu71}{RGB}{178,24,43}

\definecolor{RdBu7B}{RGB}{178,24,43}

\definecolor{RdBu72}{RGB}{239,138,98}

\definecolor{RdBu7E}{RGB}{239,138,98}

\definecolor{RdBu73}{RGB}{253,219,199}

\definecolor{RdBu7G}{RGB}{253,219,199}

\definecolor{RdBu74}{RGB}{247,247,247}

\definecolor{RdBu7H}{RGB}{247,247,247}

\definecolor{RdBu75}{RGB}{209,229,240}

\definecolor{RdBu7I}{RGB}{209,229,240}

\definecolor{RdBu76}{RGB}{103,169,207}

\definecolor{RdBu7K}{RGB}{103,169,207}

\definecolor{RdBu77}{RGB}{33,102,172}

\definecolor{RdBu7N}{RGB}{33,102,172}

\definecolor{RdBu81}{RGB}{178,24,43}

\definecolor{RdBu8B}{RGB}{178,24,43}

\definecolor{RdBu82}{RGB}{214,96,77}

\definecolor{RdBu8D}{RGB}{214,96,77}

\definecolor{RdBu83}{RGB}{244,165,130}

\definecolor{RdBu8F}{RGB}{244,165,130}

\definecolor{RdBu84}{RGB}{253,219,199}

\definecolor{RdBu8G}{RGB}{253,219,199}

\definecolor{RdBu85}{RGB}{209,229,240}

\definecolor{RdBu8I}{RGB}{209,229,240}

\definecolor{RdBu86}{RGB}{146,197,222}

\definecolor{RdBu8J}{RGB}{146,197,222}

\definecolor{RdBu87}{RGB}{67,147,195}

\definecolor{RdBu8L}{RGB}{67,147,195}

\definecolor{RdBu88}{RGB}{33,102,172}

\definecolor{RdBu8N}{RGB}{33,102,172}

\definecolor{RdBu91}{RGB}{178,24,43}

\definecolor{RdBu9B}{RGB}{178,24,43}

\definecolor{RdBu92}{RGB}{214,96,77}

\definecolor{RdBu9D}{RGB}{214,96,77}

\definecolor{RdBu93}{RGB}{244,165,130}

\definecolor{RdBu9F}{RGB}{244,165,130}

\definecolor{RdBu94}{RGB}{253,219,199}

\definecolor{RdBu9G}{RGB}{253,219,199}

\definecolor{RdBu95}{RGB}{247,247,247}

\definecolor{RdBu9H}{RGB}{247,247,247}

\definecolor{RdBu96}{RGB}{209,229,240}

\definecolor{RdBu9I}{RGB}{209,229,240}

\definecolor{RdBu97}{RGB}{146,197,222}

\definecolor{RdBu9J}{RGB}{146,197,222}

\definecolor{RdBu98}{RGB}{67,147,195}

\definecolor{RdBu9L}{RGB}{67,147,195}

\definecolor{RdBu99}{RGB}{33,102,172}

\definecolor{RdBu9N}{RGB}{33,102,172}

\definecolor{RdBu101}{RGB}{103,0,31}

\definecolor{RdBu10A}{RGB}{103,0,31}

\definecolor{RdBu102}{RGB}{178,24,43}

\definecolor{RdBu10B}{RGB}{178,24,43}

\definecolor{RdBu103}{RGB}{214,96,77}

\definecolor{RdBu10D}{RGB}{214,96,77}

\definecolor{RdBu104}{RGB}{244,165,130}

\definecolor{RdBu10F}{RGB}{244,165,130}

\definecolor{RdBu105}{RGB}{253,219,199}

\definecolor{RdBu10G}{RGB}{253,219,199}

\definecolor{RdBu106}{RGB}{209,229,240}

\definecolor{RdBu10I}{RGB}{209,229,240}

\definecolor{RdBu107}{RGB}{146,197,222}

\definecolor{RdBu10J}{RGB}{146,197,222}

\definecolor{RdBu108}{RGB}{67,147,195}

\definecolor{RdBu10L}{RGB}{67,147,195}

\definecolor{RdBu109}{RGB}{33,102,172}

\definecolor{RdBu10N}{RGB}{33,102,172}

\definecolor{RdBu1010}{RGB}{5,48,97}

\definecolor{RdBu10O}{RGB}{5,48,97}

\definecolor{RdBu111}{RGB}{103,0,31}

\definecolor{RdBu11A}{RGB}{103,0,31}

\definecolor{RdBu112}{RGB}{178,24,43}

\definecolor{RdBu11B}{RGB}{178,24,43}

\definecolor{RdBu113}{RGB}{214,96,77}

\definecolor{RdBu11D}{RGB}{214,96,77}

\definecolor{RdBu114}{RGB}{244,165,130}

\definecolor{RdBu11F}{RGB}{244,165,130}

\definecolor{RdBu115}{RGB}{253,219,199}

\definecolor{RdBu11G}{RGB}{253,219,199}

\definecolor{RdBu116}{RGB}{247,247,247}

\definecolor{RdBu11H}{RGB}{247,247,247}

\definecolor{RdBu117}{RGB}{209,229,240}

\definecolor{RdBu11I}{RGB}{209,229,240}

\definecolor{RdBu118}{RGB}{146,197,222}

\definecolor{RdBu11J}{RGB}{146,197,222}

\definecolor{RdBu119}{RGB}{67,147,195}

\definecolor{RdBu11L}{RGB}{67,147,195}

\definecolor{RdBu1110}{RGB}{33,102,172}

\definecolor{RdBu11N}{RGB}{33,102,172}

\definecolor{RdBu1111}{RGB}{5,48,97}

\definecolor{RdBu11O}{RGB}{5,48,97}

\definecolor{RdGy31}{RGB}{239,138,98}

\definecolor{RdGy3E}{RGB}{239,138,98}

\definecolor{RdGy32}{RGB}{255,255,255}

\definecolor{RdGy3H}{RGB}{255,255,255}

\definecolor{RdGy33}{RGB}{153,153,153}

\definecolor{RdGy3K}{RGB}{153,153,153}

\definecolor{RdGy41}{RGB}{202,0,32}

\definecolor{RdGy4C}{RGB}{202,0,32}

\definecolor{RdGy42}{RGB}{244,165,130}

\definecolor{RdGy4F}{RGB}{244,165,130}

\definecolor{RdGy43}{RGB}{186,186,186}

\definecolor{RdGy4J}{RGB}{186,186,186}

\definecolor{RdGy44}{RGB}{64,64,64}

\definecolor{RdGy4M}{RGB}{64,64,64}

\definecolor{RdGy51}{RGB}{202,0,32}

\definecolor{RdGy5C}{RGB}{202,0,32}

\definecolor{RdGy52}{RGB}{244,165,130}

\definecolor{RdGy5F}{RGB}{244,165,130}

\definecolor{RdGy53}{RGB}{255,255,255}

\definecolor{RdGy5H}{RGB}{255,255,255}

\definecolor{RdGy54}{RGB}{186,186,186}

\definecolor{RdGy5J}{RGB}{186,186,186}

\definecolor{RdGy55}{RGB}{64,64,64}

\definecolor{RdGy5M}{RGB}{64,64,64}

\definecolor{RdGy61}{RGB}{178,24,43}

\definecolor{RdGy6B}{RGB}{178,24,43}

\definecolor{RdGy62}{RGB}{239,138,98}

\definecolor{RdGy6E}{RGB}{239,138,98}

\definecolor{RdGy63}{RGB}{253,219,199}

\definecolor{RdGy6G}{RGB}{253,219,199}

\definecolor{RdGy64}{RGB}{224,224,224}

\definecolor{RdGy6I}{RGB}{224,224,224}

\definecolor{RdGy65}{RGB}{153,153,153}

\definecolor{RdGy6K}{RGB}{153,153,153}

\definecolor{RdGy66}{RGB}{77,77,77}

\definecolor{RdGy6N}{RGB}{77,77,77}

\definecolor{RdGy71}{RGB}{178,24,43}

\definecolor{RdGy7B}{RGB}{178,24,43}

\definecolor{RdGy72}{RGB}{239,138,98}

\definecolor{RdGy7E}{RGB}{239,138,98}

\definecolor{RdGy73}{RGB}{253,219,199}

\definecolor{RdGy7G}{RGB}{253,219,199}

\definecolor{RdGy74}{RGB}{255,255,255}

\definecolor{RdGy7H}{RGB}{255,255,255}

\definecolor{RdGy75}{RGB}{224,224,224}

\definecolor{RdGy7I}{RGB}{224,224,224}

\definecolor{RdGy76}{RGB}{153,153,153}

\definecolor{RdGy7K}{RGB}{153,153,153}

\definecolor{RdGy77}{RGB}{77,77,77}

\definecolor{RdGy7N}{RGB}{77,77,77}

\definecolor{RdGy81}{RGB}{178,24,43}

\definecolor{RdGy8B}{RGB}{178,24,43}

\definecolor{RdGy82}{RGB}{214,96,77}

\definecolor{RdGy8D}{RGB}{214,96,77}

\definecolor{RdGy83}{RGB}{244,165,130}

\definecolor{RdGy8F}{RGB}{244,165,130}

\definecolor{RdGy84}{RGB}{253,219,199}

\definecolor{RdGy8G}{RGB}{253,219,199}

\definecolor{RdGy85}{RGB}{224,224,224}

\definecolor{RdGy8I}{RGB}{224,224,224}

\definecolor{RdGy86}{RGB}{186,186,186}

\definecolor{RdGy8J}{RGB}{186,186,186}

\definecolor{RdGy87}{RGB}{135,135,135}

\definecolor{RdGy8L}{RGB}{135,135,135}

\definecolor{RdGy88}{RGB}{77,77,77}

\definecolor{RdGy8N}{RGB}{77,77,77}

\definecolor{RdGy91}{RGB}{178,24,43}

\definecolor{RdGy9B}{RGB}{178,24,43}

\definecolor{RdGy92}{RGB}{214,96,77}

\definecolor{RdGy9D}{RGB}{214,96,77}

\definecolor{RdGy93}{RGB}{244,165,130}

\definecolor{RdGy9F}{RGB}{244,165,130}

\definecolor{RdGy94}{RGB}{253,219,199}

\definecolor{RdGy9G}{RGB}{253,219,199}

\definecolor{RdGy95}{RGB}{255,255,255}

\definecolor{RdGy9H}{RGB}{255,255,255}

\definecolor{RdGy96}{RGB}{224,224,224}

\definecolor{RdGy9I}{RGB}{224,224,224}

\definecolor{RdGy97}{RGB}{186,186,186}

\definecolor{RdGy9J}{RGB}{186,186,186}

\definecolor{RdGy98}{RGB}{135,135,135}

\definecolor{RdGy9L}{RGB}{135,135,135}

\definecolor{RdGy99}{RGB}{77,77,77}

\definecolor{RdGy9N}{RGB}{77,77,77}

\definecolor{RdGy101}{RGB}{103,0,31}

\definecolor{RdGy10A}{RGB}{103,0,31}

\definecolor{RdGy102}{RGB}{178,24,43}

\definecolor{RdGy10B}{RGB}{178,24,43}

\definecolor{RdGy103}{RGB}{214,96,77}

\definecolor{RdGy10D}{RGB}{214,96,77}

\definecolor{RdGy104}{RGB}{244,165,130}

\definecolor{RdGy10F}{RGB}{244,165,130}

\definecolor{RdGy105}{RGB}{253,219,199}

\definecolor{RdGy10G}{RGB}{253,219,199}

\definecolor{RdGy106}{RGB}{224,224,224}

\definecolor{RdGy10I}{RGB}{224,224,224}

\definecolor{RdGy107}{RGB}{186,186,186}

\definecolor{RdGy10J}{RGB}{186,186,186}

\definecolor{RdGy108}{RGB}{135,135,135}

\definecolor{RdGy10L}{RGB}{135,135,135}

\definecolor{RdGy109}{RGB}{77,77,77}

\definecolor{RdGy10N}{RGB}{77,77,77}

\definecolor{RdGy1010}{RGB}{26,26,26}

\definecolor{RdGy10O}{RGB}{26,26,26}

\definecolor{RdGy111}{RGB}{103,0,31}

\definecolor{RdGy11A}{RGB}{103,0,31}

\definecolor{RdGy112}{RGB}{178,24,43}

\definecolor{RdGy11B}{RGB}{178,24,43}

\definecolor{RdGy113}{RGB}{214,96,77}

\definecolor{RdGy11D}{RGB}{214,96,77}

\definecolor{RdGy114}{RGB}{244,165,130}

\definecolor{RdGy11F}{RGB}{244,165,130}

\definecolor{RdGy115}{RGB}{253,219,199}

\definecolor{RdGy11G}{RGB}{253,219,199}

\definecolor{RdGy116}{RGB}{255,255,255}

\definecolor{RdGy11H}{RGB}{255,255,255}

\definecolor{RdGy117}{RGB}{224,224,224}

\definecolor{RdGy11I}{RGB}{224,224,224}

\definecolor{RdGy118}{RGB}{186,186,186}

\definecolor{RdGy11J}{RGB}{186,186,186}

\definecolor{RdGy119}{RGB}{135,135,135}

\definecolor{RdGy11L}{RGB}{135,135,135}

\definecolor{RdGy1110}{RGB}{77,77,77}

\definecolor{RdGy11N}{RGB}{77,77,77}

\definecolor{RdGy1111}{RGB}{26,26,26}

\definecolor{RdGy11O}{RGB}{26,26,26}

\definecolor{RdYlBu31}{RGB}{252,141,89}

\definecolor{RdYlBu3E}{RGB}{252,141,89}

\definecolor{RdYlBu32}{RGB}{255,255,191}

\definecolor{RdYlBu3H}{RGB}{255,255,191}

\definecolor{RdYlBu33}{RGB}{145,191,219}

\definecolor{RdYlBu3K}{RGB}{145,191,219}

\definecolor{RdYlBu41}{RGB}{215,25,28}

\definecolor{RdYlBu4C}{RGB}{215,25,28}

\definecolor{RdYlBu42}{RGB}{253,174,97}

\definecolor{RdYlBu4F}{RGB}{253,174,97}

\definecolor{RdYlBu43}{RGB}{171,217,233}

\definecolor{RdYlBu4J}{RGB}{171,217,233}

\definecolor{RdYlBu44}{RGB}{44,123,182}

\definecolor{RdYlBu4M}{RGB}{44,123,182}

\definecolor{RdYlBu51}{RGB}{215,25,28}

\definecolor{RdYlBu5C}{RGB}{215,25,28}

\definecolor{RdYlBu52}{RGB}{253,174,97}

\definecolor{RdYlBu5F}{RGB}{253,174,97}

\definecolor{RdYlBu53}{RGB}{255,255,191}

\definecolor{RdYlBu5H}{RGB}{255,255,191}

\definecolor{RdYlBu54}{RGB}{171,217,233}

\definecolor{RdYlBu5J}{RGB}{171,217,233}

\definecolor{RdYlBu55}{RGB}{44,123,182}

\definecolor{RdYlBu5M}{RGB}{44,123,182}

\definecolor{RdYlBu61}{RGB}{215,48,39}

\definecolor{RdYlBu6B}{RGB}{215,48,39}

\definecolor{RdYlBu62}{RGB}{252,141,89}

\definecolor{RdYlBu6E}{RGB}{252,141,89}

\definecolor{RdYlBu63}{RGB}{254,224,144}

\definecolor{RdYlBu6G}{RGB}{254,224,144}

\definecolor{RdYlBu64}{RGB}{224,243,248}

\definecolor{RdYlBu6I}{RGB}{224,243,248}

\definecolor{RdYlBu65}{RGB}{145,191,219}

\definecolor{RdYlBu6K}{RGB}{145,191,219}

\definecolor{RdYlBu66}{RGB}{69,117,180}

\definecolor{RdYlBu6N}{RGB}{69,117,180}

\definecolor{RdYlBu71}{RGB}{215,48,39}

\definecolor{RdYlBu7B}{RGB}{215,48,39}

\definecolor{RdYlBu72}{RGB}{252,141,89}

\definecolor{RdYlBu7E}{RGB}{252,141,89}

\definecolor{RdYlBu73}{RGB}{254,224,144}

\definecolor{RdYlBu7G}{RGB}{254,224,144}

\definecolor{RdYlBu74}{RGB}{255,255,191}

\definecolor{RdYlBu7H}{RGB}{255,255,191}

\definecolor{RdYlBu75}{RGB}{224,243,248}

\definecolor{RdYlBu7I}{RGB}{224,243,248}

\definecolor{RdYlBu76}{RGB}{145,191,219}

\definecolor{RdYlBu7K}{RGB}{145,191,219}

\definecolor{RdYlBu77}{RGB}{69,117,180}

\definecolor{RdYlBu7N}{RGB}{69,117,180}

\definecolor{RdYlBu81}{RGB}{215,48,39}

\definecolor{RdYlBu8B}{RGB}{215,48,39}

\definecolor{RdYlBu82}{RGB}{244,109,67}

\definecolor{RdYlBu8D}{RGB}{244,109,67}

\definecolor{RdYlBu83}{RGB}{253,174,97}

\definecolor{RdYlBu8F}{RGB}{253,174,97}

\definecolor{RdYlBu84}{RGB}{254,224,144}

\definecolor{RdYlBu8G}{RGB}{254,224,144}

\definecolor{RdYlBu85}{RGB}{224,243,248}

\definecolor{RdYlBu8I}{RGB}{224,243,248}

\definecolor{RdYlBu86}{RGB}{171,217,233}

\definecolor{RdYlBu8J}{RGB}{171,217,233}

\definecolor{RdYlBu87}{RGB}{116,173,209}

\definecolor{RdYlBu8L}{RGB}{116,173,209}

\definecolor{RdYlBu88}{RGB}{69,117,180}

\definecolor{RdYlBu8N}{RGB}{69,117,180}

\definecolor{RdYlBu91}{RGB}{215,48,39}

\definecolor{RdYlBu9B}{RGB}{215,48,39}

\definecolor{RdYlBu92}{RGB}{244,109,67}

\definecolor{RdYlBu9D}{RGB}{244,109,67}

\definecolor{RdYlBu93}{RGB}{253,174,97}

\definecolor{RdYlBu9F}{RGB}{253,174,97}

\definecolor{RdYlBu94}{RGB}{254,224,144}

\definecolor{RdYlBu9G}{RGB}{254,224,144}

\definecolor{RdYlBu95}{RGB}{255,255,191}

\definecolor{RdYlBu9H}{RGB}{255,255,191}

\definecolor{RdYlBu96}{RGB}{224,243,248}

\definecolor{RdYlBu9I}{RGB}{224,243,248}

\definecolor{RdYlBu97}{RGB}{171,217,233}

\definecolor{RdYlBu9J}{RGB}{171,217,233}

\definecolor{RdYlBu98}{RGB}{116,173,209}

\definecolor{RdYlBu9L}{RGB}{116,173,209}

\definecolor{RdYlBu99}{RGB}{69,117,180}

\definecolor{RdYlBu9N}{RGB}{69,117,180}

\definecolor{RdYlBu101}{RGB}{165,0,38}

\definecolor{RdYlBu10A}{RGB}{165,0,38}

\definecolor{RdYlBu102}{RGB}{215,48,39}

\definecolor{RdYlBu10B}{RGB}{215,48,39}

\definecolor{RdYlBu103}{RGB}{244,109,67}

\definecolor{RdYlBu10D}{RGB}{244,109,67}

\definecolor{RdYlBu104}{RGB}{253,174,97}

\definecolor{RdYlBu10F}{RGB}{253,174,97}

\definecolor{RdYlBu105}{RGB}{254,224,144}

\definecolor{RdYlBu10G}{RGB}{254,224,144}

\definecolor{RdYlBu106}{RGB}{224,243,248}

\definecolor{RdYlBu10I}{RGB}{224,243,248}

\definecolor{RdYlBu107}{RGB}{171,217,233}

\definecolor{RdYlBu10J}{RGB}{171,217,233}

\definecolor{RdYlBu108}{RGB}{116,173,209}

\definecolor{RdYlBu10L}{RGB}{116,173,209}

\definecolor{RdYlBu109}{RGB}{69,117,180}

\definecolor{RdYlBu10N}{RGB}{69,117,180}

\definecolor{RdYlBu1010}{RGB}{49,54,149}

\definecolor{RdYlBu10O}{RGB}{49,54,149}

\definecolor{RdYlBu111}{RGB}{165,0,38}

\definecolor{RdYlBu11A}{RGB}{165,0,38}

\definecolor{RdYlBu112}{RGB}{215,48,39}

\definecolor{RdYlBu11B}{RGB}{215,48,39}

\definecolor{RdYlBu113}{RGB}{244,109,67}

\definecolor{RdYlBu11D}{RGB}{244,109,67}

\definecolor{RdYlBu114}{RGB}{253,174,97}

\definecolor{RdYlBu11F}{RGB}{253,174,97}

\definecolor{RdYlBu115}{RGB}{254,224,144}

\definecolor{RdYlBu11G}{RGB}{254,224,144}

\definecolor{RdYlBu116}{RGB}{255,255,191}

\definecolor{RdYlBu11H}{RGB}{255,255,191}

\definecolor{RdYlBu117}{RGB}{224,243,248}

\definecolor{RdYlBu11I}{RGB}{224,243,248}

\definecolor{RdYlBu118}{RGB}{171,217,233}

\definecolor{RdYlBu11J}{RGB}{171,217,233}

\definecolor{RdYlBu119}{RGB}{116,173,209}

\definecolor{RdYlBu11L}{RGB}{116,173,209}

\definecolor{RdYlBu1110}{RGB}{69,117,180}

\definecolor{RdYlBu11N}{RGB}{69,117,180}

\definecolor{RdYlBu1111}{RGB}{49,54,149}

\definecolor{RdYlBu11O}{RGB}{49,54,149}

\definecolor{Spectral31}{RGB}{252,141,89}

\definecolor{Spectral3E}{RGB}{252,141,89}

\definecolor{Spectral32}{RGB}{255,255,191}

\definecolor{Spectral3H}{RGB}{255,255,191}

\definecolor{Spectral33}{RGB}{153,213,148}

\definecolor{Spectral3K}{RGB}{153,213,148}

\definecolor{Spectral41}{RGB}{215,25,28}

\definecolor{Spectral4C}{RGB}{215,25,28}

\definecolor{Spectral42}{RGB}{253,174,97}

\definecolor{Spectral4F}{RGB}{253,174,97}

\definecolor{Spectral43}{RGB}{171,221,164}

\definecolor{Spectral4J}{RGB}{171,221,164}

\definecolor{Spectral44}{RGB}{43,131,186}

\definecolor{Spectral4M}{RGB}{43,131,186}

\definecolor{Spectral51}{RGB}{215,25,28}

\definecolor{Spectral5C}{RGB}{215,25,28}

\definecolor{Spectral52}{RGB}{253,174,97}

\definecolor{Spectral5F}{RGB}{253,174,97}

\definecolor{Spectral53}{RGB}{255,255,191}

\definecolor{Spectral5H}{RGB}{255,255,191}

\definecolor{Spectral54}{RGB}{171,221,164}

\definecolor{Spectral5J}{RGB}{171,221,164}

\definecolor{Spectral55}{RGB}{43,131,186}

\definecolor{Spectral5M}{RGB}{43,131,186}

\definecolor{Spectral61}{RGB}{213,62,79}

\definecolor{Spectral6B}{RGB}{213,62,79}

\definecolor{Spectral62}{RGB}{252,141,89}

\definecolor{Spectral6E}{RGB}{252,141,89}

\definecolor{Spectral63}{RGB}{254,224,139}

\definecolor{Spectral6G}{RGB}{254,224,139}

\definecolor{Spectral64}{RGB}{230,245,152}

\definecolor{Spectral6I}{RGB}{230,245,152}

\definecolor{Spectral65}{RGB}{153,213,148}

\definecolor{Spectral6K}{RGB}{153,213,148}

\definecolor{Spectral66}{RGB}{50,136,189}

\definecolor{Spectral6N}{RGB}{50,136,189}

\definecolor{Spectral71}{RGB}{213,62,79}

\definecolor{Spectral7B}{RGB}{213,62,79}

\definecolor{Spectral72}{RGB}{252,141,89}

\definecolor{Spectral7E}{RGB}{252,141,89}

\definecolor{Spectral73}{RGB}{254,224,139}

\definecolor{Spectral7G}{RGB}{254,224,139}

\definecolor{Spectral74}{RGB}{255,255,191}

\definecolor{Spectral7H}{RGB}{255,255,191}

\definecolor{Spectral75}{RGB}{230,245,152}

\definecolor{Spectral7I}{RGB}{230,245,152}

\definecolor{Spectral76}{RGB}{153,213,148}

\definecolor{Spectral7K}{RGB}{153,213,148}

\definecolor{Spectral77}{RGB}{50,136,189}

\definecolor{Spectral7N}{RGB}{50,136,189}

\definecolor{Spectral81}{RGB}{213,62,79}

\definecolor{Spectral8B}{RGB}{213,62,79}

\definecolor{Spectral82}{RGB}{244,109,67}

\definecolor{Spectral8D}{RGB}{244,109,67}

\definecolor{Spectral83}{RGB}{253,174,97}

\definecolor{Spectral8F}{RGB}{253,174,97}

\definecolor{Spectral84}{RGB}{254,224,139}

\definecolor{Spectral8G}{RGB}{254,224,139}

\definecolor{Spectral85}{RGB}{230,245,152}

\definecolor{Spectral8I}{RGB}{230,245,152}

\definecolor{Spectral86}{RGB}{171,221,164}

\definecolor{Spectral8J}{RGB}{171,221,164}

\definecolor{Spectral87}{RGB}{102,194,165}

\definecolor{Spectral8L}{RGB}{102,194,165}

\definecolor{Spectral88}{RGB}{50,136,189}

\definecolor{Spectral8N}{RGB}{50,136,189}

\definecolor{Spectral91}{RGB}{213,62,79}

\definecolor{Spectral9B}{RGB}{213,62,79}

\definecolor{Spectral92}{RGB}{244,109,67}

\definecolor{Spectral9D}{RGB}{244,109,67}

\definecolor{Spectral93}{RGB}{253,174,97}

\definecolor{Spectral9F}{RGB}{253,174,97}

\definecolor{Spectral94}{RGB}{254,224,139}

\definecolor{Spectral9G}{RGB}{254,224,139}

\definecolor{Spectral95}{RGB}{255,255,191}

\definecolor{Spectral9H}{RGB}{255,255,191}

\definecolor{Spectral96}{RGB}{230,245,152}

\definecolor{Spectral9I}{RGB}{230,245,152}

\definecolor{Spectral97}{RGB}{171,221,164}

\definecolor{Spectral9J}{RGB}{171,221,164}

\definecolor{Spectral98}{RGB}{102,194,165}

\definecolor{Spectral9L}{RGB}{102,194,165}

\definecolor{Spectral99}{RGB}{50,136,189}

\definecolor{Spectral9N}{RGB}{50,136,189}

\definecolor{Spectral101}{RGB}{158,1,66}

\definecolor{Spectral10A}{RGB}{158,1,66}

\definecolor{Spectral102}{RGB}{213,62,79}

\definecolor{Spectral10B}{RGB}{213,62,79}

\definecolor{Spectral103}{RGB}{244,109,67}

\definecolor{Spectral10D}{RGB}{244,109,67}

\definecolor{Spectral104}{RGB}{253,174,97}

\definecolor{Spectral10F}{RGB}{253,174,97}

\definecolor{Spectral105}{RGB}{254,224,139}

\definecolor{Spectral10G}{RGB}{254,224,139}

\definecolor{Spectral106}{RGB}{230,245,152}

\definecolor{Spectral10I}{RGB}{230,245,152}

\definecolor{Spectral107}{RGB}{171,221,164}

\definecolor{Spectral10J}{RGB}{171,221,164}

\definecolor{Spectral108}{RGB}{102,194,165}

\definecolor{Spectral10L}{RGB}{102,194,165}

\definecolor{Spectral109}{RGB}{50,136,189}

\definecolor{Spectral10N}{RGB}{50,136,189}

\definecolor{Spectral1010}{RGB}{94,79,162}

\definecolor{Spectral10O}{RGB}{94,79,162}

\definecolor{Spectral111}{RGB}{158,1,66}

\definecolor{Spectral11A}{RGB}{158,1,66}

\definecolor{Spectral112}{RGB}{213,62,79}

\definecolor{Spectral11B}{RGB}{213,62,79}

\definecolor{Spectral113}{RGB}{244,109,67}

\definecolor{Spectral11D}{RGB}{244,109,67}

\definecolor{Spectral114}{RGB}{253,174,97}

\definecolor{Spectral11F}{RGB}{253,174,97}

\definecolor{Spectral115}{RGB}{254,224,139}

\definecolor{Spectral11G}{RGB}{254,224,139}

\definecolor{Spectral116}{RGB}{255,255,191}

\definecolor{Spectral11H}{RGB}{255,255,191}

\definecolor{Spectral117}{RGB}{230,245,152}

\definecolor{Spectral11I}{RGB}{230,245,152}

\definecolor{Spectral118}{RGB}{171,221,164}

\definecolor{Spectral11J}{RGB}{171,221,164}

\definecolor{Spectral119}{RGB}{102,194,165}

\definecolor{Spectral11L}{RGB}{102,194,165}

\definecolor{Spectral1110}{RGB}{50,136,189}

\definecolor{Spectral11N}{RGB}{50,136,189}

\definecolor{Spectral1111}{RGB}{94,79,162}

\definecolor{Spectral11O}{RGB}{94,79,162}

\definecolor{RdYlGn31}{RGB}{252,141,89}

\definecolor{RdYlGn3E}{RGB}{252,141,89}

\definecolor{RdYlGn32}{RGB}{255,255,191}

\definecolor{RdYlGn3H}{RGB}{255,255,191}

\definecolor{RdYlGn33}{RGB}{145,207,96}

\definecolor{RdYlGn3K}{RGB}{145,207,96}

\definecolor{RdYlGn41}{RGB}{215,25,28}

\definecolor{RdYlGn4C}{RGB}{215,25,28}

\definecolor{RdYlGn42}{RGB}{253,174,97}

\definecolor{RdYlGn4F}{RGB}{253,174,97}

\definecolor{RdYlGn43}{RGB}{166,217,106}

\definecolor{RdYlGn4J}{RGB}{166,217,106}

\definecolor{RdYlGn44}{RGB}{26,150,65}

\definecolor{RdYlGn4M}{RGB}{26,150,65}

\definecolor{RdYlGn51}{RGB}{215,25,28}

\definecolor{RdYlGn5C}{RGB}{215,25,28}

\definecolor{RdYlGn52}{RGB}{253,174,97}

\definecolor{RdYlGn5F}{RGB}{253,174,97}

\definecolor{RdYlGn53}{RGB}{255,255,191}

\definecolor{RdYlGn5H}{RGB}{255,255,191}

\definecolor{RdYlGn54}{RGB}{166,217,106}

\definecolor{RdYlGn5J}{RGB}{166,217,106}

\definecolor{RdYlGn55}{RGB}{26,150,65}

\definecolor{RdYlGn5M}{RGB}{26,150,65}

\definecolor{RdYlGn61}{RGB}{215,48,39}

\definecolor{RdYlGn6B}{RGB}{215,48,39}

\definecolor{RdYlGn62}{RGB}{252,141,89}

\definecolor{RdYlGn6E}{RGB}{252,141,89}

\definecolor{RdYlGn63}{RGB}{254,224,139}

\definecolor{RdYlGn6G}{RGB}{254,224,139}

\definecolor{RdYlGn64}{RGB}{217,239,139}

\definecolor{RdYlGn6I}{RGB}{217,239,139}

\definecolor{RdYlGn65}{RGB}{145,207,96}

\definecolor{RdYlGn6K}{RGB}{145,207,96}

\definecolor{RdYlGn66}{RGB}{26,152,80}

\definecolor{RdYlGn6N}{RGB}{26,152,80}

\definecolor{RdYlGn71}{RGB}{215,48,39}

\definecolor{RdYlGn7B}{RGB}{215,48,39}

\definecolor{RdYlGn72}{RGB}{252,141,89}

\definecolor{RdYlGn7E}{RGB}{252,141,89}

\definecolor{RdYlGn73}{RGB}{254,224,139}

\definecolor{RdYlGn7G}{RGB}{254,224,139}

\definecolor{RdYlGn74}{RGB}{255,255,191}

\definecolor{RdYlGn7H}{RGB}{255,255,191}

\definecolor{RdYlGn75}{RGB}{217,239,139}

\definecolor{RdYlGn7I}{RGB}{217,239,139}

\definecolor{RdYlGn76}{RGB}{145,207,96}

\definecolor{RdYlGn7K}{RGB}{145,207,96}

\definecolor{RdYlGn77}{RGB}{26,152,80}

\definecolor{RdYlGn7N}{RGB}{26,152,80}

\definecolor{RdYlGn81}{RGB}{215,48,39}

\definecolor{RdYlGn8B}{RGB}{215,48,39}

\definecolor{RdYlGn82}{RGB}{244,109,67}

\definecolor{RdYlGn8D}{RGB}{244,109,67}

\definecolor{RdYlGn83}{RGB}{253,174,97}

\definecolor{RdYlGn8F}{RGB}{253,174,97}

\definecolor{RdYlGn84}{RGB}{254,224,139}

\definecolor{RdYlGn8G}{RGB}{254,224,139}

\definecolor{RdYlGn85}{RGB}{217,239,139}

\definecolor{RdYlGn8I}{RGB}{217,239,139}

\definecolor{RdYlGn86}{RGB}{166,217,106}

\definecolor{RdYlGn8J}{RGB}{166,217,106}

\definecolor{RdYlGn87}{RGB}{102,189,99}

\definecolor{RdYlGn8L}{RGB}{102,189,99}

\definecolor{RdYlGn88}{RGB}{26,152,80}

\definecolor{RdYlGn8N}{RGB}{26,152,80}

\definecolor{RdYlGn91}{RGB}{215,48,39}

\definecolor{RdYlGn9B}{RGB}{215,48,39}

\definecolor{RdYlGn92}{RGB}{244,109,67}

\definecolor{RdYlGn9D}{RGB}{244,109,67}

\definecolor{RdYlGn93}{RGB}{253,174,97}

\definecolor{RdYlGn9F}{RGB}{253,174,97}

\definecolor{RdYlGn94}{RGB}{254,224,139}

\definecolor{RdYlGn9G}{RGB}{254,224,139}

\definecolor{RdYlGn95}{RGB}{255,255,191}

\definecolor{RdYlGn9H}{RGB}{255,255,191}

\definecolor{RdYlGn96}{RGB}{217,239,139}

\definecolor{RdYlGn9I}{RGB}{217,239,139}

\definecolor{RdYlGn97}{RGB}{166,217,106}

\definecolor{RdYlGn9J}{RGB}{166,217,106}

\definecolor{RdYlGn98}{RGB}{102,189,99}

\definecolor{RdYlGn9L}{RGB}{102,189,99}

\definecolor{RdYlGn99}{RGB}{26,152,80}

\definecolor{RdYlGn9N}{RGB}{26,152,80}

\definecolor{RdYlGn101}{RGB}{165,0,38}

\definecolor{RdYlGn10A}{RGB}{165,0,38}

\definecolor{RdYlGn102}{RGB}{215,48,39}

\definecolor{RdYlGn10B}{RGB}{215,48,39}

\definecolor{RdYlGn103}{RGB}{244,109,67}

\definecolor{RdYlGn10D}{RGB}{244,109,67}

\definecolor{RdYlGn104}{RGB}{253,174,97}

\definecolor{RdYlGn10F}{RGB}{253,174,97}

\definecolor{RdYlGn105}{RGB}{254,224,139}

\definecolor{RdYlGn10G}{RGB}{254,224,139}

\definecolor{RdYlGn106}{RGB}{217,239,139}

\definecolor{RdYlGn10I}{RGB}{217,239,139}

\definecolor{RdYlGn107}{RGB}{166,217,106}

\definecolor{RdYlGn10J}{RGB}{166,217,106}

\definecolor{RdYlGn108}{RGB}{102,189,99}

\definecolor{RdYlGn10L}{RGB}{102,189,99}

\definecolor{RdYlGn109}{RGB}{26,152,80}

\definecolor{RdYlGn10N}{RGB}{26,152,80}

\definecolor{RdYlGn1010}{RGB}{0,104,55}

\definecolor{RdYlGn10O}{RGB}{0,104,55}

\definecolor{RdYlGn111}{RGB}{165,0,38}

\definecolor{RdYlGn11A}{RGB}{165,0,38}

\definecolor{RdYlGn112}{RGB}{215,48,39}

\definecolor{RdYlGn11B}{RGB}{215,48,39}

\definecolor{RdYlGn113}{RGB}{244,109,67}

\definecolor{RdYlGn11D}{RGB}{244,109,67}

\definecolor{RdYlGn114}{RGB}{253,174,97}

\definecolor{RdYlGn11F}{RGB}{253,174,97}

\definecolor{RdYlGn115}{RGB}{254,224,139}

\definecolor{RdYlGn11G}{RGB}{254,224,139}

\definecolor{RdYlGn116}{RGB}{255,255,191}

\definecolor{RdYlGn11H}{RGB}{255,255,191}

\definecolor{RdYlGn117}{RGB}{217,239,139}

\definecolor{RdYlGn11I}{RGB}{217,239,139}

\definecolor{RdYlGn118}{RGB}{166,217,106}

\definecolor{RdYlGn11J}{RGB}{166,217,106}

\definecolor{RdYlGn119}{RGB}{102,189,99}

\definecolor{RdYlGn11L}{RGB}{102,189,99}

\definecolor{RdYlGn1110}{RGB}{26,152,80}

\definecolor{RdYlGn11N}{RGB}{26,152,80}

\definecolor{RdYlGn1111}{RGB}{0,104,55}

\definecolor{RdYlGn11O}{RGB}{0,104,55}

\definecolor{Set331}{RGB}{141,211,199}

\definecolor{Set33A}{RGB}{141,211,199}

\definecolor{Set332}{RGB}{255,255,179}

\definecolor{Set33B}{RGB}{255,255,179}

\definecolor{Set333}{RGB}{190,186,218}

\definecolor{Set33C}{RGB}{190,186,218}

\definecolor{Set341}{RGB}{141,211,199}

\definecolor{Set34A}{RGB}{141,211,199}

\definecolor{Set342}{RGB}{255,255,179}

\definecolor{Set34B}{RGB}{255,255,179}

\definecolor{Set343}{RGB}{190,186,218}

\definecolor{Set34C}{RGB}{190,186,218}

\definecolor{Set344}{RGB}{251,128,114}

\definecolor{Set34D}{RGB}{251,128,114}

\definecolor{Set351}{RGB}{141,211,199}

\definecolor{Set35A}{RGB}{141,211,199}

\definecolor{Set352}{RGB}{255,255,179}

\definecolor{Set35B}{RGB}{255,255,179}

\definecolor{Set353}{RGB}{190,186,218}

\definecolor{Set35C}{RGB}{190,186,218}

\definecolor{Set354}{RGB}{251,128,114}

\definecolor{Set35D}{RGB}{251,128,114}

\definecolor{Set355}{RGB}{128,177,211}

\definecolor{Set35E}{RGB}{128,177,211}

\definecolor{Set361}{RGB}{141,211,199}

\definecolor{Set36A}{RGB}{141,211,199}

\definecolor{Set362}{RGB}{255,255,179}

\definecolor{Set36B}{RGB}{255,255,179}

\definecolor{Set363}{RGB}{190,186,218}

\definecolor{Set36C}{RGB}{190,186,218}

\definecolor{Set364}{RGB}{251,128,114}

\definecolor{Set36D}{RGB}{251,128,114}

\definecolor{Set365}{RGB}{128,177,211}

\definecolor{Set36E}{RGB}{128,177,211}

\definecolor{Set366}{RGB}{253,180,98}

\definecolor{Set36F}{RGB}{253,180,98}

\definecolor{Set371}{RGB}{141,211,199}

\definecolor{Set37A}{RGB}{141,211,199}

\definecolor{Set372}{RGB}{255,255,179}

\definecolor{Set37B}{RGB}{255,255,179}

\definecolor{Set373}{RGB}{190,186,218}

\definecolor{Set37C}{RGB}{190,186,218}

\definecolor{Set374}{RGB}{251,128,114}

\definecolor{Set37D}{RGB}{251,128,114}

\definecolor{Set375}{RGB}{128,177,211}

\definecolor{Set37E}{RGB}{128,177,211}

\definecolor{Set376}{RGB}{253,180,98}

\definecolor{Set37F}{RGB}{253,180,98}

\definecolor{Set377}{RGB}{179,222,105}

\definecolor{Set37G}{RGB}{179,222,105}

\definecolor{Set381}{RGB}{141,211,199}

\definecolor{Set38A}{RGB}{141,211,199}

\definecolor{Set382}{RGB}{255,255,179}

\definecolor{Set38B}{RGB}{255,255,179}

\definecolor{Set383}{RGB}{190,186,218}

\definecolor{Set38C}{RGB}{190,186,218}

\definecolor{Set384}{RGB}{251,128,114}

\definecolor{Set38D}{RGB}{251,128,114}

\definecolor{Set385}{RGB}{128,177,211}

\definecolor{Set38E}{RGB}{128,177,211}

\definecolor{Set386}{RGB}{253,180,98}

\definecolor{Set38F}{RGB}{253,180,98}

\definecolor{Set387}{RGB}{179,222,105}

\definecolor{Set38G}{RGB}{179,222,105}

\definecolor{Set388}{RGB}{252,205,229}

\definecolor{Set38H}{RGB}{252,205,229}

\definecolor{Set391}{RGB}{141,211,199}

\definecolor{Set39A}{RGB}{141,211,199}

\definecolor{Set392}{RGB}{255,255,179}

\definecolor{Set39B}{RGB}{255,255,179}

\definecolor{Set393}{RGB}{190,186,218}

\definecolor{Set39C}{RGB}{190,186,218}

\definecolor{Set394}{RGB}{251,128,114}

\definecolor{Set39D}{RGB}{251,128,114}

\definecolor{Set395}{RGB}{128,177,211}

\definecolor{Set39E}{RGB}{128,177,211}

\definecolor{Set396}{RGB}{253,180,98}

\definecolor{Set39F}{RGB}{253,180,98}

\definecolor{Set397}{RGB}{179,222,105}

\definecolor{Set39G}{RGB}{179,222,105}

\definecolor{Set398}{RGB}{252,205,229}

\definecolor{Set39H}{RGB}{252,205,229}

\definecolor{Set399}{RGB}{217,217,217}

\definecolor{Set39I}{RGB}{217,217,217}

\definecolor{Set3101}{RGB}{141,211,199}

\definecolor{Set310A}{RGB}{141,211,199}

\definecolor{Set3102}{RGB}{255,255,179}

\definecolor{Set310B}{RGB}{255,255,179}

\definecolor{Set3103}{RGB}{190,186,218}

\definecolor{Set310C}{RGB}{190,186,218}

\definecolor{Set3104}{RGB}{251,128,114}

\definecolor{Set310D}{RGB}{251,128,114}

\definecolor{Set3105}{RGB}{128,177,211}

\definecolor{Set310E}{RGB}{128,177,211}

\definecolor{Set3106}{RGB}{253,180,98}

\definecolor{Set310F}{RGB}{253,180,98}

\definecolor{Set3107}{RGB}{179,222,105}

\definecolor{Set310G}{RGB}{179,222,105}

\definecolor{Set3108}{RGB}{252,205,229}

\definecolor{Set310H}{RGB}{252,205,229}

\definecolor{Set3109}{RGB}{217,217,217}

\definecolor{Set310I}{RGB}{217,217,217}

\definecolor{Set31010}{RGB}{188,128,189}

\definecolor{Set310J}{RGB}{188,128,189}

\definecolor{Set3111}{RGB}{141,211,199}

\definecolor{Set311A}{RGB}{141,211,199}

\definecolor{Set3112}{RGB}{255,255,179}

\definecolor{Set311B}{RGB}{255,255,179}

\definecolor{Set3113}{RGB}{190,186,218}

\definecolor{Set311C}{RGB}{190,186,218}

\definecolor{Set3114}{RGB}{251,128,114}

\definecolor{Set311D}{RGB}{251,128,114}

\definecolor{Set3115}{RGB}{128,177,211}

\definecolor{Set311E}{RGB}{128,177,211}

\definecolor{Set3116}{RGB}{253,180,98}

\definecolor{Set311F}{RGB}{253,180,98}

\definecolor{Set3117}{RGB}{179,222,105}

\definecolor{Set311G}{RGB}{179,222,105}

\definecolor{Set3118}{RGB}{252,205,229}

\definecolor{Set311H}{RGB}{252,205,229}

\definecolor{Set3119}{RGB}{217,217,217}

\definecolor{Set311I}{RGB}{217,217,217}

\definecolor{Set31110}{RGB}{188,128,189}

\definecolor{Set311J}{RGB}{188,128,189}

\definecolor{Set31111}{RGB}{204,235,197}

\definecolor{Set311K}{RGB}{204,235,197}

\definecolor{Set3121}{RGB}{141,211,199}

\definecolor{Set312A}{RGB}{141,211,199}

\definecolor{Set3122}{RGB}{255,255,179}

\definecolor{Set312B}{RGB}{255,255,179}

\definecolor{Set3123}{RGB}{190,186,218}

\definecolor{Set312C}{RGB}{190,186,218}

\definecolor{Set3124}{RGB}{251,128,114}

\definecolor{Set312D}{RGB}{251,128,114}

\definecolor{Set3125}{RGB}{128,177,211}

\definecolor{Set312E}{RGB}{128,177,211}

\definecolor{Set3126}{RGB}{253,180,98}

\definecolor{Set312F}{RGB}{253,180,98}

\definecolor{Set3127}{RGB}{179,222,105}

\definecolor{Set312G}{RGB}{179,222,105}

\definecolor{Set3128}{RGB}{252,205,229}

\definecolor{Set312H}{RGB}{252,205,229}

\definecolor{Set3129}{RGB}{217,217,217}

\definecolor{Set312I}{RGB}{217,217,217}

\definecolor{Set31210}{RGB}{188,128,189}

\definecolor{Set312J}{RGB}{188,128,189}

\definecolor{Set31211}{RGB}{204,235,197}

\definecolor{Set312K}{RGB}{204,235,197}

\definecolor{Set31212}{RGB}{255,237,111}

\definecolor{Set312L}{RGB}{255,237,111}

\definecolor{Pastel131}{RGB}{251,180,174}

\definecolor{Pastel13A}{RGB}{251,180,174}

\definecolor{Pastel132}{RGB}{179,205,227}

\definecolor{Pastel13B}{RGB}{179,205,227}

\definecolor{Pastel133}{RGB}{204,235,197}

\definecolor{Pastel13C}{RGB}{204,235,197}

\definecolor{Pastel141}{RGB}{251,180,174}

\definecolor{Pastel14A}{RGB}{251,180,174}

\definecolor{Pastel142}{RGB}{179,205,227}

\definecolor{Pastel14B}{RGB}{179,205,227}

\definecolor{Pastel143}{RGB}{204,235,197}

\definecolor{Pastel14C}{RGB}{204,235,197}

\definecolor{Pastel144}{RGB}{222,203,228}

\definecolor{Pastel14D}{RGB}{222,203,228}

\definecolor{Pastel151}{RGB}{251,180,174}

\definecolor{Pastel15A}{RGB}{251,180,174}

\definecolor{Pastel152}{RGB}{179,205,227}

\definecolor{Pastel15B}{RGB}{179,205,227}

\definecolor{Pastel153}{RGB}{204,235,197}

\definecolor{Pastel15C}{RGB}{204,235,197}

\definecolor{Pastel154}{RGB}{222,203,228}

\definecolor{Pastel15D}{RGB}{222,203,228}

\definecolor{Pastel155}{RGB}{254,217,166}

\definecolor{Pastel15E}{RGB}{254,217,166}

\definecolor{Pastel161}{RGB}{251,180,174}

\definecolor{Pastel16A}{RGB}{251,180,174}

\definecolor{Pastel162}{RGB}{179,205,227}

\definecolor{Pastel16B}{RGB}{179,205,227}

\definecolor{Pastel163}{RGB}{204,235,197}

\definecolor{Pastel16C}{RGB}{204,235,197}

\definecolor{Pastel164}{RGB}{222,203,228}

\definecolor{Pastel16D}{RGB}{222,203,228}

\definecolor{Pastel165}{RGB}{254,217,166}

\definecolor{Pastel16E}{RGB}{254,217,166}

\definecolor{Pastel166}{RGB}{255,255,204}

\definecolor{Pastel16F}{RGB}{255,255,204}

\definecolor{Pastel171}{RGB}{251,180,174}

\definecolor{Pastel17A}{RGB}{251,180,174}

\definecolor{Pastel172}{RGB}{179,205,227}

\definecolor{Pastel17B}{RGB}{179,205,227}

\definecolor{Pastel173}{RGB}{204,235,197}

\definecolor{Pastel17C}{RGB}{204,235,197}

\definecolor{Pastel174}{RGB}{222,203,228}

\definecolor{Pastel17D}{RGB}{222,203,228}

\definecolor{Pastel175}{RGB}{254,217,166}

\definecolor{Pastel17E}{RGB}{254,217,166}

\definecolor{Pastel176}{RGB}{255,255,204}

\definecolor{Pastel17F}{RGB}{255,255,204}

\definecolor{Pastel177}{RGB}{229,216,189}

\definecolor{Pastel17G}{RGB}{229,216,189}

\definecolor{Pastel181}{RGB}{251,180,174}

\definecolor{Pastel18A}{RGB}{251,180,174}

\definecolor{Pastel182}{RGB}{179,205,227}

\definecolor{Pastel18B}{RGB}{179,205,227}

\definecolor{Pastel183}{RGB}{204,235,197}

\definecolor{Pastel18C}{RGB}{204,235,197}

\definecolor{Pastel184}{RGB}{222,203,228}

\definecolor{Pastel18D}{RGB}{222,203,228}

\definecolor{Pastel185}{RGB}{254,217,166}

\definecolor{Pastel18E}{RGB}{254,217,166}

\definecolor{Pastel186}{RGB}{255,255,204}

\definecolor{Pastel18F}{RGB}{255,255,204}

\definecolor{Pastel187}{RGB}{229,216,189}

\definecolor{Pastel18G}{RGB}{229,216,189}

\definecolor{Pastel188}{RGB}{253,218,236}

\definecolor{Pastel18H}{RGB}{253,218,236}

\definecolor{Pastel191}{RGB}{251,180,174}

\definecolor{Pastel19A}{RGB}{251,180,174}

\definecolor{Pastel192}{RGB}{179,205,227}

\definecolor{Pastel19B}{RGB}{179,205,227}

\definecolor{Pastel193}{RGB}{204,235,197}

\definecolor{Pastel19C}{RGB}{204,235,197}

\definecolor{Pastel194}{RGB}{222,203,228}

\definecolor{Pastel19D}{RGB}{222,203,228}

\definecolor{Pastel195}{RGB}{254,217,166}

\definecolor{Pastel19E}{RGB}{254,217,166}

\definecolor{Pastel196}{RGB}{255,255,204}

\definecolor{Pastel19F}{RGB}{255,255,204}

\definecolor{Pastel197}{RGB}{229,216,189}

\definecolor{Pastel19G}{RGB}{229,216,189}

\definecolor{Pastel198}{RGB}{253,218,236}

\definecolor{Pastel19H}{RGB}{253,218,236}

\definecolor{Pastel199}{RGB}{242,242,242}

\definecolor{Pastel19I}{RGB}{242,242,242}

\definecolor{Set131}{RGB}{228,26,28}

\definecolor{Set13A}{RGB}{228,26,28}

\definecolor{Set132}{RGB}{55,126,184}

\definecolor{Set13B}{RGB}{55,126,184}

\definecolor{Set133}{RGB}{77,175,74}

\definecolor{Set13C}{RGB}{77,175,74}

\definecolor{Set141}{RGB}{228,26,28}

\definecolor{Set14A}{RGB}{228,26,28}

\definecolor{Set142}{RGB}{55,126,184}

\definecolor{Set14B}{RGB}{55,126,184}

\definecolor{Set143}{RGB}{77,175,74}

\definecolor{Set14C}{RGB}{77,175,74}

\definecolor{Set144}{RGB}{152,78,163}

\definecolor{Set14D}{RGB}{152,78,163}

\definecolor{Set151}{RGB}{228,26,28}

\definecolor{Set15A}{RGB}{228,26,28}

\definecolor{Set152}{RGB}{55,126,184}

\definecolor{Set15B}{RGB}{55,126,184}

\definecolor{Set153}{RGB}{77,175,74}

\definecolor{Set15C}{RGB}{77,175,74}

\definecolor{Set154}{RGB}{152,78,163}

\definecolor{Set15D}{RGB}{152,78,163}

\definecolor{Set155}{RGB}{255,127,0}

\definecolor{Set15E}{RGB}{255,127,0}

\definecolor{Set161}{RGB}{228,26,28}

\definecolor{Set16A}{RGB}{228,26,28}

\definecolor{Set162}{RGB}{55,126,184}

\definecolor{Set16B}{RGB}{55,126,184}

\definecolor{Set163}{RGB}{77,175,74}

\definecolor{Set16C}{RGB}{77,175,74}

\definecolor{Set164}{RGB}{152,78,163}

\definecolor{Set16D}{RGB}{152,78,163}

\definecolor{Set165}{RGB}{255,127,0}

\definecolor{Set16E}{RGB}{255,127,0}

\definecolor{Set166}{RGB}{255,255,51}

\definecolor{Set16F}{RGB}{255,255,51}

\definecolor{Set171}{RGB}{228,26,28}

\definecolor{Set17A}{RGB}{228,26,28}

\definecolor{Set172}{RGB}{55,126,184}

\definecolor{Set17B}{RGB}{55,126,184}

\definecolor{Set173}{RGB}{77,175,74}

\definecolor{Set17C}{RGB}{77,175,74}

\definecolor{Set174}{RGB}{152,78,163}

\definecolor{Set17D}{RGB}{152,78,163}

\definecolor{Set175}{RGB}{255,127,0}

\definecolor{Set17E}{RGB}{255,127,0}

\definecolor{Set176}{RGB}{255,255,51}

\definecolor{Set17F}{RGB}{255,255,51}

\definecolor{Set177}{RGB}{166,86,40}

\definecolor{Set17G}{RGB}{166,86,40}

\definecolor{Set181}{RGB}{228,26,28}

\definecolor{Set18A}{RGB}{228,26,28}

\definecolor{Set182}{RGB}{55,126,184}

\definecolor{Set18B}{RGB}{55,126,184}

\definecolor{Set183}{RGB}{77,175,74}

\definecolor{Set18C}{RGB}{77,175,74}

\definecolor{Set184}{RGB}{152,78,163}

\definecolor{Set18D}{RGB}{152,78,163}

\definecolor{Set185}{RGB}{255,127,0}

\definecolor{Set18E}{RGB}{255,127,0}

\definecolor{Set186}{RGB}{255,255,51}

\definecolor{Set18F}{RGB}{255,255,51}

\definecolor{Set187}{RGB}{166,86,40}

\definecolor{Set18G}{RGB}{166,86,40}

\definecolor{Set188}{RGB}{247,129,191}

\definecolor{Set18H}{RGB}{247,129,191}

\definecolor{Set191}{RGB}{228,26,28}

\definecolor{Set19A}{RGB}{228,26,28}

\definecolor{Set192}{RGB}{55,126,184}

\definecolor{Set19B}{RGB}{55,126,184}

\definecolor{Set193}{RGB}{77,175,74}

\definecolor{Set19C}{RGB}{77,175,74}

\definecolor{Set194}{RGB}{152,78,163}

\definecolor{Set19D}{RGB}{152,78,163}

\definecolor{Set195}{RGB}{255,127,0}

\definecolor{Set19E}{RGB}{255,127,0}

\definecolor{Set196}{RGB}{255,255,51}

\definecolor{Set19F}{RGB}{255,255,51}

\definecolor{Set197}{RGB}{166,86,40}

\definecolor{Set19G}{RGB}{166,86,40}

\definecolor{Set198}{RGB}{247,129,191}

\definecolor{Set19H}{RGB}{247,129,191}

\definecolor{Set199}{RGB}{153,153,153}

\definecolor{Set19I}{RGB}{153,153,153}

\definecolor{Pastel231}{RGB}{179,226,205}

\definecolor{Pastel23A}{RGB}{179,226,205}

\definecolor{Pastel232}{RGB}{253,205,172}

\definecolor{Pastel23B}{RGB}{253,205,172}

\definecolor{Pastel233}{RGB}{203,213,232}

\definecolor{Pastel23C}{RGB}{203,213,232}

\definecolor{Pastel241}{RGB}{179,226,205}

\definecolor{Pastel24A}{RGB}{179,226,205}

\definecolor{Pastel242}{RGB}{253,205,172}

\definecolor{Pastel24B}{RGB}{253,205,172}

\definecolor{Pastel243}{RGB}{203,213,232}

\definecolor{Pastel24C}{RGB}{203,213,232}

\definecolor{Pastel244}{RGB}{244,202,228}

\definecolor{Pastel24D}{RGB}{244,202,228}

\definecolor{Pastel251}{RGB}{179,226,205}

\definecolor{Pastel25A}{RGB}{179,226,205}

\definecolor{Pastel252}{RGB}{253,205,172}

\definecolor{Pastel25B}{RGB}{253,205,172}

\definecolor{Pastel253}{RGB}{203,213,232}

\definecolor{Pastel25C}{RGB}{203,213,232}

\definecolor{Pastel254}{RGB}{244,202,228}

\definecolor{Pastel25D}{RGB}{244,202,228}

\definecolor{Pastel255}{RGB}{230,245,201}

\definecolor{Pastel25E}{RGB}{230,245,201}

\definecolor{Pastel261}{RGB}{179,226,205}

\definecolor{Pastel26A}{RGB}{179,226,205}

\definecolor{Pastel262}{RGB}{253,205,172}

\definecolor{Pastel26B}{RGB}{253,205,172}

\definecolor{Pastel263}{RGB}{203,213,232}

\definecolor{Pastel26C}{RGB}{203,213,232}

\definecolor{Pastel264}{RGB}{244,202,228}

\definecolor{Pastel26D}{RGB}{244,202,228}

\definecolor{Pastel265}{RGB}{230,245,201}

\definecolor{Pastel26E}{RGB}{230,245,201}

\definecolor{Pastel266}{RGB}{255,242,174}

\definecolor{Pastel26F}{RGB}{255,242,174}

\definecolor{Pastel271}{RGB}{179,226,205}

\definecolor{Pastel27A}{RGB}{179,226,205}

\definecolor{Pastel272}{RGB}{253,205,172}

\definecolor{Pastel27B}{RGB}{253,205,172}

\definecolor{Pastel273}{RGB}{203,213,232}

\definecolor{Pastel27C}{RGB}{203,213,232}

\definecolor{Pastel274}{RGB}{244,202,228}

\definecolor{Pastel27D}{RGB}{244,202,228}

\definecolor{Pastel275}{RGB}{230,245,201}

\definecolor{Pastel27E}{RGB}{230,245,201}

\definecolor{Pastel276}{RGB}{255,242,174}

\definecolor{Pastel27F}{RGB}{255,242,174}

\definecolor{Pastel277}{RGB}{241,226,204}

\definecolor{Pastel27G}{RGB}{241,226,204}

\definecolor{Pastel281}{RGB}{179,226,205}

\definecolor{Pastel28A}{RGB}{179,226,205}

\definecolor{Pastel282}{RGB}{253,205,172}

\definecolor{Pastel28B}{RGB}{253,205,172}

\definecolor{Pastel283}{RGB}{203,213,232}

\definecolor{Pastel28C}{RGB}{203,213,232}

\definecolor{Pastel284}{RGB}{244,202,228}

\definecolor{Pastel28D}{RGB}{244,202,228}

\definecolor{Pastel285}{RGB}{230,245,201}

\definecolor{Pastel28E}{RGB}{230,245,201}

\definecolor{Pastel286}{RGB}{255,242,174}

\definecolor{Pastel28F}{RGB}{255,242,174}

\definecolor{Pastel287}{RGB}{241,226,204}

\definecolor{Pastel28G}{RGB}{241,226,204}

\definecolor{Pastel288}{RGB}{204,204,204}

\definecolor{Pastel28H}{RGB}{204,204,204}

\definecolor{Set231}{RGB}{102,194,165}

\definecolor{Set23A}{RGB}{102,194,165}

\definecolor{Set232}{RGB}{252,141,98}

\definecolor{Set23B}{RGB}{252,141,98}

\definecolor{Set233}{RGB}{141,160,203}

\definecolor{Set23C}{RGB}{141,160,203}

\definecolor{Set241}{RGB}{102,194,165}

\definecolor{Set24A}{RGB}{102,194,165}

\definecolor{Set242}{RGB}{252,141,98}

\definecolor{Set24B}{RGB}{252,141,98}

\definecolor{Set243}{RGB}{141,160,203}

\definecolor{Set24C}{RGB}{141,160,203}

\definecolor{Set244}{RGB}{231,138,195}

\definecolor{Set24D}{RGB}{231,138,195}

\definecolor{Set251}{RGB}{102,194,165}

\definecolor{Set25A}{RGB}{102,194,165}

\definecolor{Set252}{RGB}{252,141,98}

\definecolor{Set25B}{RGB}{252,141,98}

\definecolor{Set253}{RGB}{141,160,203}

\definecolor{Set25C}{RGB}{141,160,203}

\definecolor{Set254}{RGB}{231,138,195}

\definecolor{Set25D}{RGB}{231,138,195}

\definecolor{Set255}{RGB}{166,216,84}

\definecolor{Set25E}{RGB}{166,216,84}

\definecolor{Set261}{RGB}{102,194,165}

\definecolor{Set26A}{RGB}{102,194,165}

\definecolor{Set262}{RGB}{252,141,98}

\definecolor{Set26B}{RGB}{252,141,98}

\definecolor{Set263}{RGB}{141,160,203}

\definecolor{Set26C}{RGB}{141,160,203}

\definecolor{Set264}{RGB}{231,138,195}

\definecolor{Set26D}{RGB}{231,138,195}

\definecolor{Set265}{RGB}{166,216,84}

\definecolor{Set26E}{RGB}{166,216,84}

\definecolor{Set266}{RGB}{255,217,47}

\definecolor{Set26F}{RGB}{255,217,47}

\definecolor{Set271}{RGB}{102,194,165}

\definecolor{Set27A}{RGB}{102,194,165}

\definecolor{Set272}{RGB}{252,141,98}

\definecolor{Set27B}{RGB}{252,141,98}

\definecolor{Set273}{RGB}{141,160,203}

\definecolor{Set27C}{RGB}{141,160,203}

\definecolor{Set274}{RGB}{231,138,195}

\definecolor{Set27D}{RGB}{231,138,195}

\definecolor{Set275}{RGB}{166,216,84}

\definecolor{Set27E}{RGB}{166,216,84}

\definecolor{Set276}{RGB}{255,217,47}

\definecolor{Set27F}{RGB}{255,217,47}

\definecolor{Set277}{RGB}{229,196,148}

\definecolor{Set27G}{RGB}{229,196,148}

\definecolor{Set281}{RGB}{102,194,165}

\definecolor{Set28A}{RGB}{102,194,165}

\definecolor{Set282}{RGB}{252,141,98}

\definecolor{Set28B}{RGB}{252,141,98}

\definecolor{Set283}{RGB}{141,160,203}

\definecolor{Set28C}{RGB}{141,160,203}

\definecolor{Set284}{RGB}{231,138,195}

\definecolor{Set28D}{RGB}{231,138,195}

\definecolor{Set285}{RGB}{166,216,84}

\definecolor{Set28E}{RGB}{166,216,84}

\definecolor{Set286}{RGB}{255,217,47}

\definecolor{Set28F}{RGB}{255,217,47}

\definecolor{Set287}{RGB}{229,196,148}

\definecolor{Set28G}{RGB}{229,196,148}

\definecolor{Set288}{RGB}{179,179,179}

\definecolor{Set28H}{RGB}{179,179,179}

\definecolor{Dark231}{RGB}{27,158,119}

\definecolor{Dark23A}{RGB}{27,158,119}

\definecolor{Dark232}{RGB}{217,95,2}

\definecolor{Dark23B}{RGB}{217,95,2}

\definecolor{Dark233}{RGB}{117,112,179}

\definecolor{Dark23C}{RGB}{117,112,179}

\definecolor{Dark241}{RGB}{27,158,119}

\definecolor{Dark24A}{RGB}{27,158,119}

\definecolor{Dark242}{RGB}{217,95,2}

\definecolor{Dark24B}{RGB}{217,95,2}

\definecolor{Dark243}{RGB}{117,112,179}

\definecolor{Dark24C}{RGB}{117,112,179}

\definecolor{Dark244}{RGB}{231,41,138}

\definecolor{Dark24D}{RGB}{231,41,138}

\definecolor{Dark251}{RGB}{27,158,119}

\definecolor{Dark25A}{RGB}{27,158,119}

\definecolor{Dark252}{RGB}{217,95,2}

\definecolor{Dark25B}{RGB}{217,95,2}

\definecolor{Dark253}{RGB}{117,112,179}

\definecolor{Dark25C}{RGB}{117,112,179}

\definecolor{Dark254}{RGB}{231,41,138}

\definecolor{Dark25D}{RGB}{231,41,138}

\definecolor{Dark255}{RGB}{102,166,30}

\definecolor{Dark25E}{RGB}{102,166,30}

\definecolor{Dark261}{RGB}{27,158,119}

\definecolor{Dark26A}{RGB}{27,158,119}

\definecolor{Dark262}{RGB}{217,95,2}

\definecolor{Dark26B}{RGB}{217,95,2}

\definecolor{Dark263}{RGB}{117,112,179}

\definecolor{Dark26C}{RGB}{117,112,179}

\definecolor{Dark264}{RGB}{231,41,138}

\definecolor{Dark26D}{RGB}{231,41,138}

\definecolor{Dark265}{RGB}{102,166,30}

\definecolor{Dark26E}{RGB}{102,166,30}

\definecolor{Dark266}{RGB}{230,171,2}

\definecolor{Dark26F}{RGB}{230,171,2}

\definecolor{Dark271}{RGB}{27,158,119}

\definecolor{Dark27A}{RGB}{27,158,119}

\definecolor{Dark272}{RGB}{217,95,2}

\definecolor{Dark27B}{RGB}{217,95,2}

\definecolor{Dark273}{RGB}{117,112,179}

\definecolor{Dark27C}{RGB}{117,112,179}

\definecolor{Dark274}{RGB}{231,41,138}

\definecolor{Dark27D}{RGB}{231,41,138}

\definecolor{Dark275}{RGB}{102,166,30}

\definecolor{Dark27E}{RGB}{102,166,30}

\definecolor{Dark276}{RGB}{230,171,2}

\definecolor{Dark27F}{RGB}{230,171,2}

\definecolor{Dark277}{RGB}{166,118,29}

\definecolor{Dark27G}{RGB}{166,118,29}

\definecolor{Dark281}{RGB}{27,158,119}

\definecolor{Dark28A}{RGB}{27,158,119}

\definecolor{Dark282}{RGB}{217,95,2}

\definecolor{Dark28B}{RGB}{217,95,2}

\definecolor{Dark283}{RGB}{117,112,179}

\definecolor{Dark28C}{RGB}{117,112,179}

\definecolor{Dark284}{RGB}{231,41,138}

\definecolor{Dark28D}{RGB}{231,41,138}

\definecolor{Dark285}{RGB}{102,166,30}

\definecolor{Dark28E}{RGB}{102,166,30}

\definecolor{Dark286}{RGB}{230,171,2}

\definecolor{Dark28F}{RGB}{230,171,2}

\definecolor{Dark287}{RGB}{166,118,29}

\definecolor{Dark28G}{RGB}{166,118,29}

\definecolor{Dark288}{RGB}{102,102,102}

\definecolor{Dark28H}{RGB}{102,102,102}

\definecolor{Paired31}{RGB}{166,206,227}

\definecolor{Paired3A}{RGB}{166,206,227}

\definecolor{Paired32}{RGB}{31,120,180}

\definecolor{Paired3B}{RGB}{31,120,180}

\definecolor{Paired33}{RGB}{178,223,138}

\definecolor{Paired3C}{RGB}{178,223,138}

\definecolor{Paired41}{RGB}{166,206,227}

\definecolor{Paired4A}{RGB}{166,206,227}

\definecolor{Paired42}{RGB}{31,120,180}

\definecolor{Paired4B}{RGB}{31,120,180}

\definecolor{Paired43}{RGB}{178,223,138}

\definecolor{Paired4C}{RGB}{178,223,138}

\definecolor{Paired44}{RGB}{51,160,44}

\definecolor{Paired4D}{RGB}{51,160,44}

\definecolor{Paired51}{RGB}{166,206,227}

\definecolor{Paired5A}{RGB}{166,206,227}

\definecolor{Paired52}{RGB}{31,120,180}

\definecolor{Paired5B}{RGB}{31,120,180}

\definecolor{Paired53}{RGB}{178,223,138}

\definecolor{Paired5C}{RGB}{178,223,138}

\definecolor{Paired54}{RGB}{51,160,44}

\definecolor{Paired5D}{RGB}{51,160,44}

\definecolor{Paired55}{RGB}{251,154,153}

\definecolor{Paired5E}{RGB}{251,154,153}

\definecolor{Paired61}{RGB}{166,206,227}

\definecolor{Paired6A}{RGB}{166,206,227}

\definecolor{Paired62}{RGB}{31,120,180}

\definecolor{Paired6B}{RGB}{31,120,180}

\definecolor{Paired63}{RGB}{178,223,138}

\definecolor{Paired6C}{RGB}{178,223,138}

\definecolor{Paired64}{RGB}{51,160,44}

\definecolor{Paired6D}{RGB}{51,160,44}

\definecolor{Paired65}{RGB}{251,154,153}

\definecolor{Paired6E}{RGB}{251,154,153}

\definecolor{Paired66}{RGB}{227,26,28}

\definecolor{Paired6F}{RGB}{227,26,28}

\definecolor{Paired71}{RGB}{166,206,227}

\definecolor{Paired7A}{RGB}{166,206,227}

\definecolor{Paired72}{RGB}{31,120,180}

\definecolor{Paired7B}{RGB}{31,120,180}

\definecolor{Paired73}{RGB}{178,223,138}

\definecolor{Paired7C}{RGB}{178,223,138}

\definecolor{Paired74}{RGB}{51,160,44}

\definecolor{Paired7D}{RGB}{51,160,44}

\definecolor{Paired75}{RGB}{251,154,153}

\definecolor{Paired7E}{RGB}{251,154,153}

\definecolor{Paired76}{RGB}{227,26,28}

\definecolor{Paired7F}{RGB}{227,26,28}

\definecolor{Paired77}{RGB}{253,191,111}

\definecolor{Paired7G}{RGB}{253,191,111}

\definecolor{Paired81}{RGB}{166,206,227}

\definecolor{Paired8A}{RGB}{166,206,227}

\definecolor{Paired82}{RGB}{31,120,180}

\definecolor{Paired8B}{RGB}{31,120,180}

\definecolor{Paired83}{RGB}{178,223,138}

\definecolor{Paired8C}{RGB}{178,223,138}

\definecolor{Paired84}{RGB}{51,160,44}

\definecolor{Paired8D}{RGB}{51,160,44}

\definecolor{Paired85}{RGB}{251,154,153}

\definecolor{Paired8E}{RGB}{251,154,153}

\definecolor{Paired86}{RGB}{227,26,28}

\definecolor{Paired8F}{RGB}{227,26,28}

\definecolor{Paired87}{RGB}{253,191,111}

\definecolor{Paired8G}{RGB}{253,191,111}

\definecolor{Paired88}{RGB}{255,127,0}

\definecolor{Paired8H}{RGB}{255,127,0}

\definecolor{Paired91}{RGB}{166,206,227}

\definecolor{Paired9A}{RGB}{166,206,227}

\definecolor{Paired92}{RGB}{31,120,180}

\definecolor{Paired9B}{RGB}{31,120,180}

\definecolor{Paired93}{RGB}{178,223,138}

\definecolor{Paired9C}{RGB}{178,223,138}

\definecolor{Paired94}{RGB}{51,160,44}

\definecolor{Paired9D}{RGB}{51,160,44}

\definecolor{Paired95}{RGB}{251,154,153}

\definecolor{Paired9E}{RGB}{251,154,153}

\definecolor{Paired96}{RGB}{227,26,28}

\definecolor{Paired9F}{RGB}{227,26,28}

\definecolor{Paired97}{RGB}{253,191,111}

\definecolor{Paired9G}{RGB}{253,191,111}

\definecolor{Paired98}{RGB}{255,127,0}

\definecolor{Paired9H}{RGB}{255,127,0}

\definecolor{Paired99}{RGB}{202,178,214}

\definecolor{Paired9I}{RGB}{202,178,214}

\definecolor{Paired101}{RGB}{166,206,227}

\definecolor{Paired10A}{RGB}{166,206,227}

\definecolor{Paired102}{RGB}{31,120,180}

\definecolor{Paired10B}{RGB}{31,120,180}

\definecolor{Paired103}{RGB}{178,223,138}

\definecolor{Paired10C}{RGB}{178,223,138}

\definecolor{Paired104}{RGB}{51,160,44}

\definecolor{Paired10D}{RGB}{51,160,44}

\definecolor{Paired105}{RGB}{251,154,153}

\definecolor{Paired10E}{RGB}{251,154,153}

\definecolor{Paired106}{RGB}{227,26,28}

\definecolor{Paired10F}{RGB}{227,26,28}

\definecolor{Paired107}{RGB}{253,191,111}

\definecolor{Paired10G}{RGB}{253,191,111}

\definecolor{Paired108}{RGB}{255,127,0}

\definecolor{Paired10H}{RGB}{255,127,0}

\definecolor{Paired109}{RGB}{202,178,214}

\definecolor{Paired10I}{RGB}{202,178,214}

\definecolor{Paired1010}{RGB}{106,61,154}

\definecolor{Paired10J}{RGB}{106,61,154}

\definecolor{Paired111}{RGB}{166,206,227}

\definecolor{Paired11A}{RGB}{166,206,227}

\definecolor{Paired112}{RGB}{31,120,180}

\definecolor{Paired11B}{RGB}{31,120,180}

\definecolor{Paired113}{RGB}{178,223,138}

\definecolor{Paired11C}{RGB}{178,223,138}

\definecolor{Paired114}{RGB}{51,160,44}

\definecolor{Paired11D}{RGB}{51,160,44}

\definecolor{Paired115}{RGB}{251,154,153}

\definecolor{Paired11E}{RGB}{251,154,153}

\definecolor{Paired116}{RGB}{227,26,28}

\definecolor{Paired11F}{RGB}{227,26,28}

\definecolor{Paired117}{RGB}{253,191,111}

\definecolor{Paired11G}{RGB}{253,191,111}

\definecolor{Paired118}{RGB}{255,127,0}

\definecolor{Paired11H}{RGB}{255,127,0}

\definecolor{Paired119}{RGB}{202,178,214}

\definecolor{Paired11I}{RGB}{202,178,214}

\definecolor{Paired1110}{RGB}{106,61,154}

\definecolor{Paired11J}{RGB}{106,61,154}

\definecolor{Paired1111}{RGB}{255,255,153}

\definecolor{Paired11K}{RGB}{255,255,153}

\definecolor{Paired121}{RGB}{166,206,227}

\definecolor{Paired12A}{RGB}{166,206,227}

\definecolor{Paired122}{RGB}{31,120,180}

\definecolor{Paired12B}{RGB}{31,120,180}

\definecolor{Paired123}{RGB}{178,223,138}

\definecolor{Paired12C}{RGB}{178,223,138}

\definecolor{Paired124}{RGB}{51,160,44}

\definecolor{Paired12D}{RGB}{51,160,44}

\definecolor{Paired125}{RGB}{251,154,153}

\definecolor{Paired12E}{RGB}{251,154,153}

\definecolor{Paired126}{RGB}{227,26,28}

\definecolor{Paired12F}{RGB}{227,26,28}

\definecolor{Paired127}{RGB}{253,191,111}

\definecolor{Paired12G}{RGB}{253,191,111}

\definecolor{Paired128}{RGB}{255,127,0}

\definecolor{Paired12H}{RGB}{255,127,0}

\definecolor{Paired129}{RGB}{202,178,214}

\definecolor{Paired12I}{RGB}{202,178,214}

\definecolor{Paired1210}{RGB}{106,61,154}

\definecolor{Paired12J}{RGB}{106,61,154}

\definecolor{Paired1211}{RGB}{255,255,153}

\definecolor{Paired12K}{RGB}{255,255,153}

\definecolor{Paired1212}{RGB}{177,89,40}

\definecolor{Paired12L}{RGB}{177,89,40}

\definecolor{Accent31}{RGB}{127,201,127}

\definecolor{Accent3A}{RGB}{127,201,127}

\definecolor{Accent32}{RGB}{190,174,212}

\definecolor{Accent3B}{RGB}{190,174,212}

\definecolor{Accent33}{RGB}{253,192,134}

\definecolor{Accent3C}{RGB}{253,192,134}

\definecolor{Accent41}{RGB}{127,201,127}

\definecolor{Accent4A}{RGB}{127,201,127}

\definecolor{Accent42}{RGB}{190,174,212}

\definecolor{Accent4B}{RGB}{190,174,212}

\definecolor{Accent43}{RGB}{253,192,134}

\definecolor{Accent4C}{RGB}{253,192,134}

\definecolor{Accent44}{RGB}{255,255,153}

\definecolor{Accent4D}{RGB}{255,255,153}

\definecolor{Accent51}{RGB}{127,201,127}

\definecolor{Accent5A}{RGB}{127,201,127}

\definecolor{Accent52}{RGB}{190,174,212}

\definecolor{Accent5B}{RGB}{190,174,212}

\definecolor{Accent53}{RGB}{253,192,134}

\definecolor{Accent5C}{RGB}{253,192,134}

\definecolor{Accent54}{RGB}{255,255,153}

\definecolor{Accent5D}{RGB}{255,255,153}

\definecolor{Accent55}{RGB}{56,108,176}

\definecolor{Accent5E}{RGB}{56,108,176}

\definecolor{Accent61}{RGB}{127,201,127}

\definecolor{Accent6A}{RGB}{127,201,127}

\definecolor{Accent62}{RGB}{190,174,212}

\definecolor{Accent6B}{RGB}{190,174,212}

\definecolor{Accent63}{RGB}{253,192,134}

\definecolor{Accent6C}{RGB}{253,192,134}

\definecolor{Accent64}{RGB}{255,255,153}

\definecolor{Accent6D}{RGB}{255,255,153}

\definecolor{Accent65}{RGB}{56,108,176}

\definecolor{Accent6E}{RGB}{56,108,176}

\definecolor{Accent66}{RGB}{240,2,127}

\definecolor{Accent6F}{RGB}{240,2,127}

\definecolor{Accent71}{RGB}{127,201,127}

\definecolor{Accent7A}{RGB}{127,201,127}

\definecolor{Accent72}{RGB}{190,174,212}

\definecolor{Accent7B}{RGB}{190,174,212}

\definecolor{Accent73}{RGB}{253,192,134}

\definecolor{Accent7C}{RGB}{253,192,134}

\definecolor{Accent74}{RGB}{255,255,153}

\definecolor{Accent7D}{RGB}{255,255,153}

\definecolor{Accent75}{RGB}{56,108,176}

\definecolor{Accent7E}{RGB}{56,108,176}

\definecolor{Accent76}{RGB}{240,2,127}

\definecolor{Accent7F}{RGB}{240,2,127}

\definecolor{Accent77}{RGB}{191,91,23}

\definecolor{Accent7G}{RGB}{191,91,23}

\definecolor{Accent81}{RGB}{127,201,127}

\definecolor{Accent8A}{RGB}{127,201,127}

\definecolor{Accent82}{RGB}{190,174,212}

\definecolor{Accent8B}{RGB}{190,174,212}

\definecolor{Accent83}{RGB}{253,192,134}

\definecolor{Accent8C}{RGB}{253,192,134}

\definecolor{Accent84}{RGB}{255,255,153}

\definecolor{Accent8D}{RGB}{255,255,153}

\definecolor{Accent85}{RGB}{56,108,176}

\definecolor{Accent8E}{RGB}{56,108,176}

\definecolor{Accent86}{RGB}{240,2,127}

\definecolor{Accent8F}{RGB}{240,2,127}

\definecolor{Accent87}{RGB}{191,91,23}

\definecolor{Accent8G}{RGB}{191,91,23}

\definecolor{Accent88}{RGB}{102,102,102}

\definecolor{Accent8H}{RGB}{102,102,102}

\makeatletter
\newcommand{\stepsubequation}[1][]{%
  \ifmeasuring@
  \else
    \refstepcounter{parentequation}%
    \protected@xdef\theparentequation{\arabic{parentequation}}%
    \ifdefined\theHparentequation
      \protected@xdef\theHparentequation{\arabic{parentequation}}%
    \fi
    \setcounter{equation}{0}%
    \if\relax\detokenize{#1}\relax\else
      \edef\@currentlabel{\theparentequation}%
      \ltx@label{#1}%
    \fi
  \fi
}
\makeatother

\newcommand*\diff{\mathop{}\!\mathrm{d}}

\newcommand\copyrighttext{%
    \footnotesize \textcopyright 2026 IEEE.  Personal use of this material is permitted.  Permission from IEEE must be obtained for all other uses, in any current or future media, including reprinting/republishing this material for advertising or promotional purposes, creating new collective works, for resale or redistribution to servers or lists, or reuse of any copyrighted component of this work in other works.
}

\newcommand\copyrightnotice{%
    \tikzexternaldisable
    \begin{tikzpicture}[remember picture,overlay]
        \node[anchor=south,yshift=10pt] at (current page.south) {\fbox{\parbox{\dimexpr\textwidth-\fboxsep-\fboxrule\relax}{\copyrighttext}}};
    \end{tikzpicture}%
    \tikzexternalenable
}

\IEEEoverridecommandlockouts                              
\usepackage{amsthm}
\newtheorem{remark}{Remark}

\title{\LARGE \bf
	MoCCA: A Movable Circle Probability of Collision Approximation
}

\author{Tobias Kern$^{1}$, Christian Birkner$^{1}$ 
	\thanks{$^{1}$CARISSMA Institute of Safety in Future Mobility, Technische Hochschule Ingolstadt, Ingolstadt, Germany, e-mail:
		{\tt\small \{tobiasbenjamin.kern, christian.birkner\}@thi.de}}%
}

\begin{document}    
	\maketitle
	
    \thispagestyle{empty}
	\pagestyle{empty}
    \copyrightnotice
	\begin{abstract}               

        In automated driving, crash mitigation is crucial to ensure passenger safety. Accurate avoidance requires precise knowledge of the object's position and orientation. However, sensor noise and occlusions often result in tracking and prediction uncertainties. To account for these uncertainties, estimating the Probability of Collision (POC) is a critical requirement. While Monte Carlo sampling is a common estimation technique, its high computational demand and stochastic nature often render it unsuitable for real-time applications. Analytical POC calculations are simplified by approximating vehicle geometries using circular bounds. While multi-circle approximations offer higher fidelity than a single circumscribed circle, they significantly increase computational complexity. This paper proposes a shape approximation algorithm, MoCCA\footnotemark[2], which utilizes a single circle for each vehicle, optimized to minimize the relative distance between them. MoCCA maintains a computational efficiency comparable to standard single-circle techniques while reducing over-conservatism. To address the potential underestimation of POC inherent in partial coverage, we establish an upper bound for the approximation error, demonstrating that it depends primarily on inter-vehicle distance and orientation variance. Furthermore, we introduce a safety distance margin that can be calibrated solely based on orientation variance.
	\end{abstract}
	

	
	\section{Introduction}

    In automated driving systems, the robust avoidance of detected objects is a fundamental requirement. For this, the poses, i.e., positions and orientations, of the surrounding objects have to be known. However, due to inherent sensor noise and limited knowledge of the automated driving system's surroundings, object detection, tracking, and prediction are subject to uncertainty. \\
    To account for this uncertainty, the probability of collision (POC) is a commonly used measure \cite{althoffModelBasedProbabilisticCollision2009,schreierIntegratedApproachManeuverBased2016,schwartingSafeNonlinearTrajectory2018,gouletProbabilisticConstraintTightening2022}. Historically, the POC calculations originate in orbital mechanics \cite{alfriendProbabilityCollisionError1999,kenchanCollisionProbabilityAnalyses2001,jamesleeforsterjrParametricAnalysisOrbital1992,pateraGeneralMethodCalculating2001,alfanoNumericalImplementationSpherical2005}, where the pose uncertainty is commonly modeled using a multivariate normal distribution, as the object is tracked over time \cite{alfriendProbabilityCollisionError1999,kenchanCollisionProbabilityAnalyses2001}. 
    Similarly, in the automated driving domain, the pose uncertainty is also modeled as normally distributed \cite{gouletProbabilisticConstraintTightening2022,tolksdorfFastCollisionProbability2024,hossamParticleBasedCollisionProbability2026}. To estimate the POC between two objects, the shape of the objects as well as the conditions of collision have to be known. Thus, a collision is defined as the overlap of two shapes. To accommodate arbitrary shapes, Monte Carlo simulation (MCS) is typically used \cite{schreierIntegratedApproachManeuverBased2016,hossamParticleBasedCollisionProbability2026,lambertCollisionProbabilityAssessment2008}. Although the shapes are approximated well, many samples are needed to estimate the POC accurately. Also, MCS may underapproximate the POC depending on which random samples are drawn, resulting in an undesired safety risk. To guarantee non-underapproximate behavior, analytical methods of approximating the POC were developed. As vehicle shapes can be considered rectangular, the POC calculation needs to solve a computationally intensive three-dimensional integral. To decrease the computational load for arbitrary shapes, the rectangular shape can be circumscribed by, e.g., a single circle. as shown in \cite{alfanoNumericalImplementationSpherical2005,dutoitProbabilisticCollisionChecking2011}. Using a single circumscribing circle provides the benefit of being orientation independent, reducing the number of integration variables to two. If the x- and y-coordinates are uncorrelated, the integral can be further reduced to 1D. But using a single circle to approximate a rectangle introduces conservatism. To reduce this conservatism, \cite{schwartingSafeNonlinearTrajectory2018} uses ellipses, which reduces the conservative POC estimation but only considers the positional uncertainty and neglects the orientation uncertainty. Whereas \cite{tolksdorfFastCollisionProbability2024} uses multiple overlapping circles to encapsulate the rectangular shape of the vehicle, considering also the orientation uncertainty. Although these methods offer high fidelity and outperform MCS in terms of speed, they remain computationally more expensive than the single circle methods.\\  
    To bridge this gap, this work presents MoCCA \footnote[2]{the code is hosted at \url{https://github.com/tok4623/MoCCA}}, a novel shape approximation utilizing a single circle that moves continuously along the vehicle's centerline. This approach enables a POC calculation that is nearly as computationally efficient as the traditional single circle methods. Given that two vehicles can only collide at their point of closest approach, we dynamically place one circle on the centerline of each vehicle, minimizing the inter-circle distance. While this reduces over-conservatism, it introduces the risk of underestimating the POC. Therefore, this paper also derives a theoretical upper bound for the resulting approximation error to ensure safety and reliability.
	
	\section{Problem Statement}
\begin{figure}
        \centering
        \includegraphics{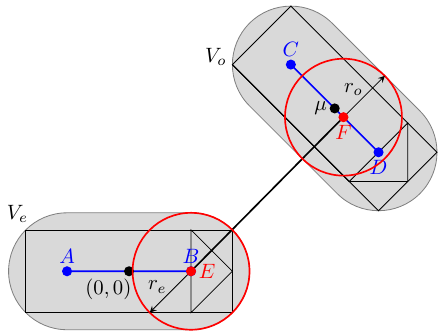}
        \caption{Visualization of the problem statement. The gray shadows visualize the possible envelope of the shape approximation. The blue lines indicate the centerlines along which the red circles around $E$ and $F$ can move.}
        \label{fig:ProblemStatement}
    \end{figure}

	Assume two vehicles, the ego vehicle $V_e$ and the opponent vehicle $V_o$, approaching each other, see Figure \ref{fig:ProblemStatement}.
	Let the center-point of $V_o$ be relatively located to the center-point of $V_e$ at $\mu=(\mu_x, \mu_y, \mu_\theta)$, where $\mu_x \in \mathbb{R}$ is the distance in $x$, the distance in $y$ is $\mu_y \in \mathbb{R}$, and $\mu_\theta \in [0, 2\pi)$ is the relative rotation of $V_o$ to $V_e$. Hence, detection algorithms aren't perfect, $\mu$ is assumed to be normally distributed with covariance matrix $\Sigma=\mathrm{diag}(\sigma_x^2, \sigma_y^2, \sigma_\theta^2) \in \mathbb{R}^{3\times 3}$. Additionally, let $l_o$ and $w_o$ be the length and width of $V_o$, and let $l_e$ and $w_e$ be the length and width of $V_e$. 
    We define two points $A=(x_A, y_A)^T \in \mathbb{R}^2$ and $B=(x_B, y_B)^T \in \mathbb{R}^2$ on the centerline of $V_e$. The points $A$ and $B$ can be placed symmetrically anywhere on the centerline. To ensure that the shape extends the same on each side, $A$ is located $\frac{w_e}{2}$ in front of the rear-end of $V_e$. And, on the opposite, $B$ is located at a distance of $\frac{w_e}{2}$ behind the front-end of $V_e$. Similarly, points $C$ and $D$ are defined for $V_o$.
    Let a point 
    \begin{align}
        E(t) &= l_E(t)\begin{bmatrix}
            1\\0
        \end{bmatrix}
    \end{align}
    lie on $\vv{AB}$, with 
    \begin{align}
        l_E(t) = t\frac{l_e-w_e}{2}, \label{eq:l_E}
    \end{align}
    where $t\in[-1,1]$ indicates the normalized shift of $E(t)$ from the center-point of $V_e$ towards point $A$ or $B$. Also, let 
    \begin{align}
        F(s) &=\begin{bmatrix}
         \mu_x\\
         \mu_y
     \end{bmatrix} + l_F(s)
     \begin{bmatrix}
         \cos{\mu_\theta} \\
         \sin{\mu_\theta}
     \end{bmatrix}, 
    \end{align}
    respectively, be a point that lies on $\vv{CD}$, with
    \begin{align}
        l_F(s) = s\frac{l_o-w_o}{2}, \label{eq:l_F}
    \end{align}
    where $s\in[-1,1]$ indicates the normalized shift from the center-point of $V_o$ to point $C$ or $D$. Also, let the distance between points $E(t)$ and $F(s)$ be minimal for $t=t_\mathrm{min}$ and $s=s_\mathrm{min}$. 
    Additionally, we declare a single circle for $V_e$ at point $E(t)$ of radius $r_e=\sqrt{\frac{w_e^2}{2}}$ and as well for $V_o$ at point $F(s)$ of radius $r_o=\sqrt{\frac{w_o^2}{2}}$. \\
    Hence, the point $F(s)$ may be shifted from the center of $V_o$ by $l_F(s)$ (\ref{eq:l_F}), and the orientation used in the translation process is uncertain, the covariance $\Sigma_F(s)$ at point $F(s)$ needs to be determined by propagating the variance of $\Sigma$ using the Jacobian 
    \begin{align}
        J_F(s) &=\begin{bmatrix}
            1 & 0 & -l_F(s)\sin{\mu_\theta} \\
            0 & 1 & \hspace{1em}l_F(s)\cos{\mu_\theta}
        \end{bmatrix}
    \end{align}
    of $F(s)$. Finally, $\Sigma_F$ can be derived by
    \begin{align}
        \Sigma_F(s) &= J_F(s)\,\Sigma\,J_F(s)^T. \label{eq:sigma_prop}
    \end{align}
    In the following the points $E(t)$ at {$t = \nolinebreak t_\mathrm{min}$}, $F(s)$ at $s=s_\mathrm{min}$, as well as $\Sigma_F(s)$ at $s=s_\mathrm{min}$ will be noted without their dependency on $t$ or $s$.

    \subsection{Collision Probability of Single Circles}
    A collision between two shapes occurs if these two shapes overlap.
    The POC $P_\mathrm{coll}$ is, therefore, the integration over the circular area with radius $r_c = r_e+r_o$ of the two dimensional probability density function (PDF)
    \begin{multline}
        P_\mathrm{coll} = \iint\limits_{|X|^2 \leq r_c^2} \frac{1}{2\pi\sqrt{\det(\Sigma_F)}} \exp{\left(-\frac{1}{2}\right. } \\
         \left.(X-\mu_F)^T\Sigma_F^{-1}(X-\mu_F)\vphantom{\frac{}{}}\right) \diff X \label{eq:P_coll}   
    \end{multline}
    centered around point $E$, where $\mu_F$ is the mean relative position of $F$ with respect to $E$ in $x$ and $y$, and ${X=(x, y)^T\in \mathbb{R}^2}$ is a point in $x$ and $y$. 

    \subsection{Approximation Error} 
\begin{figure}
        \centering
            \includegraphics{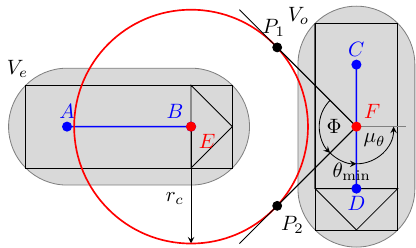}
        \caption{Depiction of determining the underapproximation error of our approach.}
        \label{fig:approx_error}
    \end{figure}

     As the vehicular shapes are partially covered using the circles around $E$ and $F$, as shown in Figure \ref{fig:ProblemStatement}, the algorithm underapproximates the POC. Hence, the shape not covered by the circles may collide if the orientation differs. Therefore, to upper-bound the error, the probability of $V_o$'s orientations, which lead to a collision with $V_e$, has to be found. In Figure \nolinebreak\ref{fig:approx_error}, the setup is shown. The points $P_1$ and $P_2$ are the contact points between the tangential lines of the circle with radius $r_c$ through $F$. The angle $\theta_\mathrm{min}$ is the minimum possible angle that would result in a collision. The angle $\Phi$ is the angular difference between $\theta_\mathrm{min}$ and  $\theta_\mathrm{max}$, indicating the range of possible collision angles. With that, it is shown that, if $V_o$ crosses slightly in front of $V_e$ ($|\mu_F|=r_c$), the angle $\Phi$ opens up to $180\text{\,°}$. Due to symmetry, i.e. the collision may occur in clockwise and counter-clockwise rotation, this results in a $100\text{\,\%}$ approximation error $e_\mathrm{approx}$. Meaning, if the orientation differs from its mean, $V_o$ and $V_e$ are colliding without the algorithm noticing.
 	Calculating
    \begin{align}
        P_2 &= 
        \begin{bmatrix}
            \cos\Theta & -\sin\Theta \\
            \sin\Theta & \cos\Theta
        \end{bmatrix}
        \, \frac{1}{|\mu_F|}\,
        \begin{bmatrix}
            r_c^2\\
            -r_c\,\sqrt{|\mu_F|^2-r_c^2}
        \end{bmatrix} \label{eq:P1_approx_error} \\
        \text{with} \nonumber \\
        \Theta &= \tan^{-1}{\left(\frac{\mu_{F,y}}{\mu_{F,x}}\right)}
    \end{align}
    is sufficient as $P_1$ is always symmetrical to $P_2$.
    Next, defining angular difference
    \begin{align}
        \Phi &= 2\sin^{-1}{\left(\frac{r_c}{|\mu_F|}\right)}
    \end{align}
    between $\theta_\mathrm{min}$ and $\theta_\mathrm{max}$, i.e.
    \begin{align}
        \theta_{\mathrm{max}} &= \theta_\mathrm{min} + \Phi.
    \end{align}
    Using the symmetry of $P_1$ and $P_2$, as well as the orthogonality $\vv{CD}\perp\vv{EF}$, results in
    \begin{align}
        \theta_\mathrm{min} &= \frac{\pi-\Phi}{2}. 
    \end{align}
    Due to symmetry of rotation, the definition of $\theta_\mathrm{min}$ relative to $\mu_\theta$, and the periodicity of the angular distribution, we define $e_\mathrm{approx}$ as:
    \begin{multline}
        e_\mathrm{approx} = 2\sum_{\beta=-\infty}^{\infty}\int_{\theta_\mathrm{min}+2\pi\beta}^{\theta_\mathrm{max}+2\pi\beta} \frac{1}{\sqrt{2\pi\sigma_\theta^2}} \\
        \exp{\left(-\left(\frac{\theta}{\sqrt{2}\sigma_\theta}\right)^2\right)} \diff \theta. \label{eq:approx_error}
    \end{multline}
    \begin{remark}
        If $|\mu_F|<r_c$, the circles overlap, and therefore no probable rotation can change this collision.\\ 
        For $\lim_{|\mu_F| \to \infty} \Phi = 2\sin^{-1}(0)=0$, meaning that at infinite distance the angular interval within an undetected collision may occur is $0$, and with that also $e_\mathrm{approx}=0$. \\
        But, if $|\mu_F| = r_c$ the denominator of the $\sin^{-1}$ in $\Phi$ becomes $r_c$. Therefore $\lim_{|\mu_F| \to r_c} \Phi = \pi $, resulting in any rotation between $+\pi$ and $-\pi$ can lead to a possible collision. As the integral of a wrapped normal distribution (\ref{eq:approx_error}) over an interval of length of $2\pi$ equals $1$ \cite{mardiaDirectionalStatistics2010}, the approximation error results here in $e_\mathrm{approx}=1$.
    \end{remark}
    The upper bound of $e_\mathrm{approx}$ at the distance $|\mu_F|=r_c$ is $100\text{\,\%}$. Thus, an additional safety distance $d_s$ shall be added, such that $r_c=r_e+r_o+d_s$, to set the maximum possible underapproximation based on $\sigma_\theta$. 
    For $6\,\sigma_\theta \leq \pi$, meaning the probability is nearly completely contained in one single period, we could redefine the approximation error 
    \begin{align}
        e_\mathrm{approx} &\approx 1 - \int_{0}^{\left(\frac{\theta_\mathrm{min}}{\sigma_\theta}\right)^2}\chi_1^2(\theta)\diff \theta.
    \end{align}
    If we now leverage the relationship between the normal distribution, the Mahalanobis distance and the $\chi_1^2$-distribution \cite{mardiaMultivariateAnalysis2023}, we find for a Mahalanobis distance of $n$ 
    \begin{align}
        n\,\sigma_\theta &= \theta_\mathrm{min} = \frac{\pi}{2} - \sin^{-1}\left(\frac{r_e+r_o}{r_e+r_o+d_s}\right),
    \end{align}
    the safety distance 
    \begin{align}
        d_s              &= \frac{r_e+r_o}{\sin\left(\frac{\pi}{2} - n\,\sigma_\theta\right)}-r_e-r_o.
    \end{align}
    For $n=3$, following the $3\,\sigma$-rule for normal distributions, the maximum approximation error can be reduced to $0.3\text{\,\%}$ if the additional safety distance $d_s$ is added to $r_e+r_o$. 
	
	\section{Methodology}
    This section is structured as follows: at first, the calculation of points $E$ and $F$ is shown. Then, the integration method is introduced. At last, a simulation example is set up running on an Intel i7-12800HX with $32\text{\,GB}$ RAM.

    \subsection{Find Minimal Distance Points}
    As mentioned, the points $E$ and $F$ lie on $\vv{AB}$ and $\vv{CD}$ respectively. 
    The line segments $\vv{AB}$ and $\vv{CD}$ can be described as 
    \begin{alignat}{2}
        \vv{AB} = g(t) &= l_E(t) 
        \begin{bmatrix}
         1\\
         0
        \end{bmatrix} \text{,} \\
        \vv{CD} = h(s) &=\begin{bmatrix}
         \mu_x\\
         \mu_y
        \end{bmatrix} + l_F(s)
        \begin{bmatrix}
         \cos\mu_\theta\\
         \sin\mu_\theta
        \end{bmatrix}
    \end{alignat}
    with $t \in [-1,1]$ and $s \in [-1,1]$. Therefore, the distance $d(t,s)$ is defined as $|h(s)-g(t)|$ resulting in
    \begin{subequations}
        \begin{align}
            c_1 &= (x_C-x_A)^2 + y_C^2 \\
            c_2 &= 2((x_C-x_A)(x_D-x_C) + y_C\,(y_D-y_C)) \\
            c_3 &= 2(x_C-x_A)(x_B-x_A) \\ 
            c_4 &= 2(x_D-x_C)(x_B-x_A) \\
            c_5 &= (x_D-x_C)^2 + (y_D-y_C)^2 \\
            c_6 &= (x_B-x_A)^2 \\
            d(t,s) &= \sqrt{c_1 + s\,c_2 - t\,c_3 - s\,t\,c_4 + s^2\,c_5 + t^2\,c_6}\text{,} \label{eq:dist_g_h}
        \end{align}
    \end{subequations}
    where $x$ indicates the x-coordinate and $y$ the y-coordinate of the point in the subscript.
    To find the points $E$ and $F$, the optimization problem
    \begin{alignat}{2}
        \min_{t,s}{}\quad   & d(t,s) = |h(s)-g(t)|        \nonumber\\
        \text{s.\,t.}\quad  & -1 \leq t \leq 1  \label{eq:min_dist}\\
                            & -1 \leq s \leq 1  \nonumber
    \end{alignat}
    has to be solved.
    In a two-dimensional space, the lines can either intersect or be parallel. Therefore, the solution is either the intersection point or infinitely many solutions. As the variables $s$ and $t$ are limited between $-1$ and $1$, and the equations resulting from (\ref{eq:min_dist}) are linear, two solution pairs 
    
    \begin{subequations}
        \begin{align}
            s_1 &= \mathrm{clamp}\left(\frac{c_3\,c_4 - 2\,c_2\,c_6}{4\,c_5\,c_6 - c_4^2},\,\ -1,\ 1\right) \text{,} \\
            t_1 &= \mathrm{clamp}\left(\frac{c_2 + 2\,s_1\,c_5}{c_4},\ -1,\ 1\right)\text{,}
        \end{align}
    \end{subequations}
    and
    \begin{subequations}
        \begin{align}
            t_2 &= \mathrm{clamp}\left(\frac{2\,c_3\,c_5 - c_2\,c_4}{4\,c_5\,c_6 - c_4^2},\ -1,\ 1\right) \text{,} \\
            s_2 &= \mathrm{clamp}\left(\frac{t_2\,c_4 - c_2}{2\,c_5},\ -1,\ 1\right)\text{,} 
        \end{align}
    \end{subequations}
    \noindent
    for non-parallel $h(s)$ and $g(t)$ are found.
    It is worth mentioning that $t_1$,\ $s_1$,\ $t_2$, and $s_2$ are also limited between $-1$ and $1$.
    Therefore, the points on $\vv{AB}$ and $\vv{CD}$ are 
    \begin{align}
        E &= \begin{cases}
                g(t_1) & d(t_1, s_1) < d(t_2,s_2) \\
                g(t_2) & \text{otherwise}
             \end{cases}, \\
        F &= \begin{cases}
                h(s_1) & d(t_1, s_1) < d(t_2,s_2) \\
                h(s_2) & \text{otherwise}
             \end{cases} \text{.}
    \end{align}
    For parallel $g(t)$ and $h(s)$ the distances of each point ($A$,$B$,$C$,$D$) to the respective other line have to be calculated, i.e. for $A$ and $B$ the distance to line $h(s)$, and for $C$ and $D$ the distance to line $g(t)$, see (\ref{eq:d_par_A})-(\ref{eq:d_par_D}).
    \begin{align}
        \vv{A\,h(s_A)}\cdot\vv{CD} &= 0\text{,} & d_A &= |\vv{A\,h(s_A)}| \label{eq:d_par_A}\\
        \vv{B\,h(s_B)}\cdot\vv{CD} &= 0\text{,} & d_B &= |\vv{B\,h(s_B)}| \label{eq:d_par_B}\\
        \vv{C\,g(t_C)}\cdot\vv{AB} &= 0\text{,} & d_C &= |\vv{C\,g(t_C)}| \label{eq:d_par_C}\\
        \vv{D\,g(t_D)}\cdot\vv{AB} &= 0\text{,} & d_D &= |\vv{D\,g(t_D)}| \label{eq:d_par_D}
    \end{align}
    where $s_A\in[-1,1]$, $s_B\in[-1,1]$, $t_C\in[-1,1]$, and ${t_D \in [-1,1]}$. Therefore, the necessary points are
    \begin{align}
        E &= \begin{cases}
                A      & \min(d_A,d_B,d_C,d_D)==d_A\\
                B      & \min(d_A,d_B,d_C,d_D)==d_B\\
                g(t_C) & \min(d_A,d_B,d_C,d_D)==d_C\\
                g(t_D) & \text{otherwise}
            \end{cases} \text{,} \label{eq:E_par} 
    \end{align}
    \begin{align}
        F &= \begin{cases}
                h(s_A) & \min(d_A,d_B,d_C,d_D)==d_A\\
                h(s_B) & \min(d_A,d_B,d_C,d_D)==d_B\\
                C      & \min(d_A,d_B,d_C,d_D)==d_C\\
                D      & \text{otherwise}
            \end{cases} \label{eq:F_par}
    \end{align}
    in case the lines $g(t)$ and $h(s)$ are parallel.

    \subsection{Integration Method}  
    As defined by \cite{pateraGeneralMethodCalculating2001,alfanoNumericalImplementationSpherical2005}, the POC is calculated by (\ref{eq:P_coll}).
    If $x$ and $y$ are uncorrelated, the PDF $p_E(x,y)$ can be split into 
    \begin{align}
        p_{E,x}(x) &= \frac{1}{\sqrt{2\,\pi}\sigma_{F,x}} \exp{-\frac{(x-\mu_{F,x})^2}{2\,\sigma_{F,x}^2}}\text{,}
    \end{align}
    \begin{align}
        p_{E,y}(y) &= \frac{1}{\sqrt{2\,\pi}\sigma_{F,y}} \exp{-\frac{(y-\mu_{F,y})^2}{2\,\sigma_{F,y}^2}}\text{.}
    \end{align}
    Furthermore, the integral of (\ref{eq:P_coll}) can be split into
    \begin{align}
        P_\mathrm{coll} &= \int_{-r_c}^{r_c} p_{E,x}(x) \int_{-\sqrt{r_c^2-x^2}}^{\sqrt{r_c^2-x^2}} p_{E,y}(y) \diff y \diff x \text{.}
    \end{align}
    Using the cumulative distribution function (CDF), the inner integral can be solved to
    \begin{multline}
        P_\mathrm{coll} = \frac{1}{2} \int_{-r_c}^{r_c} p_{E,x}(x) 
        \left[\mathrm{erf}\left(\frac{\sqrt{r_c^2-x^2}-\mu_{F,y}}{\sqrt{2}\,\sigma_{F,y}}\right) \right. -\\
        \left. \mathrm{erf}\left(\frac{-\sqrt{r_c^2-x^2}-\mu_{F,y}}{\sqrt{2}\,\sigma_{F,y}}\right)\right] \diff x \label{eq:POC_simplified} \text{.}
    \end{multline}
    Notably (\ref{eq:POC_simplified}) has to be solved numerically.
    But, $x$ and $y$ are not uncorrelated as described in equation (\ref{eq:sigma_prop}). Thus, the PDF has to be decorrelated to use equation (\ref{eq:POC_simplified}). As $\Sigma_F$ is symmetric the problem can be rotated such that the axes align with the eigenvectors of $\Sigma_F$ \cite{gouletProbabilisticConstraintTightening2022}.

    \subsection{Simulation Example}
\begin{figure}
    \centering
        \begin{tabular}{cc}
            Scenario A & Scenario B \\
            \includegraphics{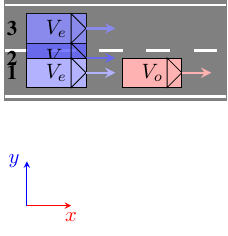}  &  \includegraphics{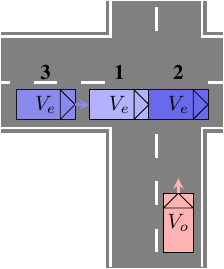}
        \end{tabular}
        \caption{Scenario \textbf{A}) testcase \textbf{1}: Head-on colliding vehicles at $100\text{\,\%}$ overlap; testcase \textbf{2}: Head-on colliding vehicles at $50\text{\,\%}$ overlap; testcase \textbf{3}: $V_e$ passing $V_o$ at a y-distance of $1.1\text{\,m}$; Scenario \textbf{B}) testcase \textbf{1}: $V_o$ crosses $V_e$ at an x-distance of $1.1\text{\,m}$; testcase \textbf{2}: $V_o$ colliding sideways with the center of $V_e$; testcase \textbf{3}: $V_e$ and $V_o$ approach the crossing at the same speed from the same distance.}
        \label{fig:Scenarios}
    \end{figure}

    Two scenarios, shown in Figure \ref{fig:Scenarios}, are used to check the calculation time as well as the accuracy of the proposed method. Scenario A is a passing scenario, and scenario B is a crossing scenario. For scenario A, three cases are tested: $100\text{\,\%}$ overlap, $50\text{\,\%}$ overlap and passing at a distance of half the width of the vehicle. In scenario B, also three cases are tested: $V_o$ passes in front of $V_e$ at a distance of half the width of the vehicle, $V_o$ drives through the still standing $V_e$, and $V_o$ and $V_e$ drive through each other. For all test cases $\Sigma=[0.25\text{\,m}^2, 0.25\text{\,m}^2, 0.25 \text{\,rad}^2]$ is assumed, the length of $V_e$ and $V_o$ is set to $5\text{\,m}$, and the width of both vehicles is fixed at $2.2\text{\,m}$. For comparison, MCS and the algorithms of \cite{pateraGeneralMethodCalculating2001, tolksdorfFastCollisionProbability2024} are evaluated. The Monte Carlo simulation is conducted using $10\text{\,k}$ samples. For the numerical integration of (\ref{eq:POC_simplified}), used in \cite{pateraGeneralMethodCalculating2001, tolksdorfFastCollisionProbability2024}, and MoCCA $80$ substeps are used. All methods are implemented using CasADi's Python API \cite{anderssonCasADiSoftwareFramework2019}.

	\section{Results and Discussion}
\begin{figure}
	    \centering
	    \includegraphics{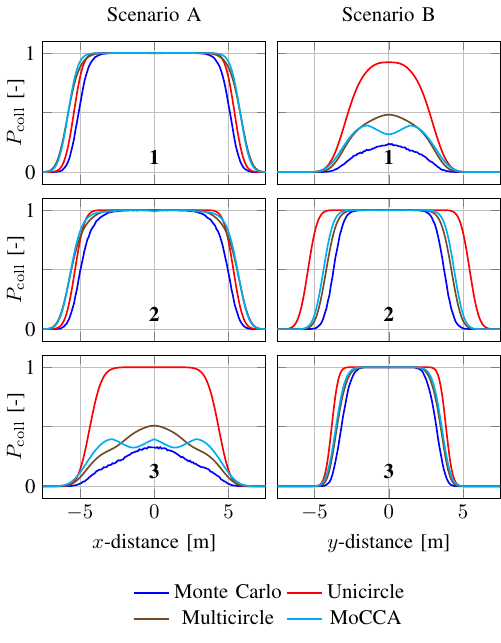}
	    \caption{Collision probabilities for each scenario (left: Scenario A; right: Scenario B) and testcase (from top to bottom: \textbf{1}--\textbf{3}).}
	    \label{fig:results_poc}
	\end{figure}

\begin{figure}
        \centering
        \includegraphics{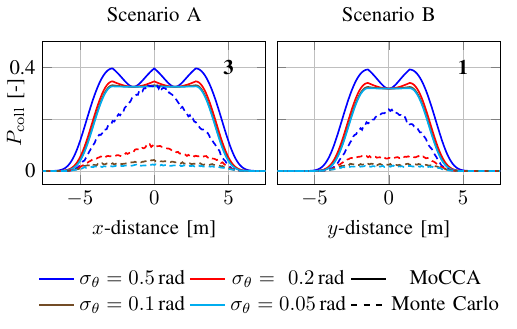}
        \caption{Collision probabilities for testcase 3 of Scenario A and testcase 1 of Scenario B with a variation on $\sigma_\theta$. For $\sigma_\theta=0.5\text{\,rad}$, the curves are the same as in Figure \ref{fig:results_poc}.}
        \label{fig:results_poc_sigma_theta}
    \end{figure}

	The results of the instantaneous POC calculation over the lateral and longitudinal distances, respectively, are shown in Figure \ref{fig:results_poc}. 
    For testcase 1 of scenario A, the unicircle method has the least deviation from the rectangular MCS. The MCS is considered the ground truth. The multicircle method and our method behave similarly up to an x-distance of $\pm5.5\text{\,m}$. Due to the error propagation of (\ref{eq:sigma_prop}), our approach shows a higher collision probability than the multicircle approach, if the vehicles are closer than $5.5\text{\,m}$. The early increase in collision probability of the circular methods can be explained by the different radii and positions of the circles, whereby the multi-circle method and our method have a similar circle radius due to the parameterization. The collision probability of the unicircle method increases last, as the unicircle method reaches not as far to the front and back as the other two approximations, $2.73\text{\,m}$ versus $3.05\text{\,m}$ (multicircle) and $2.95\text{\,m}$ (MoCCA), measured at the corresponding vehicle's center.\\
    For testcase 2 of scenario A, minor differences to testcase 1 of scenario A are observable. The collision probability increases more slowly for all approaches except the unicircle method. That is due to the higher possibility that the vehicle is oriented such that no collision occurs. Our approach shows 2 local maxima between $-2.5\text{\,m}$ and $0\text{\,m}$, and between $0\text{\,m}$ and $2.5\text{\,m}$. These local maxima originate from the error propagation (\ref{eq:sigma_prop}), as the area of the collision circle with radius $r_c$ around $E$ includes $\mu_F$, the collision probability increases the smaller the variance. The cusp at the x-distance of $0\text{\,m}$ results from the jump of $E$ from $B$ to $A$, respectively $F$ jumps from $D$ to $C$, see (\ref{eq:E_par}) and (\ref{eq:F_par}.\\
    In testcase 3 of scenario A, the cusp is better observable. Instead of local maxima, local minima are observable. Hence, $V_e$ is passing $V_o$ at a further distance, the probability inside the collision circle with radius $r_c$ around $E$ is reducing now, as $\mu_F$ is not inside this area anymore. The dependency on $\sigma_\theta$ of the magnitude of these extrema is visualized in Figure \ref{fig:results_poc_sigma_theta}. The maximum approximation error is $49.6\text{\,\%}$ for the inter-circle distance of $0.19\text{\,m}$ and $\sigma_\theta=0.5\text{\,rad}$. The collision probability of the multicircle method follows the contour of the MCS collision probability scaled by a positive factor. The unicircle method indicates still a certain collision.\\
    In testcase 1 of scenario B, a small deviation between the multicircle and our approach up to a y-distance of $\pm1.9\text{\,m}$ is observable. The local minimum at a y-distance of $0\text{\,m}$ is explained by the propagation of the covariance matrix (\ref{eq:sigma_prop}). If the covariance matrix had not been propagated, the curve would flatten at the probability of the local minimum, indicating the underapproximation, see Figure \ref{fig:results_poc_sigma_theta}. The maximum underapproximation error is here also $49.6\text{\,\%}$ as it is the same inter-circle distance and $\sigma_\theta$. The POC of the multicircle approach follows the contour of the POC of the MCS method again scaled by a positive factor, like in testcase 3 of scenario A. The unicircle method indicates the highest POC, as the overapproximation at the sides is the highest of all other methods, $1.63\text{\,m}$ shape extension to one side versus $0.28\text{\,m}$ (multicircle) and $0.46\text{\,m}$ (MoCCA).\\
    Testcase 2 of scenario B highlights the dependence of POC on the considered radii. Hence, the covariance matrix in the y-direction is here the same for all algorithms. The difference when the POC increases between the unicircle and MCS method shows the conservative approximation of the unicircle method at the sides of the rectangle, similar to testcase 3 of scenario A. \\
    Testcase 3 of scenario B differs by the direction of entry and exit, and by the effect of error propagation from testcase 2 of scenario B, which is why the POC rises later and falls earlier that in testcase 2.
    
    \noindent For Figure \ref{fig:results_poc_sigma_theta}, the variances $\sigma_x^2$ and $\sigma_y^2$ were kept at $0.25\text{\,m}^2$ and only $\sigma_\theta$ was changed between $0.05\text{\,rad}$ and $0.5\text{\,rad}$ to showcase the dependency of the local maxima or minima on $\sigma_\theta$.
    As mentioned before, the smaller the orientation uncertainty $\sigma_\theta$, the smaller the amplitude of the extrema and nearly flatten out at $\sigma_\theta<0.1\text{\,rad}$. The residual $P_\mathrm{coll}=0.325$ of MoCCA and 
    $P_\mathrm{coll}=0.023$ of the MCS resides solely on the positional variance. The difference is explained by the shape extension to the front and side.  

    \begin{table}[]
        \centering
        \caption{Computation times for the algorithms over $10\text{\,k}$ simulations. The rectangular Monte Carlo simulation utilizes $10\text{\,k}$ samples. The circular approaches use $80$ substeps to integrate over the distribution.}
        \label{tab:results_timing}
        \begin{tabular}{l|r|r|r}
            \toprule
            Method               &             Median &               Mean & Standard Deviation\\ 
            \midrule 
            Monte Carlo Sampling &  $13.4\text{\,ms}$ &  $13.5\text{\,ms}$ &   $0.7\text{\,ms}$\\
            Unicircle            &  $36.8\text{\,µs}$ &  $38.0\text{\,µs}$ &   $4.7\text{\,µs}$\\
            Multicircle          & $593.0\text{\,µs}$ & $601.0\text{\,µs}$ &  $33.9\text{\,µs}$\\
            MoCCA                &  $43.3\text{\,µs}$ &  $44.6\text{\,µs}$ &   $4.9\text{\,µs}$\\
            \bottomrule
        \end{tabular}
    \end{table}
    \noindent Table \ref{tab:results_timing} shows the calculation times for each algorithm's iteration. The mean calculation time of the MCS method, including the sample generation, is with $13.5\text{\,ms}$ approximately $22.5$ times slower than the multi-circle approach at $601\text{\,µs}$, which is by itself $13.5$ times slower than MoCCA at $44.6\text{\,µs}$. The unicircle method is still the fastest at $38\text{\,µs}$. 
    \begin{figure}
	    \centering
	    \includegraphics{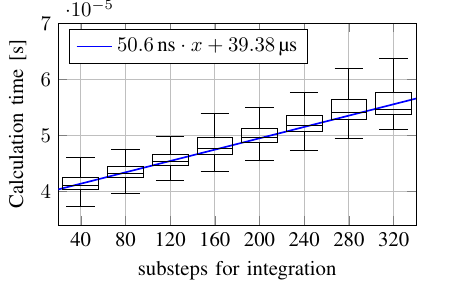}
	    \caption{Calculation times for MoCCA across different integration substep counts. Blue line: linear regression for the median calculation time with R$^2=0.993$.}
	    \label{fig:results_various_substeps}
	\end{figure}
    \noindent Figure \ref{fig:results_various_substeps} shows the dependency of the calculation time on the number of substeps used for solving the integration of \ref{eq:POC_simplified}. The blue line in Figure \ref{fig:results_various_substeps} represents the linear regression of the median calculation times with a fitness of R$^2=0.993$.
    
    
	\section{Conclusion and Future Work}
    We proposed MoCCA, a novel algorithm that estimates the probability of collision, assuming the collision can only occur between the two closest points. For that, we determine these points on the centerlines of two vehicles or objects. With that, the used collision circles may be set smaller than with the unicircle method, which overestimates rectangular objects heavily on the longer side of the rectangle. In terms of calculation speed, it shows that the unicircle method is only $6.6\text{\,µs}$ faster than MoCCA, which computes in $44.6\text{\,µs}$. In comparison with the Monte Carlo simulation approach, MoCCA is $303$ times faster. \\
    Also, we showed the main disadvantage of MoCCA on underestimating the probability of collision, but also upper-bound this approximation error by introducing an additional safety distance. As the upper bound of the approximation error is dependent on the variance of orientation, setting the safety distance accordingly leads to a confidence-based safety distance. If additional accuracy is needed, the multicircle approach of \cite{tolksdorfFastCollisionProbability2024} is a good trade-off of computation time and approximation accuracy. Another limitation of MoCCA is that if both vehicles are parallel, the selection of the reference points is not ideal, resulting in an unsteady collision probability calculation. \\
    This limitation could be worked on in the future by using, e.g, the Mahalanobis distance instead of the Euclidean distance to determine the points $E$ and $F$, like proposed in \cite{modeniniRelationsCollisionProbability2022}. With regard to the upper bound of the approximation error, the boundaries might be tightened further by considering also the distance between the vehicles as well as the shifts of the circles from the center of each vehicle. Another open research question is how fast the computation has to be for collision avoidance.

	\bibliography{references}         

@book{mardiaMultivariateAnalysis2023,
	address = {Hoboken, NJ},
	edition = {Second edition},
	series = {Wiley series in probability and statistics},
	title = {Multivariate analysis},
	isbn = {978-1-118-89251-0 978-1-118-89252-7},
	publisher = {Wiley},
	author = {Mardia, K. V. and Kent, J. T. and Bibby, J. M.},
	year = {2023},
}

@article{hossamParticleBasedCollisionProbability2026,
	title = {A {Particle}-{Based} {Collision} {Probability} {Estimation} {Framework} for {Uncertainty}-{Aware} {Risk} {Evaluation} in {Autonomous} {Vehicles}},
	volume = {14},
	copyright = {https://creativecommons.org/licenses/by/4.0/legalcode},
	issn = {2169-3536},
	url = {https://ieeexplore.ieee.org/document/11433426/},
	doi = {10.1109/ACCESS.2026.3672985},
	urldate = {2026-05-06},
	journal = {IEEE Access},
	author = {Hossam, Abdallah and Jiménez-Bermejo, Víctor and Villagra, Jorge and Navas, Francisco and Milanés, Vicente},
	year = {2026},
	pages = {39912--39925},
}

@article{althoffModelBasedProbabilisticCollision2009,
	title = {Model-{Based} {Probabilistic} {Collision} {Detection} in {Autonomous} {Driving}},
	volume = {10},
	copyright = {https://ieeexplore.ieee.org/Xplorehelp/downloads/license-information/IEEE.html},
	issn = {1524-9050, 1558-0016},
	url = {http://ieeexplore.ieee.org/document/4895669/},
	doi = {10.1109/TITS.2009.2018966},
	number = {2},
	urldate = {2026-05-06},
	journal = {IEEE Transactions on Intelligent Transportation Systems},
	author = {Althoff, M. and Stursberg, O. and Buss, M.},
	month = jun,
	year = {2009},
	pages = {299--310},
}

@article{dutoitProbabilisticCollisionChecking2011,
	title = {Probabilistic {Collision} {Checking} {With} {Chance} {Constraints}},
	volume = {27},
	copyright = {https://ieeexplore.ieee.org/Xplorehelp/downloads/license-information/IEEE.html},
	issn = {1552-3098, 1941-0468},
	url = {http://ieeexplore.ieee.org/document/5738354/},
	doi = {10.1109/TRO.2011.2116190},
	number = {4},
	urldate = {2026-05-06},
	journal = {IEEE Transactions on Robotics},
	author = {Du Toit, Noel E. and Burdick, J. W.},
	month = aug,
	year = {2011},
	pages = {809--815},
}

@book{mardiaDirectionalStatistics2010,
	address = {Chichester New York},
	series = {Wiley series in probability and statistics},
	title = {Directional statistics},
	isbn = {978-0-471-95333-3 978-0-470-31781-5 978-0-470-31697-9},
	language = {eng},
	publisher = {J. Wiley},
	editor = {Mardia, Kantilal V. and Jupp, Peter E.},
	year = {2010},
}

@article{anderssonCasADiSoftwareFramework2019,
	title = {{CasADi}: a software framework for nonlinear optimization and optimal control},
	volume = {11},
	issn = {1867-2949, 1867-2957},
	shorttitle = {{CasADi}},
	url = {http://link.springer.com/10.1007/s12532-018-0139-4},
	doi = {10.1007/s12532-018-0139-4},
	language = {en},
	number = {1},
	urldate = {2026-05-04},
	journal = {Mathematical Programming Computation},
	author = {Andersson, Joel A. E. and Gillis, Joris and Horn, Greg and Rawlings, James B. and Diehl, Moritz},
	month = mar,
	year = {2019},
	pages = {1--36},
}

@article{modeniniRelationsCollisionProbability2022,
	title = {Relations {Between} {Collision} {Probability}, {Mahalanobis} {Distance}, and {Confidence} {Intervals} for {Conjunction} {Assessment}},
	volume = {59},
	issn = {0022-4650},
	url = {https://arc.aiaa.org/doi/10.2514/1.A35234},
	doi = {10.2514/1.A35234},
	number = {4},
	urldate = {2026-02-03},
	journal = {Journal of Spacecraft and Rockets},
	publisher = {American Institute of Aeronautics and Astronautics},
	author = {Modenini, Dario and Curzi, Giacomo and Locarini, Alfredo},
	month = jul,
	year = {2022},
	pages = {1125--1134},
}

@article{schwartingSafeNonlinearTrajectory2018,
	title = {Safe {Nonlinear} {Trajectory} {Generation} for {Parallel} {Autonomy} {With} a {Dynamic} {Vehicle} {Model}},
	volume = {19},
	issn = {1558-0016},
	url = {https://ieeexplore.ieee.org/abstract/document/8207782},
	doi = {10.1109/TITS.2017.2771351},
	number = {9},
	urldate = {2026-02-02},
	journal = {IEEE Transactions on Intelligent Transportation Systems},
	author = {Schwarting, Wilko and Alonso-Mora, Javier and Paull, Liam and Karaman, Sertac and Rus, Daniela},
	month = sep,
	year = {2018},
	pages = {2994--3008},
}

@article{schreierIntegratedApproachManeuverBased2016,
	title = {An {Integrated} {Approach} to {Maneuver}-{Based} {Trajectory} {Prediction} and {Criticality} {Assessment} in {Arbitrary} {Road} {Environments}},
	volume = {17},
	issn = {1558-0016},
	url = {https://ieeexplore.ieee.org/abstract/document/7412746},
	doi = {10.1109/TITS.2016.2522507},
	number = {10},
	urldate = {2026-02-02},
	journal = {IEEE Transactions on Intelligent Transportation Systems},
	author = {Schreier, Matthias and Willert, Volker and Adamy, Jürgen},
	month = oct,
	year = {2016},
	pages = {2751--2766},
}

@article{alfriendProbabilityCollisionError1999,
	title = {Probability of {Collision} {Error} {Analysis}},
	volume = {1},
	copyright = {https://www.springernature.com/gp/researchers/text-and-data-mining},
	issn = {1388-3828, 1572-9664},
	url = {https://link.springer.com/10.1023/A:1010056509803},
	doi = {10.1023/A:1010056509803},
	language = {en},
	number = {1},
	urldate = {2026-02-02},
	journal = {Space Debris},
	author = {Alfriend, Kyle T. and Akella, Maruthi R. and Frisbee, Joseph and Foster, James L. and Lee, Deok-Jin and Wilkins, Matthew},
	month = mar,
	year = {1999},
	pages = {21--35},
}

@article{pateraGeneralMethodCalculating2001,
	title = {General {Method} for {Calculating} {Satellite} {Collision} {Probability}},
	volume = {24},
	issn = {0731-5090, 1533-3884},
	url = {https://arc.aiaa.org/doi/10.2514/2.4771},
	doi = {10.2514/2.4771},
	language = {en},
	number = {4},
	urldate = {2026-02-02},
	journal = {Journal of Guidance, Control, and Dynamics},
	author = {Patera, Russell P.},
	month = jul,
	year = {2001},
	pages = {716--722},
}

@incollection{kenchanCollisionProbabilityAnalyses2001,
	series = {Proceedings of a conference sponsored {NASA} {Goddard} {Space} {Flight} {Center}},
	title = {Collision {Probability} {Analyses} {For} {Earth}-{Orbiting} {Satellites}},
	url = {https://ntrs.nasa.gov/api/citations/20010084958/downloads/20010084958.pdf},
	urldate = {2026-02-02},
	booktitle = {2001 {Flight} {Mechanics} {Symposium}},
	publisher = {National Aeronautics and Space Administration},
	author = {Ken, Chan},
	month = jun,
	year = {2001},
	pages = {53--67},
}

@inproceedings{tolksdorfFastCollisionProbability2024,
	title = {Fast {Collision} {Probability} {Estimation} for {Automated} {Driving} using {Multi}-circular {Shape} {Approximations}},
	issn = {2642-7214},
	url = {https://ieeexplore.ieee.org/abstract/document/10588731},
	doi = {10.1109/IV55156.2024.10588731},
	urldate = {2026-01-29},
	booktitle = {2024 {IEEE} {Intelligent} {Vehicles} {Symposium} ({IV})},
	author = {Tolksdorf, Leon and Birkner, Christian and Tejada, Arturo and Van De Wouw, Nathan},
	month = jun,
	year = {2024},
	note = {ISSN: 2642-7214},
	pages = {2529--2536},
}

@inproceedings{lambertCollisionProbabilityAssessment2008,
	title = {Collision {Probability} {Assessment} for {Speed} {Control}},
	issn = {2153-0017},
	url = {https://ieeexplore.ieee.org/document/4732692},
	doi = {10.1109/ITSC.2008.4732692},
	urldate = {2026-01-29},
	booktitle = {2008 11th {International} {IEEE} {Conference} on {Intelligent} {Transportation} {Systems}},
	author = {Lambert, Alain and Gruyer, Dominique and Pierre, Guillaume Saint and Ndjeng, Alexandre Ndjeng},
	month = oct,
	year = {2008},
	note = {ISSN: 2153-0017},
	pages = {1043--1048},
}

@article{alfanoNumericalImplementationSpherical2005,
	title = {A {Numerical} {Implementation} of {Spherical} {Object} {Collision} {Probability}},
	volume = {53},
	issn = {2195-0571},
	url = {https://doi.org/10.1007/BF03546397},
	doi = {10.1007/BF03546397},
	language = {en},
	number = {1},
	urldate = {2026-01-29},
	journal = {The Journal of the Astronautical Sciences},
	author = {Alfano, Salvatore},
	month = mar,
	year = {2005},
	pages = {103--109},
}

@article{jamesleeforsterjrParametricAnalysisOrbital1992,
	title = {A {Parametric} {Analysis} of {Orbital} {Debris} {Collision} {Probability} and {Maneuver} {Rate} for {Space} {Vehicles}},
	url = {https://stacks.stanford.edu/file/druid:dg552pb6632/Foster-estes-parametric_analysis_of_orbital_debris_collision_probability.pdf},
	publisher = {NASA/JSC-25898},
	author = {{James Lee Forster, Jr} and {Herbert S. Estes}},
	month = aug,
	year = {1992},
}

@article{gouletProbabilisticConstraintTightening2022,
	title = {Probabilistic constraint tightening techniques for trajectory planning with predictive control},
	volume = {359},
	issn = {00160032},
	url = {https://linkinghub.elsevier.com/retrieve/pii/S0016003222003957},
	doi = {10.1016/j.jfranklin.2022.06.005},
	language = {en},
	number = {12},
	urldate = {2026-01-29},
	journal = {Journal of the Franklin Institute},
	author = {Goulet, Nathan and Wang, Qian and Ayalew, Beshah},
	month = aug,
	year = {2022},
	pages = {6142--6172},
}
	\bibliographystyle{ieeetr}                                                    
	
	

\end{document}